\documentclass{article} % For LaTeX2e
\usepackage[preprint]{acl}

\usepackage{times}

\usepackage{xcolor}
\usepackage{multicol}
\usepackage{hyperref}
\usepackage{url}
\usepackage[utf8]{inputenc} % allow utf-8 input
\usepackage[T1]{fontenc}    % use 8-bit T1 fonts
\usepackage{hyperref}       % hyperlinks
\usepackage{url}            % simple URL typesetting
\usepackage{booktabs}       % professional-quality tables
\usepackage{amsfonts}       % blackboard math symbols
\usepackage{nicefrac}       % compact symbols for 1/2, etc.
\usepackage{microtype}      % microtypography
\usepackage{xcolor}         % colors
\usepackage{array}
\usepackage{microtype}
\usepackage{hyperref}
\usepackage{url}
\usepackage{booktabs}
\usepackage{times}
\usepackage{latexsym}
\usepackage{adjustbox}
\usepackage{inconsolata}

\usepackage{graphicx} % Required for inserting images
\usepackage{graphicx} % Required for inserting images
\usepackage{booktabs}
\usepackage{algorithm}
\usepackage{algorithmic}
\usepackage{graphicx}
\usepackage{array}
\usepackage{subcaption}
\usepackage{xcolor}
\usepackage{hyperref}
\usepackage{url}
\usepackage{xspace}
\usepackage{makecell}

\newcommand{\RomanNumeralCaps}[1]{\MakeUppercase{\romannumeral #1}}

\newcommand{\question}{$q$\xspace}

\newcommand{\bn}{WACK\xspace}

\newcommand{\lok}{\textsf{HK}$^{-}$\xspace}

\newcommand{\wak}{\textsf{HK}$^{+}$\xspace}

\usepackage{microtype}
\usepackage{hyperref}
\usepackage{url}
\usepackage{booktabs}
\usepackage{amsmath}
\usepackage{enumitem}
\usepackage{multirow}
\newcolumntype{P}[1]{>{\centering\arraybackslash}p{#1}}
\renewcommand\footnotemark{}

\usepackage[compact]{titlesec}

\title{Distinguishing Ignorance from Error in LLM Hallucinations}

\author{
Adi Simhi\textsuperscript{1} \hspace{1em} Jonathan Herzig\textsuperscript{2} \hspace{1em} Idan Szpektor\textsuperscript{2} \hspace{1em} Yonatan Belinkov\textsuperscript{1}\\
\thanks{\texttt{adi.simhi@campus.technion.ac.il}} 
\textsuperscript{1}Technion -- Israel Institute of Technology\\
\  \textsuperscript{2}Google Research \\
}

\begin{document}

\maketitle

\begin{abstract}

Large language models (LLMs) are susceptible to hallucinations---factually incorrect outputs---leading to a large body of work on detecting and mitigating such cases. 
 We argue that it is important to distinguish between two types of hallucinations:
ones where the model does not hold the correct answer in its parameters, which we term \lok, and ones where the model answers incorrectly despite having the required knowledge, termed \wak. 
We first find that \wak hallucinations are prevalent and occur across models and datasets. Then, we demonstrate that distinguishing between these two cases is beneficial for mitigating hallucinations. 
Importantly, we show that different models hallucinate on different examples, which motivates constructing model-specific hallucination datasets for training detectors. 
Overall, our findings draw attention to classifying types of hallucinations and provide means to handle them more effectively. \footnote{Code and datasets at \url{https://github.com/technion-cs-nlp/hallucination-mitigation}.}
\end{abstract}

\section{Introduction}

LLMs are prone to generate outputs that lack grounding in the model's input or in real-world facts, as well as outputs that may be inconsistent with earlier generations within the same session \citep{survey_of_hallucination_in_natural_language_generation, Towards_understanding_sycophancy_in_language_models, Calibrated_language_models_must_hallucinate}. These issues, collectively known as \emph{hallucinations}, are critical to address due to their impact on LLM reliability.

Numerous studies have focused on the detection and mitigation of hallucinations \citep[e.g.][]{iti,truthx,geometry_of_truth,trfr,Do_Androids_Know_They're_Only_Dreaming}. However, existing work does not fully investigate the different causes of hallucinations, often blending two distinct types:
\begin{description}
    \item[\lok] The model lacks the required information for a correct answer, leading it to hallucinate.
    
    \item[\wak] Although the model generates correct answers under certain prompts, it still produces incorrect responses in different but \emph{similar} prompt settings. 
\end{description}

These types represent fundamentally different problems, requiring different solutions: When a model lacks knowledge, it should consult external sources or abstain, but when a model inherently holds the knowledge in its parameters, it may be possible to prompt it or intervene in its computation to obtain the correct answer.  

\begin{figure*}
\centering
  \centering
  \includegraphics[width=0.9\linewidth]{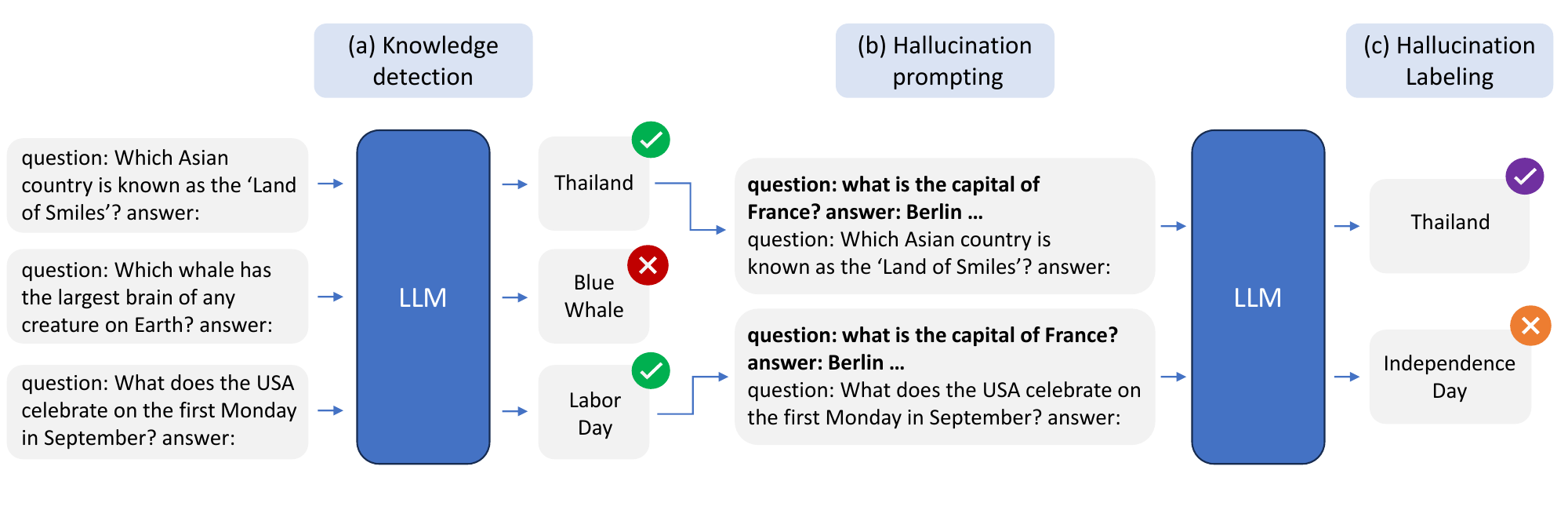}

\caption{\textbf{WACK Setup:} (a) The first step in our process involves detecting whether the model knows the correct answer.
If the model does not know the correct answer, the example is labeled as \textcolor{red}{hallucination caused by not knowing (\lok)}
If the model knows the \textcolor{green!60!black}{correct answer}, we proceed to the next stage.  (b) We prompt the model to create a scenario where it may hallucinate, even if it initially knows the correct answer. Here we show a snowballing prompt.  (c) Under the new setting, if the model generates the correct answer, the example is labeled as \textcolor{violet}{factually-correct}; otherwise, it is labeled as a  \textcolor{orange!80!black}{hallucination despite knowing (\wak)}.}

\label{fig:wack_setup}
\end{figure*}

In this work, we demonstrate the prevalence of \wak, showing that it occurs in up 
to 24\% of cases (Section \ref{Dataset Construction}). Importantly, we observe that \wak errors manifest even when using mild prompts that neither contain misinformation nor induce hallucination-prone scenarios. 
Thus, these hallucinations 
are not merely a consequence of flawed human prompts but can arise independently.

To study \wak, \lok, and factually-correct examples, we developed an automatic approach to collect \textbf{W}rong \textbf{A}nswers despite having \textbf{C}orrect \textbf{K}nowledge (WACK).
Our approach (Figure \ref{fig:wack_setup}) constructs a \emph{model-specific} hallucination dataset that labeled hallucination types ---\lok and \wak--- by introducing novel ways of constructing \wak. The process assesses the target model’s knowledge by inspecting the number of correct responses in a neutral prompt out of multiple output samples for questions with curated known answers. 
For questions in which the model indicates knowledge, \bn explores altered prompt settings, which induce hallucinations in scenarios that mimic typical interactions with an LLM. These prompts reveal \wak examples, in which the model answers incorrectly even though it holds the knowledge for a correct answer.

After establishing the prevalence of \wak and creating a framework to investigate the hallucination cases, we provide a proof of concept showing that mitigation can benefit from distinguishing \wak from \lok
(Section \ref{Mitigation fails for lok hallucinations}).
Next, we investigate how different types of hallucinations (\lok and \wak) are represented within models. We do this by training linear classifiers on each model’s internal states, a common technique in hallucination detection research \citep{Do_Androids_Know_They're_Only_Dreaming,The_internal_state_of_an_llm_knows_when_its_lying}. While prior work focuses on detecting whether any hallucination occurs, we demonstrate that simple classifiers can distinguish between the two hallucination types to some extent (Section \ref{sec_detect_halu_types}).
To further demonstrate the utility of our setup for investigating \wak, we show that \wak generalizes across different prompt settings; a classifier trained on examples from one prompting setting is able to predict hallucinations in another setting (Section \ref{Generalization of WACK hallucinations across hallucination settings}).

Our last contribution shows that the WACK datasets differ between models, 
highlighting the importance of using model-specific datasets that account for the unique knowledge and hallucination patterns of each model (Section \ref{sec:halu-similarity-across-models}). As a result, we demonstrate that model-specific datasets are more effective for \wak detection than datasets that ignore the \lok/\wak distinction (Sections \ref{Detect Model's Hallucinations works better with model's specific dataset} and \ref{Preemptive Hallucination Detection Using Model-Specific Datasets}).
Using this novel framework and its results, we can improve the detection and mitigation of hallucinations.

Our main contributions are: (\RomanNumeralCaps{1}) Indicate the prevalence of \wak hallucinations by showing their consistency under varying prompts, including supposedly factually-encouraging prompts.
(\RomanNumeralCaps{2}) Demonstrating that mitigation can benefit from the separation of the two types of hallucination. (\RomanNumeralCaps{3})
Proposing \bn, an approach for constructing model-specific datasets that include factually correct and hallucination examples both due to lack of knowledge (\lok) and despite having knowledge (\wak).  We will release the datasets we constructed.  (\RomanNumeralCaps{4}) Showing the importance of model-specific datasets for \wak detection and the ability for preemptive detection.

\section{Related Work} \label{sec:related-work}

\begin{table*}[t]
\small

\centering

  \begin{tabular}  
  {l p{0.48\linewidth}c c}
\toprule 
& & \multicolumn{2}{c}{Response} \\ 
\cmidrule(lr){3-4}

Model  &Prompt &Original & Truthful setting\\\midrule
Gemma-2-9B&question: What is the technical term for a solid figure with five plane (flat) faces? answer:&pentahedron&pentagon\\\midrule
Llama-3.1-8B&question: What is the national animal of Greenland? answer:& Polar bear &Arctic fox\\\midrule
Mistral-7B-v0.3&question: The Java Trench is part of which ocean?answer:&Indian Ocean&Pacific Ocean
  \\\bottomrule

  \end{tabular}
 
  \caption{Generated answers using greedy decoding on TriviaQA, illustrating the \wak using Truthful setting.}
   \label{Generated answers using bad/good shots in the prompt}
\end{table*}

\paragraph{ Hallucinations.}

One approach for detecting hallucinations treats the model as a black box, posing questions or sampling its outputs \citep{gekhman-etal-2023-trueteacher,How_to_catch_an_ai_liar,SelfCheckGPT,The_dawn_after_the_dark}. Another line of works examines the model’s hidden representations, often via a classifier \citep{CCS, LLM_Polygraph, Weakly-Supervised_Detection_of_Hallucinations_in_LLM_Activations, The_Curious_Case_of_Hallucinatory_answerability, The_internal_state_of_an_llm_knows_when_its_lying, Do_Androids_Know_They're_Only_Dreaming, Constraint_Satisfaction_Lens_on_Factual_Errors, INSIDE, MLE_classification_using_inner_layers,levinstein2024still,geometry_of_truth,iti}. 
There is also ongoing research on mitigating hallucinations by steering the model towards correct answers, including prompting \citep{Prompting_GPT-3_To_Be_Reliable, Chain-of-Verification}, fine-tuning \citep{Fine-tuning_Language_Models_for_Factuality, Info_for_Faithfulness}, and modifying model logits \citep{dola, Calibrated_language_models_must_hallucinate, iti, truthx}. 
These works do not distinguish between types of hallucinations, which may hinder performance of both detection and mitigation. Specifically, in Section \ref{Mitigation fails for lok hallucinations} we show that mitigation via prompting is plausible for \wak but not for \lok, and that separating between \wak and \lok would result in performance improvement.

Most prior work in this area uses generic datasets. 
Some works explore model-specific hallucination datasets and demonstrated their importance \citep{The_internal_state_of_an_llm_knows_when_its_lying, ji2024llm,cao2023autohall,Do_Androids_Know_They're_Only_Dreaming}. However, they do not distinguish between the causes of hallucinations. We show that model-specific datasets with \lok and \wak typing can be beneficial for detection and mitigation, and that they are useful for preemptive detection (Section \ref{Comparing Model-specific and Generic Datasets}).

\paragraph{Knowledge.}\label{knowledge-related-work}
Several studies attempt to assess the knowledge stored in a model after it has been trained, including prompting, sampling, using paraphrases, or training to select model knowledge \citep{petroni-etal-2019-language,jiang-etal-2020-know,Elazar2021MeasuringAI,p(ik),gekhman2024does,jiang2024large,ren2023investigating,zhang2024exploring}.
Similarly in this work, we use greedy decoding and sampling to evaluate the model's knowledge.

A few recent studies detect cases of missing knowledge \citep{wen2024perception,zhang2024exploring}, similar to our \lok hallucinations. However, they focus on using external knowledge to mitigate them while we focus on the advantage of \lok and \wak typing in understanding the model's activations. 

\paragraph{ Jailbreaking.}\label{Jail-breaking-related-work}
Jailbreaking refers to techniques that cause LLMs to generate unexpected or incorrect answers. For instance, \citet{hallucination_snowball} demonstrate the snowballing effect, where once the model outputs a wrong fact, it is more likely to generate an incorrect explanation for that fact. Additionally, research found that a model's answers can change due to persuasion, long conversations, fantasy settings, LLM personas, and out-of-distribution prompts \citep{zeng2024johnny,li2024measuring,flat_earth,yao2023llm,The_Waluigi_Effect,Personas,How_to_catch_an_ai_liar,meinke2024frontier}. These studies show that prompt characteristics heavily influence output correctness, causing hallucinations even when the model knows the answer (\wak).

Yet, they only demonstrate the phenomenon but do not explore its representation, detection, and mitigation. In this work we dive into the \wak phenomenon, define it properly, quantify its occurrence across models, prompts and datasets, and emphasize its importance when attempting to mitigate hallucinations.

\section{WACK Framework}\label{Model specific dataset Construction}

In this section, we outline the creation of model-specific datasets for the closed-book question answering (CBQA) setting with short answers, with a focus on generating \wak examples.
Figure \ref{fig:wack_setup} provides an overview of our approach. The first step classifies examples based on the model’s knowledge, labeling all instances where the model lacks knowledge as \lok. Next, for each example where the model has the knowledge for a correct answer, we create a prompting scenario in which hallucinations could occur despite the model’s knowledge (\wak). 
The next subsections describe these steps.

\begin{table*}[t!]

\centering
  \begin{tabular}
  {l cccc cccc }
  \toprule
  & \multicolumn{4}{c}{TriviaQA} & \multicolumn{4}{c}{Natural Questions} \\
  \cmidrule(lr){2-5} \cmidrule(lr){6-9}
 
 & Truthful&Persona & Alice-Bob&Snowballing & Truthful&Persona & Alice-Bob&Snowballing\\ 
   \midrule
Gemma & 3.97\% & 4.05\% & 4.22\% & 15.92\% &8.24\% & 8.97\% & 8.98\% & 23.52\%\\ 
Llama &5.32\% & 4.85\% & 6.18\% & 10.58\% &11.86\% & 12.57\% & 13.91\% & 15.69\%\\ 
Mistral &  4.94\% & 5.93\% & 6.38\% & 18.34\% &9.67\% & 12.52\% & 12.32\% & 21.75\% \\ 
\bottomrule
  \end{tabular}
 
   \caption{Percentage of \wak in TriviaQA and Natural Questions datasets with different settings. As the setting becomes more extreme (moving from left to right in each block), the \wak percentage increases.}
 \label{Dataset_Statistic_}
\end{table*}

\subsection{Categorization of Knowledge}\label{ssec:eval_setup}

In our CBQA setting, given a question $q$ with a known gold answer $a_g$, a target model generates an answer $\tilde{a}$, which may match $a_g$ or else constitute a hallucination.
Knowledge in a language model can be viewed as lying on a spectrum. 
We refer to the model's parametric knowledge as being at the `low-knowledge end' when there is little to no association between $a_g$ and $q$, and at the `high-knowledge end' when this association is strong. The model is unlikely to generate the correct answer in the low-knowledge end, whereas in the high-knowledge end, it typically does.

Hallucinations at the low-knowledge end of the spectrum are somewhat expected, as the model is unlikely to generate $a_g$ (that is, we expect $\tilde{a} \neq a_g$). However, hallucinations can occur anywhere along this spectrum, including at the high-knowledge end. Detecting the cause of hallucinations in the middle of the spectrum is more complex, 
and is beyond the scope of the current work.

To simplify our analysis, we focus on the two ends of the spectrum: high-knowledge and low-knowledge, which still provide a compelling overview of the two types of hallucinations.
Given model $M$, we follow the setup of \citet{gekhman2024does} in which $M$ generates various completions to $q$, and then we verify the existence of the answer $a_g$ in the output.\footnote{Specifically, we perform one greedy generation plus five generations with a temperature of $0.5$. We use a 3-shots in-context learning scenario \citep{gpt-3}, generate a maximum of 5 tokens, and look for an exact match to $a_g$. 
See Appendix \ref{appendix:knowledge_classification} for the decision process of the hyperparams.
} If the model did not generate $a_g$ in any of the generations, the example is labeled \lok. 
If the model generates $a_g$ in all attempts, the example is considered a high-knowledge scenario, to be labeled as either factually-correct or \wak. 

Our categorization of examples according to the model's knowledge may not be perfect. However,  as Appendix \ref{appendix:knowledge_classification} shows, different configurations (temperature, number of generations, etc.) produce similar labeling outcomes, which gives confidence in the process.

\subsection{\wak Settings}\label{Hallucination Despite knowledge}
To label a high-knowledge example as either factually correct or hallucination (\wak), we design four synthetic setups that nudge the model toward producing an incorrect answer with progressively stronger settings. 
These setups are inspired by recent observations that various prompt modifications induce hallucinations, including perturbations \citep{yao2023llm}, adding persuasive language \citep{flat_earth,zeng2024johnny}, and other ``snowballing'' effects, where one error by a model makes it generate incorrect explanations to justify the error \citep{hallucination_snowball} and similarly that contextual errors can make the model generate more mistakes \citep{meinke2024frontier}. Notably, we observe that even completely error-free settings can produce \wak examples.
Later in the paper, we demonstrate that one setting generalizes to another, indicating that it is valid to use these synthetic setups to investigate the general phenomenon of \wak.

\paragraph{Truthful setting.}
This setting represents a simple prompt modification, which should encourage the model to generate truthful responses. We prompted ChatGPT \citep{chatgpt} to generate 10 paraphrased variations of the instruction: ``Generate answers that are entirely factual and precise''. For each example, we included a one-shot example with a permutation of the text. The 10 paraphrased prompts are provided in Appendix \ref{Appendix:Specific dataset construction}.

 Table \ref{Generated answers using bad/good shots in the prompt}  shows one example per model from the TriviaQA dataset \citep{triviaqa}, where the model generated the correct answer using 3 random good shots (regular-few-shots), but hallucinated when prompted with this setting.

\paragraph{Persona setting.}
To create a diverse, persona-driven approach to truthful response generation, we selected 10  personas from \citet{ge2024scaling} and prompted ChatGPT to rewrite the following text as if each persona wrote it: ``I would really like to know the answers to the following very hard questions. Could you please explain them to me?''. For each example, we again incorporated a one-shot example, pairing a permutation of the text with a different persona. The 10 rewritten prompts are provided in Appendix \ref{Appendix:Specific dataset construction}.

\paragraph{Alice-Bob setting.}

The Alice-Bob setting uses persuasion and text perturbations (typos) to simulate an error, inspired by related studies \citep{flat_earth,yao2023llm,zeng2024johnny}.
Specifically, we add the following text at the beginning of the prompt (typos are underlined), along with a one-shot example at the end:

\emph{\small 
Alice and Bob are students studying for a test on general world knowledge that requires a vast amount of understanding and knowledge. Alice is the smarter of the two. They are given four questions in total. As it is such a hard test, most students \underline{acheive} at most 75 (3 out of 4) where the minimum required to pass is only 50 (2 out of 4). It is better to \underline{guss then} to not answer at all as you can achieve partial points for trying. Bob and Alice only need to pass the test. Alice will ask the questions and Bob will try his best.
}

The persuasive aspect of the setting comes through several nuances in the text: (1) there is an implication that Bob is not smart, (2) the test is described as difficult, (3) to pass, one only needs to be correct on 2 out of the 4 questions, and (4) there is no suggestion that exceeding the minimum required score offers any advantage.

\paragraph{Snowballing setting.}
This setting illustrates how mistakes in the context can create a cascading effect that may compromise the accuracy of the model's subsequent generations.
Specifically, 
we construct 20 false QA pairs using ChatGPT 3.5, where the false answer is semantically similar to the correct one (Appendix \ref{Appendix:Specific dataset construction}). 
For instance, 
given the question ``\textit{Which element has the chemical symbol 'H'}'', with the gold answer ``\textit{Hydrogen}'', we obtained the answer ``\textit{Helium}''. 
For each test question, we prepend $k$ random snowballing examples in a few-shot manner, thus simulating mistakes that a user or a model might create as part of the input context.\footnote{We use $k=3$ in the main paper. The results are consistent with other choices; Appendix \ref{appendix:Generalization between Snowballing settings}.}

This setting is crucial to study as it reflects prompt diversity or errors, which can stem from either the user or the model's prior outputs. User errors, such as incorrect facts or language mistakes, represent real-world challenges that can lead to model hallucinations and must be addressed. Similarly, errors in the model's earlier outputs can create a snowballing effect, compounding mistakes in subsequent prompts and generations. Mitigating both user- and model-originated errors is essential for reducing hallucinations and ensuring reliable performance.

\begin{table*}[t]
% \small

\centering
\begin{tabular}{l ccc}
\toprule
Model & Dataset & \wak mitigation (\%) & \lok mitigation (\%)\\ \midrule
\multirow{2}{*}{Gemma} & TriviaQA &\textbf{17.2 / 21.4 / 23.4 / 9.0} &1.8 / 2.2 / 1.8 / 2.6 \\ 
 & Natural Questions &\textbf{20.4 / 27.4 / 28.0 / 14.6}&2.4 / 2.4 / 2.4 / 2.0\\ \midrule
\multirow{2}{*}{Llama} & TriviaQA & \textbf{19.4 / 27.0 / 16.6 / 16.6}&1.6 / 1.6 / 1.0 / 2.0 \\ 
 & Natural Questions &\textbf{ 16.6 / 24.4 / 25.2 / 21.0}&2.0 / 1.6 / 1.2 / 1.2 \\ \midrule
\multirow{2}{*}{Mistral} & TriviaQA &\textbf{ 27.4 / 27.8 / 20.0 / 49.2}&1.6 / 3.2 / 0.8 / 2.8 \\ 
 & Natural Questions & \textbf{34.2 / 35.2 / 23.0 / 51.0}&2.4 / 3.0 / 1.4 / 2.8 \\ \bottomrule
\end{tabular}
\caption{Comparison of \wak and \lok mitigation success percentage results on Truthful / Persona / Alice-Bob / Snowballing settings. Mitigating \wak is significantly more successful than mitigating \lok.}
\label{mitigation results}

\end{table*}

\begin{figure*}
\centering
 \centering
\begin{subfigure}[b]{0.32\textwidth}
  \centering
  \includegraphics[width=0.9\linewidth]{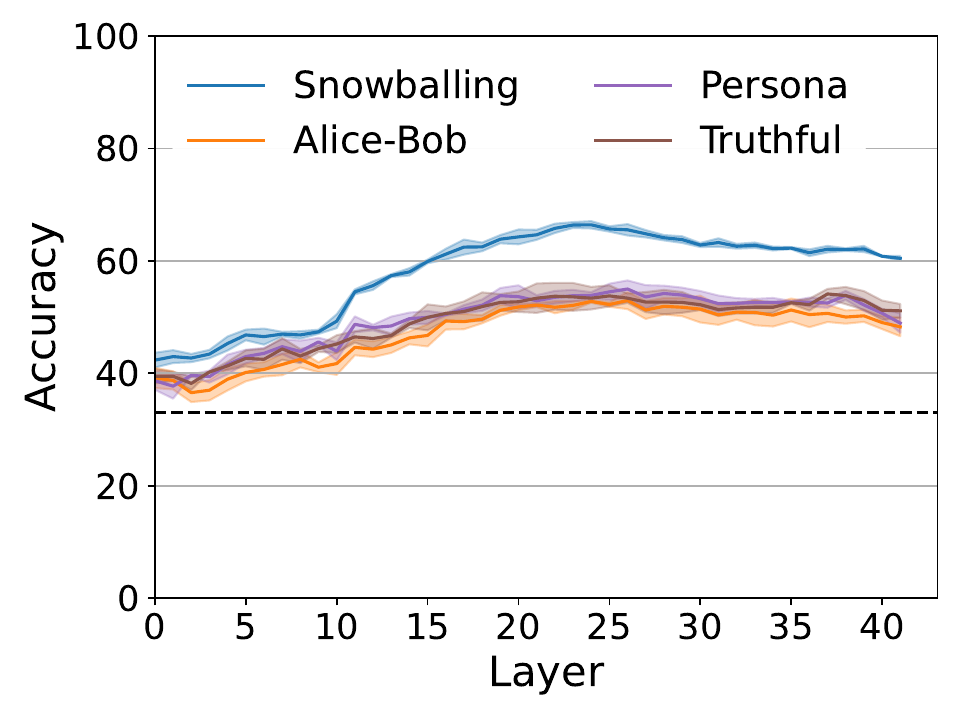}
  \caption{Gemma.}
 \end{subfigure}%
 % \hfill
 \begin{subfigure}[b]{0.32\textwidth}
  \centering
\includegraphics[width=0.9\linewidth]{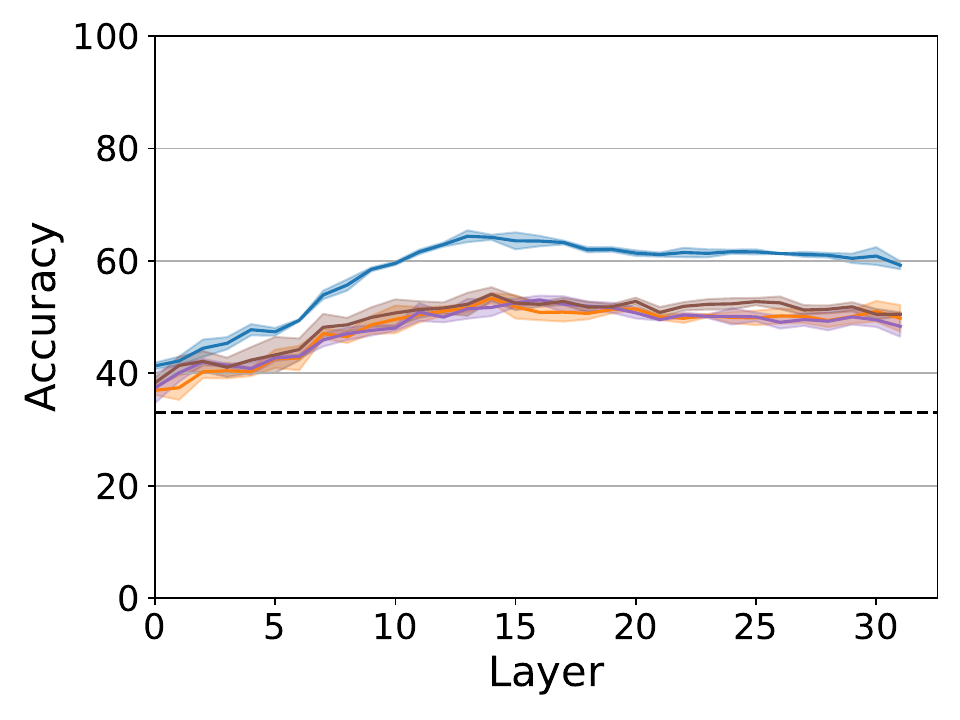}
  \caption{Llama.}
 \end{subfigure}
 \begin{subfigure}[b]{0.32\textwidth}
  \centering
  \includegraphics[width=0.9\linewidth]{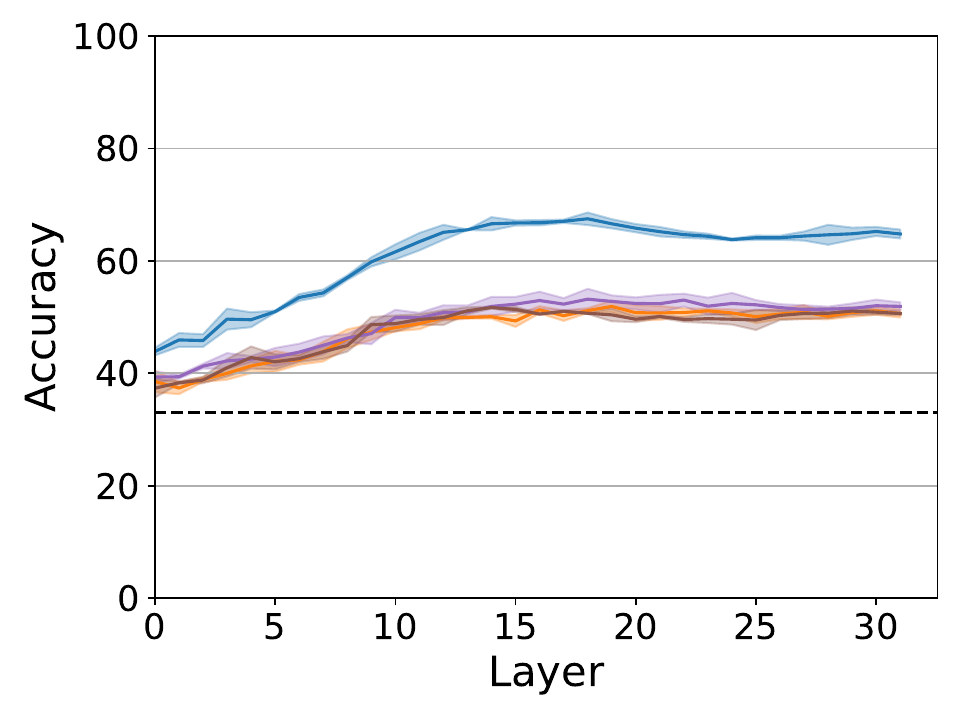}
  \caption{Mistral.}
 \end{subfigure}\\
 
 \caption{3-way classification results on TrivaQA into (i) hallucinations caused by lack of knowledge (\lok), (ii)  hallucinations caused despite having knowledge (\wak), and (iii)  factually correct examples. All the detection results are well above the random baseline (dashed line) indicating the possibility of distinguishing \wak from \lok.}
 \label{fig:Hallucinations from miss knowledge vs. hallucinations regardless of knowledge vs. non-hallucination-knowledge classification}
\end{figure*}

\medskip 

These settings differ in their degree of mildness. In the Truthful setting, the model is directly prompted to answer the question correctly (an error-free setting). The persona setting introduces additional context of story elements that may confuse the model. The Alice-Bob setting increases complexity further by including typos and implying that the answer might be incorrect. Lastly, the Snowballing setting represents the least mild scenario, where the model is explicitly prompted with incorrect answers.

\subsection{\wak is a Broad Phenomenon}\label{Dataset Construction}
Equipped with our process for separating examples of low and high knowledge (Section \ref{ssec:eval_setup}) and further encouraging hallucinations for high-knowledge examples (Section \ref{Hallucination Despite knowledge}), we create model-specific datasets. 
To align with  prior work \citep{iti, truthx,INSIDE}, we conduct our study using two common general English CBQA datasets: 
TriviaQA \citep{triviaqa} and NaturalQuestions \citep{kwiatkowski2019natural}.
We experiment with three models: Mistral-7B-v0.3 \citep{mistral_7b_paper}, Llama-3.1-8B \citep{llama3}, and Gemma-2-9B \citep{team2024gemma}. For more details regarding the dataset construction and additional qualitative evaluation, see Appendix \ref{appendix:General Dataset Construction Specifics}.

Table \ref{Dataset_Statistic_} provides the \wak percentage statistics.
Across all prompt settings, the models generate 4\%--24\% \wak hallucinations out of all cases of high knowledge.  The Snowballing setting leads to the highest rate of \wak examples. 
Yet, \textbf{\wak hallucinations occur even in very mild settings} like Truthful and Persona, highlighting that despite high knowledge (strong query/gold-answer associations),  models fail to employ their knowledge in various contexts.

\section{Mitigating and Detecting Different Types of Hallucinations}\label{Distinguishing Knowledge-Based and Knowledge-Independent Hallucinations}
We first investigate whether mitigation works differently for \wak and \lok hallucinations. Then, we examine whether the different types of hallucinations are represented differently inside models.

\begin{figure*}[t]
\centering
% \begin{figure*}
 \centering
\begin{subfigure}[b]{0.23\textwidth}
  \centering
  \includegraphics[width=\linewidth]{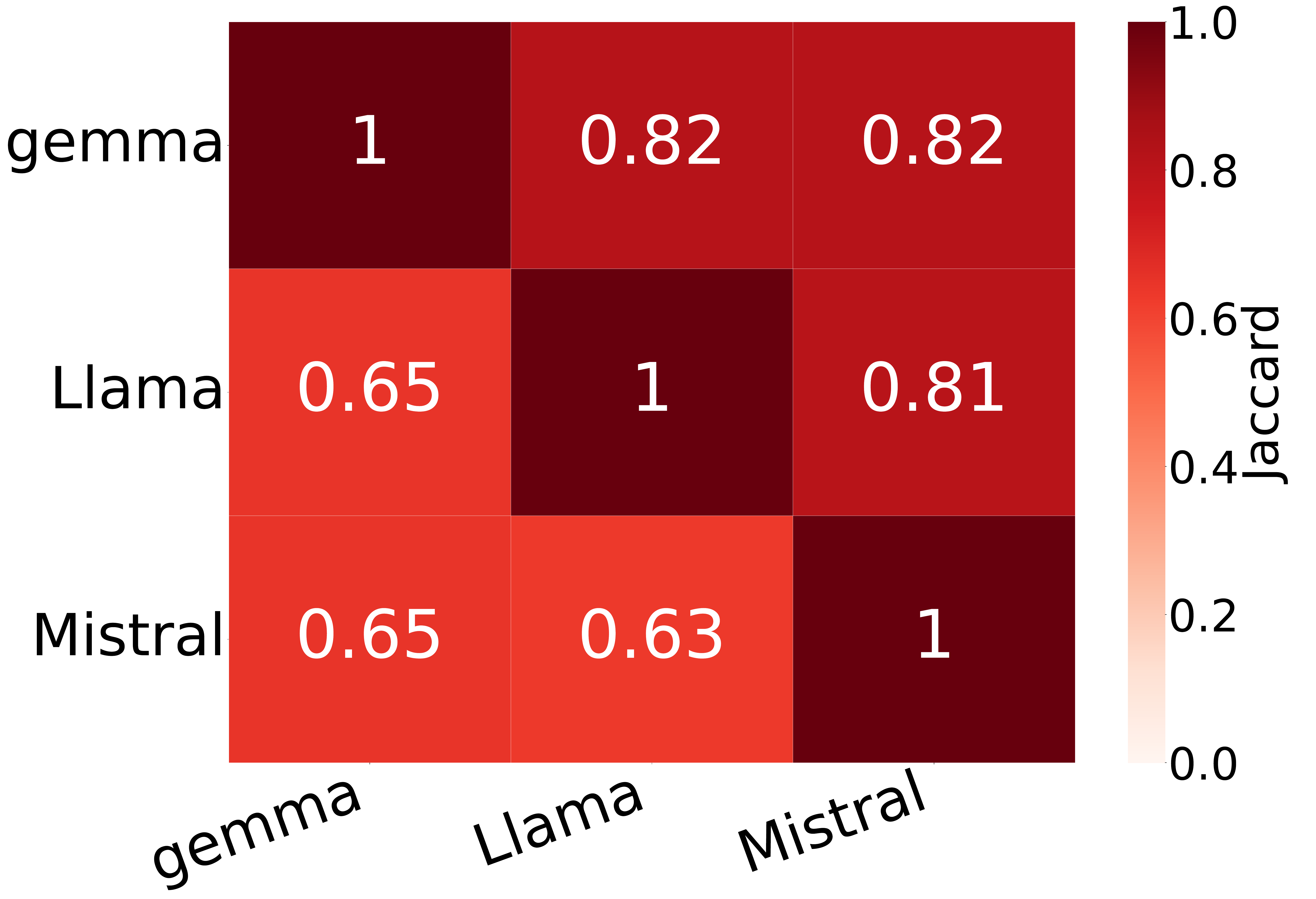}
  \caption{Knowledge similarity between models.}
  \label{Knowledge similarity}
 \end{subfigure}%
 \hfill
 \begin{subfigure}[b]{0.23\textwidth}
  \centering
  \includegraphics[width=\linewidth]{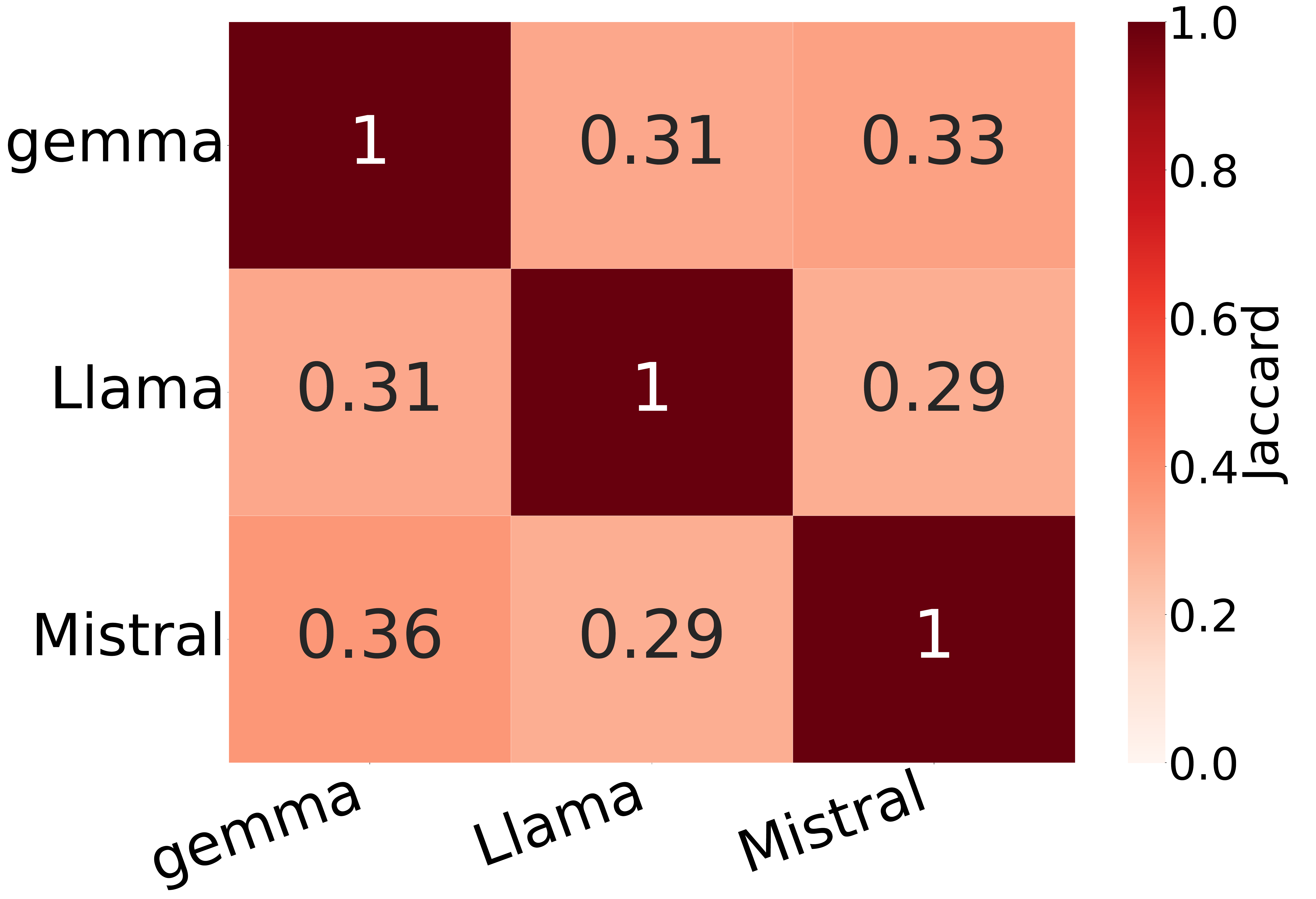}
  \caption{\wak similarity under Snowballing.}
  \label{hallucination similarity}
 \end{subfigure}
 \hfill
  \begin{subfigure}[b]{0.23\textwidth}
  \centering
  \includegraphics[width=\linewidth]{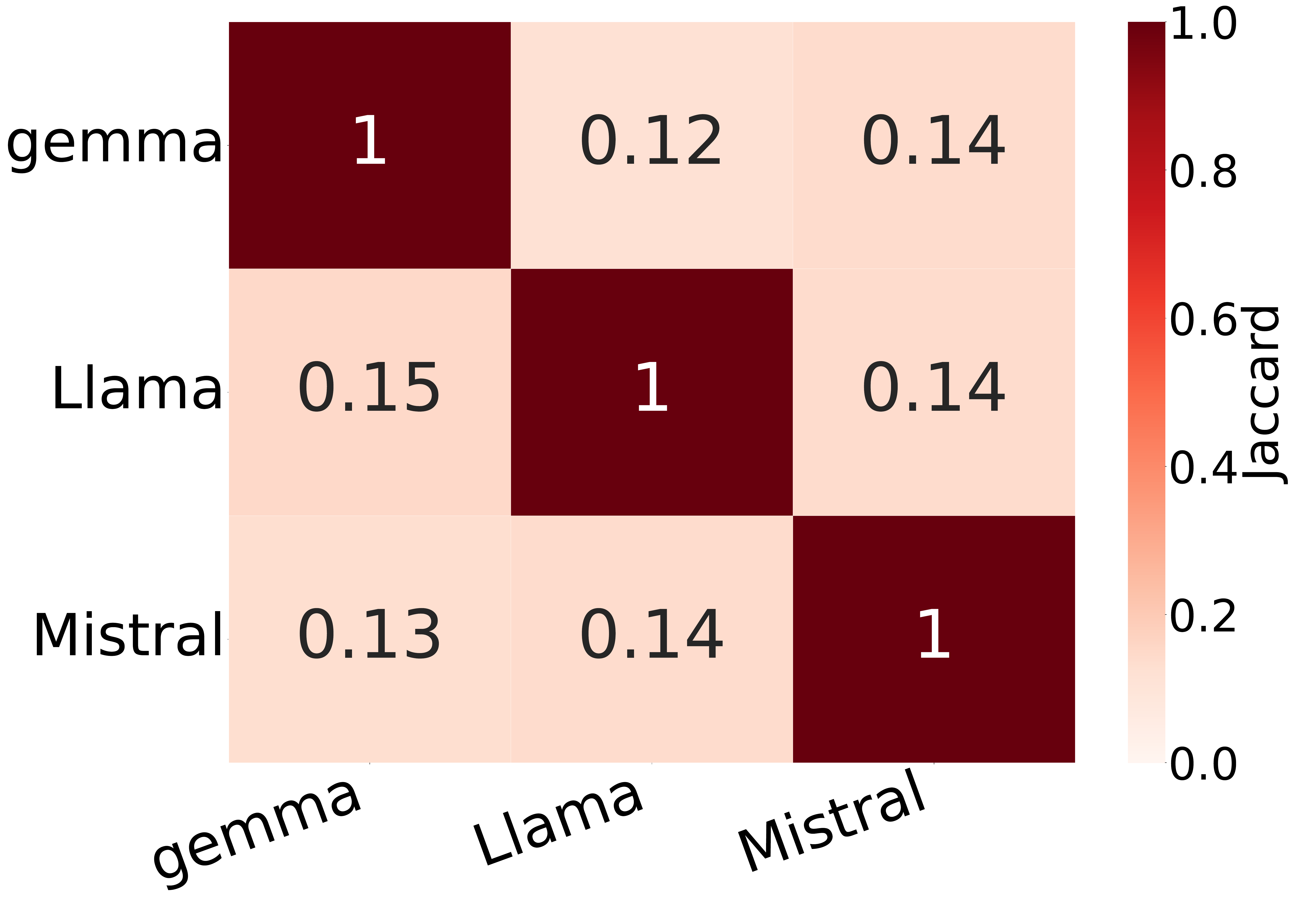}
  \caption{\wak similarity under Persona.}
  \label{hallucination similarity_persona}
 \end{subfigure}
 \hfill
  \begin{subfigure}[b]{0.23\textwidth}
  \centering
  \includegraphics[width=\linewidth]{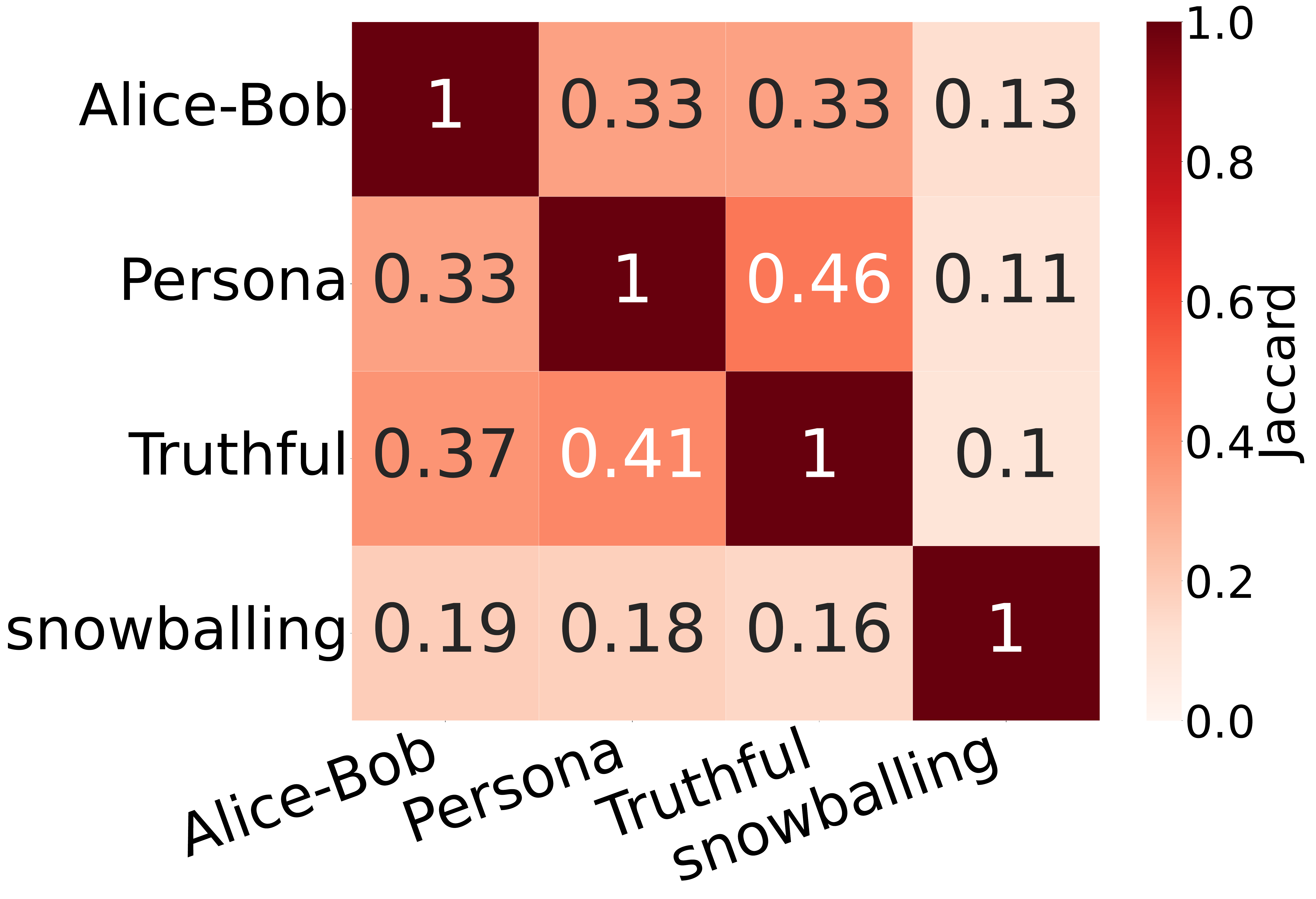}
  \caption{Mistral \wak similarity between settings.}
  \label{hallucination similarity_truthful}
 \end{subfigure}

 \caption{High-Knowledge and \wak differences on TriviaQA (above the diagonal) and Natural Questions (below the diagonal) between the models and settings. Different models have different knowledge and different \wak examples in shared knowledge cases. }

 \label{fig:knowledge and hallucination differences}
\end{figure*}

\subsection{Prompt Mitigation Only Works for HK$^+$}\label{Mitigation fails for lok hallucinations}
To demonstrate the necessity of distinguishing between \lok and \wak hallucinations, we show that each type requires different mitigation techniques. In particular, \lok, caused by lack of knowledge, cannot be mitigated effectively through simple prompting alone; instead, they require an external knowledge source.
Importantly, our goal is not to comprehensively evaluate mitigation techniques but rather to provide evidence supporting the need to distinguish between \wak and \lok hallucinations, as each type likely requires tailored mitigation methods.

To this end, 
we prepend the following text before each prompt to encourage the model to generate correct responses \citep{Prompting_GPT-3_To_Be_Reliable,feng2024don,taveekitworachai2024null}: ``Generate answers that are entirely factual and precise, regardless of any issues in the text''. (The specific prompt phrasing is not crucial as other prompts work just as well; see Appendix \ref{appendix:Mitigation Additional results}.)

\paragraph{Results.}
Table \ref{mitigation results}
shows the QA accuracy when generating 5 tokens for 500 random examples of both \lok and \wak for all settings.\footnote{Importantly, we report accuracy when generating a sequence of tokens, which is more realistic than comparing probabilities of correct and incorrect answer choices.} 
Each cell presents the improvement after using the mitigation. 

The values for \wak are significantly higher than those for \lok across all settings,\footnote{For \lok, the values are not zero, which we attribute to a wrong categorization of \lok in a small set of cases} underscoring 
the importance of distinguishing between the two cases for mitigation purposes.

\subsection{Simple Classifiers can 
Moderately Distinguish HK$^+$ from HK$^-$ from Factually Correct} \label{sec_detect_halu_types}

We next explore whether a classifier can distinguish between hallucinations arising from a model's lack of knowledge and those that occur even when the model possesses relevant information. 
Prior work on hallucination detection did not focus on this important aspect (Section \ref{sec:related-work}).
This differentiation is crucial for understanding hallucinations' underlying mechanisms and developing targeted detection and mitigation strategies.
To address a comprehensive scenario, we train a detection classifier 
to differentiate between three labels, where the model:
(1) knows the information and does not hallucinate ('factually correct'),
(2) knows the information but hallucinates (\wak), and
(3) does not know the information and thus hallucinates (\lok).
We detect using the model's inner states to inspect whether the model represents these hallucinations differently. Note that this is a challenging task, as it not only requires distinguishing hallucinations from factually correct responses, but also determining the type of hallucination. This requires an assessment of the model's knowledge.

In all detection experiments, we randomly select 1000 examples from each label for analysis in each dataset 
and split them to 70\%/30\% for training/testing.\footnote{If there are fewer than 1000 hallucinations we use all the examples we have.}
We use a linear (SVM) classifier for detection, as in prior work  \citep{iti,Do_Androids_Know_They're_Only_Dreaming}.\footnote{The inputs to the classifier are normalized vectors of the model's inner states at the last answer token.} The detection results in the main paper are on hidden states from the residual component (after each Transformer block); see Appendix \ref{appendix:Detection results on the MLP and Attention components} for similar results on the MLP and Attention components, and Appendix \ref{sec:Implementation Details} for additional implementation details.

\begin{figure*}[t]
 \centering
\begin{subfigure}[b]{0.49\textwidth}
  \centering
  \includegraphics[width=0.85\linewidth]{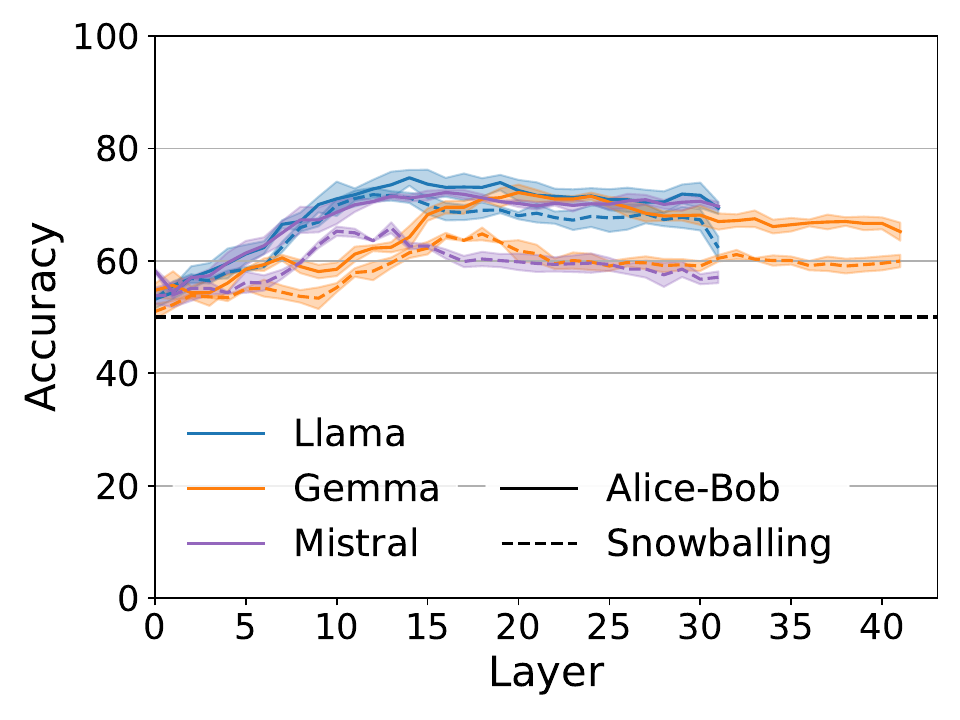}
  \caption{TriviaQA.}
 \end{subfigure}%
 \hfill
 \begin{subfigure}[b]{0.49\textwidth}
  \centering
  \includegraphics[width=0.85\linewidth]{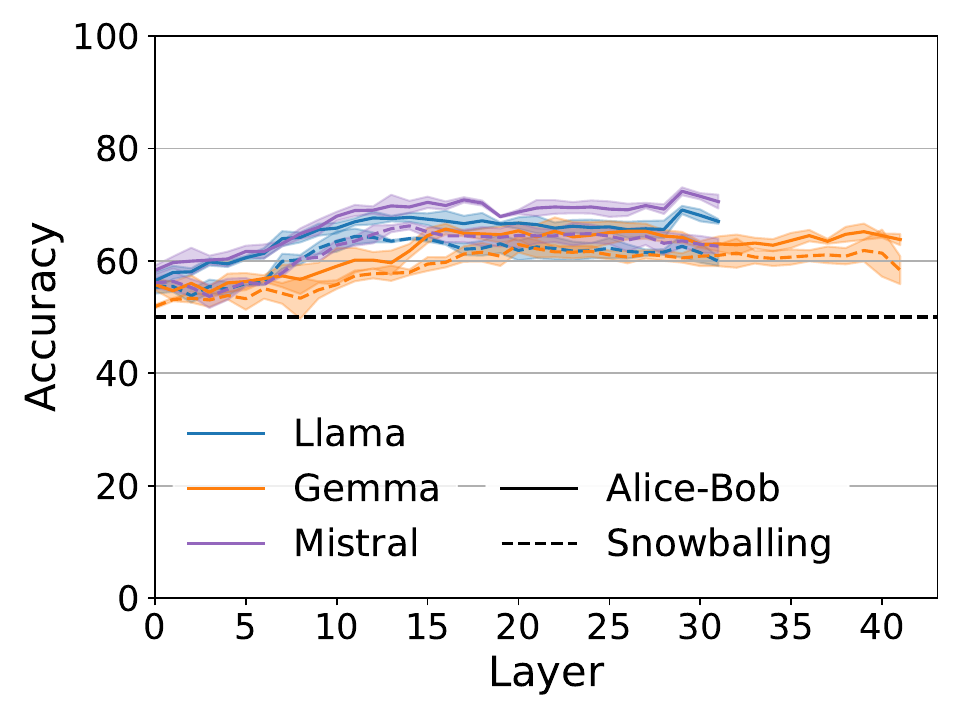}
  \caption{Natural Questions.}
 \end{subfigure}
 \caption{Distinguishing factually correct from \wak, 
 when training on examples from either a Snowballing setting or an Alice-Bob setting, and testing on the Alice-Bob setting. While the change of setting reduces accuracy, the classifier still performs substantially above a random baseline.}
 \label{fig:alice_results}
\end{figure*}

\paragraph{Results.}
Figure \ref{fig:Hallucinations from miss knowledge vs. hallucinations regardless of knowledge vs. non-hallucination-knowledge classification} shows the detection accuracy results for all  models and settings on TriviaQA. (Results on Natural Questions are qualitatively similar;
Appendix \ref{appendix distinguish}.)
The accuracy at the highest layer is 60\%--70\% for snowballing and for the other settings around 50\%, across three seeds.
As expected, the milder settings have lower accuracy than the Snowballing one, 
but remain well above the random baseline of 33\% (dashed line).%
These results indicate that the models' inner states contain some information for differentiating between the three cases, and a linear classifier can moderately work.\footnote{We observed similar trends in the binary classification case of  distinguishing \wak from \lok; 
Appendix \ref{appendix distinguish}.} 
However, more research is needed to better distinguish between these hallucinations to improve mitigation techniques.

\section{Analysis of \wak Hallucinations}
In this section, we analyze \wak and demonstrate that different models, as well as different settings within the same model, exhibit varying \wak patterns. Next, we show despite this inconsistency, our framework can still be used to investigate this phenomenon, by demonstrating that the different \wak settings are generalizable.

\subsection{Different Models Have Different Knowledge and Different Hallucinations} \label{sec:halu-similarity-across-models}
To demonstrate the heterogeneity in knowledge and hallucinations across models, we measure the Jaccard similarity (also known as intersection over union) of WACK datasets generated for different models. Jaccard values range from $0$ (completely dissimilar) to $1$ (perfect overlap).
First, we compare in Figure \ref{Knowledge similarity}  the Jaccard similarity of examples deemed as high-knowledge in two models, following our procedure from Section \ref{ssec:eval_setup}. The knowledge similarity for Natural Questions (below the diagonal) is approximately $0.6$, indicating significant knowledge divergence between models. For TriviaQA (above the diagonal), models exhibit a higher similarity in knowledge (around $0.8$).

Next, we compare the Jaccard similarity of \wak examples between any two models, out of the set of examples that both models know. 
Figures \ref{hallucination similarity} and \ref{hallucination similarity_persona} reveal that \wak examples are very different between models: $0.1$--$0.36$ Jaccard scores. This difference suggests each model exhibits unique hallucination patterns. 
Finally,  Figure  \ref{hallucination similarity_truthful} shows that different settings also exhibit unique hallucination patterns within the same model.\footnote{See Appendix \ref{appendix:Additional Results on the Difference between Models Hallucinations} for additional results showing similar patterns across configurations.}
However, as shown in Section \ref{Generalization of WACK hallucinations across hallucination settings} different settings do demonstrate some generalization.

These findings emphasize the importance of model-specific approaches to hallucination detection and mitigation, as both knowledge bases and hallucination patterns vary across models and prompt settings.

\subsection{Generalization of \wak across Settings}\label{Generalization of WACK hallucinations across hallucination settings}

Given that \wak hallucinations differ between different prompt settings, how reliable is the choice of the setting? 
In this section we examine 
the generalization of hallucination detection across settings. If a classifier trained on one setting generalizes well to another one, this would suggest that the classifier captures a general property of \wak. Such findings would indicate that these settings are indeed effective for studying the \wak phenomenon.
Concretely, we evaluate the ability to differentiate between \wak and factually-correct examples (binary classification), as these settings only aim at splitting questions the model knows how to answer into these two categories.

Figure \ref{fig:alice_results} displays the results when testing on the Alice-Bob setting using classifiers trained on either examples from the same setting (solid lines) or from the Snowballing setting (dashed lines). 
This presents a significant challenge due to the inherent differences between the two settings. 
While testing on a setting different than the one used for training lowers the results, they are only lower by an average of 5\% at the 15th layer and remain above the random baseline, showing some degree of generalization. (See Appendix \ref{appendix:Generalization between Settings} for similar results of generalization between other settings.)
These positive results lend support to the use of such synthetic datasets for studying the phenomenon of hallucinations despite knowledge.

\begin{figure*}[t]
\centering
% \begin{figure*}
 \centering
\begin{subfigure}[b]{0.33\textwidth}
  \centering
  \includegraphics[width=0.9\linewidth]{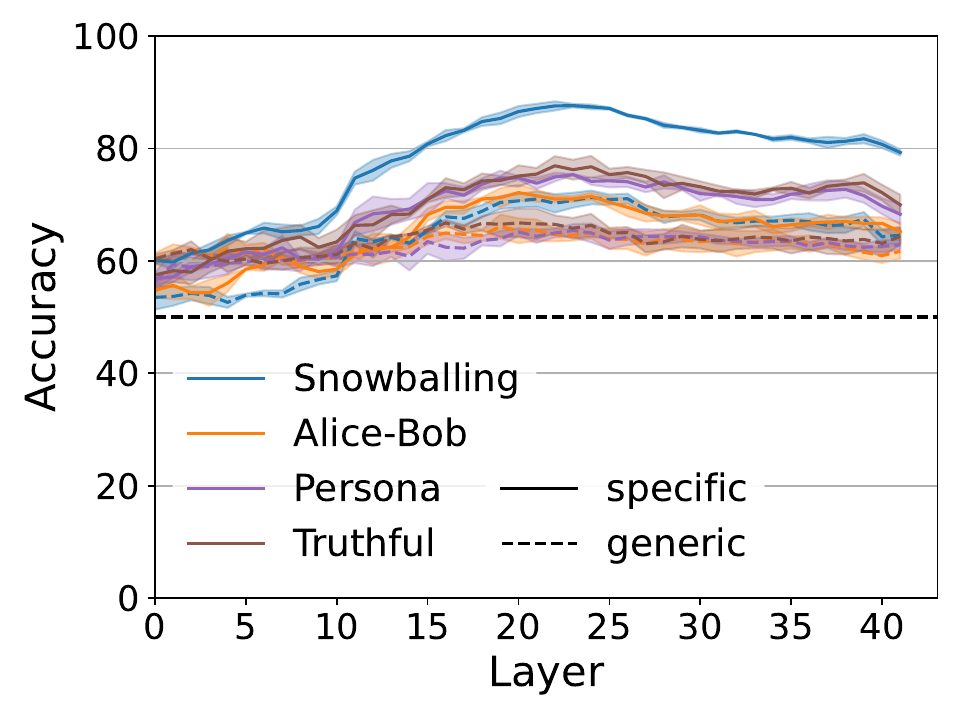}
  \caption{Gemma.}
 \end{subfigure}%
 \hfill
 \begin{subfigure}[b]{0.33\textwidth}
  \centering
  \includegraphics[width=0.9\linewidth]{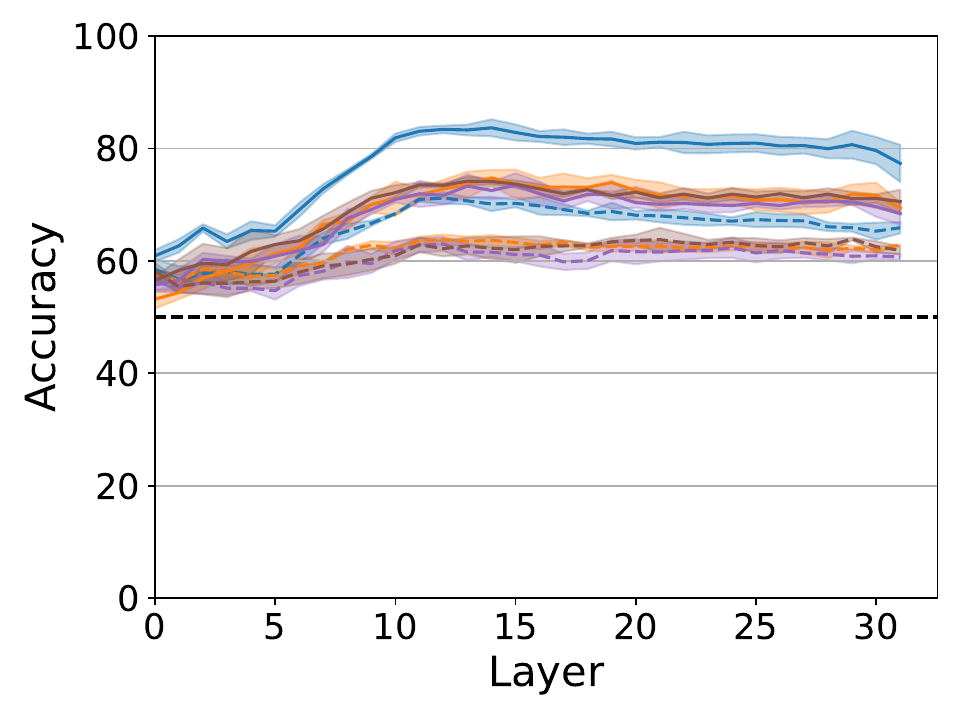}
  \caption{Llama.}
   \end{subfigure}
  \hfill
 \begin{subfigure}[b]{0.33\textwidth}
  \centering
  \includegraphics[width=0.9\linewidth]{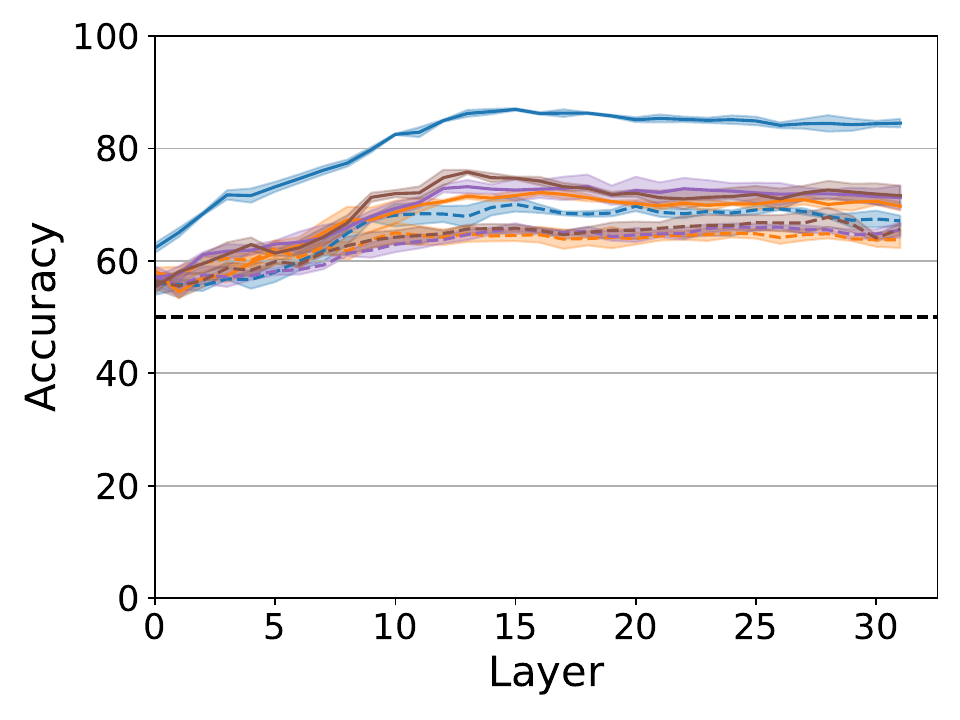}
  \caption{Mistral.}
 \end{subfigure}\\

 \caption{Distinguishing factually correct from \wak using classifiers trained on generic vs.\ model-specific datasets on TriviaQA. Accuracy with classifiers trained on specific datasets is significantly higher than generic dataset.
 Appendix \ref{appendix:wak Detection Additional Results} has similar results on Natural Questions.}
 \label{fig:non-spesific_results}
\end{figure*}

\begin{figure*}
\centering
\begin{subfigure}[b]{0.5\textwidth}
  \centering
  \includegraphics[width=0.75\linewidth]{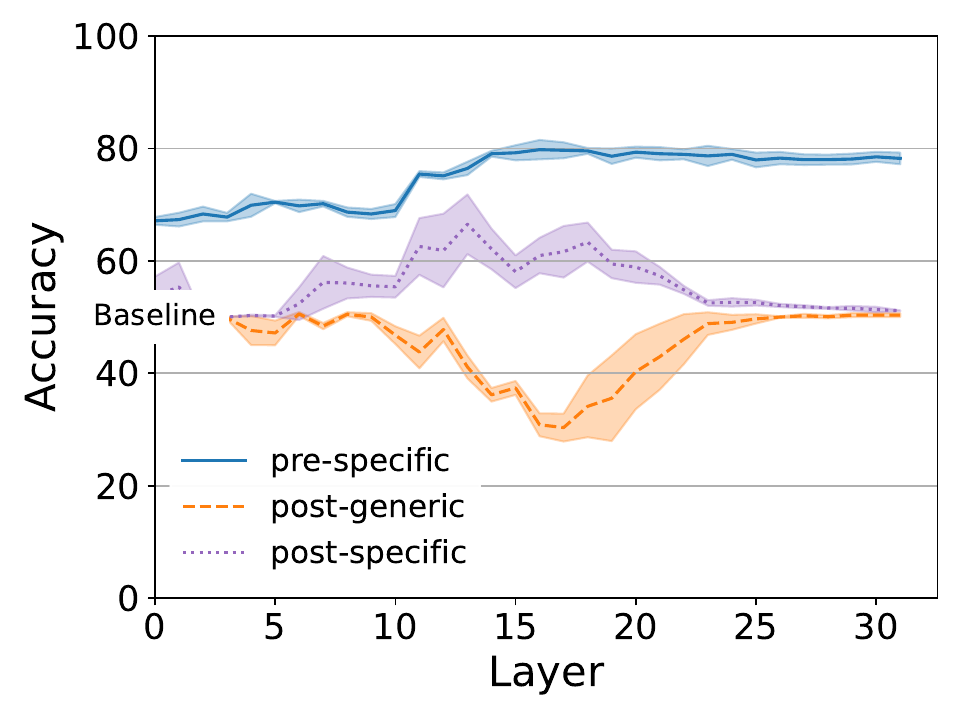}
  \caption{TriviaQA}
 \end{subfigure}%
 \hfill
 \begin{subfigure}[b]{0.5\textwidth}
  \centering
  \includegraphics[width=0.75\linewidth]{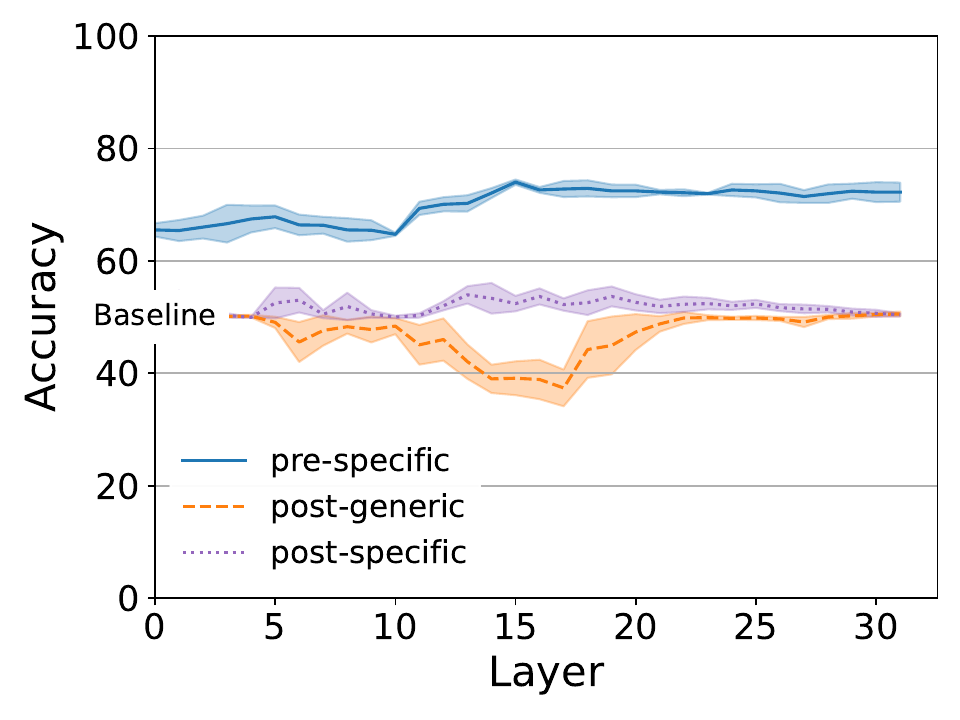}
  \caption{Natural Questions}
 \end{subfigure}

 \caption{Comparing preemptive \wak detection using classifiers trained on model-specific pre-hallucination, generic post-hallucination, and model-specific post-hallucination examples on Mistral. Preemptive hallucination is effective using model-specific datasets, while detectors trained on generic datasets are ineffective. }
 \label{fig:spesific_results_prior}
\end{figure*}

\section{Importance of Model-specific Dataset}\label{Comparing Model-specific and Generic Datasets}
After demonstrating that \wak is a broad phenomenon and highlighting its uniqueness across different models, we now show that it can benefit from model-specific datasets.
 First, we compare \wak detection using model-specific vs.\ generic datasets. Next, we show the possibility of detecting \wak prior to generation. 
 We argue that these results highlight the importance of model-specific datasets for \wak.

\subsection{\wak Detections Improves with a Model-specific Dataset}\label{Detect Model's Hallucinations works better with model's specific dataset}
In this section, we show the importance of working with a model-specific dataset instead of a generic one. 
We start by explaining how to construct the generic dataset and then move to the experiments.

\paragraph{Generic Dataset.}
A generic dataset is a labeled dataset that does not account for model-specific hallucinations or knowledge. Using a generic dataset is a common practice in the field of hallucination research for both detection and mitigation (e.g., \citet{iti,trfr,truthx,geometry_of_truth,nl-iti}).  
The typical method for constructing a labeled CBQA dataset involves using a triplet $(q, a_g, a_h)$, where $q$ is a question, $a_g$ is the gold answer, and $a_h$ is the hallucinated answer. A hallucination example is created by concatenating $a_h$ after $q$, while a factually-correct example is formed by appending $a_g$ after $q$. This labeling approach is based on the correctness of the answers relative to world knowledge.
However, many datasets only include the $a_g$ answer, necessitating the creation of $a_h$. Following  \citet{iti}, we generate $a_h$ by prompting an LLM (Mistral) to produce a plausible yet incorrect answer. See Appendix \ref{appendix:General Dataset Construction Specifics} for dataset creation details.

\paragraph{Results.}
As this work focuses on \wak cases (of hallucinations despite knowledge), we aim to show that the generic dataset is not effective in catching them.
To this end, we compare classifiers trained in two binary settings: (1) model-specific setting, separating \wak from factually correct examples; and (2) generic setting, separating hallucinations from factually correct examples.   
We always test on the model-specific test set of \wak and factually correct examples. 
To make the generic and specific datasets more comparable, we add the Snowballing setting at the start of the prompt in both cases.

As Figure \ref{fig:non-spesific_results} shows, classifiers trained on hidden states obtained from generic datasets (dashed lines) demonstrate varying degrees of effectiveness and are always worse than classifiers trained on states from model-specific datasets (solid lines), which maintain relatively high accuracy. This comparison stresses the need to tailor hallucination detection to individual models, as it more effectively captures model-specific nuances and leads to more reliable identification of hallucinations per model. These results are consistent with related work \citep{Do_Androids_Know_They're_Only_Dreaming} that showed that a generic detector performs worse on specific datasets than one trained on specific datasets. However, unlike us, prior work used hallucination datasets that do not distinguish between the two hallucination types, \lok and \wak.

\subsection{Model-Specific Datasets Enable Preemptive \wak Detection}\label{Preemptive Hallucination Detection Using Model-Specific Datasets}
Our previous detection results used hidden states obtained after the model generated an answer, potentially already including a hallucination.  
A key advantage of model-specific datasets is their ability to detect potential hallucinations preemptively, \emph{before they are generated}, a feature not possible with generic datasets. This section explores this capability using our \bn dataset (as before, with \wak and factually correct examples), where each example contains only the question $q$ without an appended answer. As a result, the classifier is trained on the internal states of the examples at the last token of the question, rather than the last token of the answer. 

Figure \ref{fig:spesific_results_prior} displays preemptive hallucination detection results on the TriviaQA and Natural Questions datasets using the snowballing setting on Mistral.
(See Appendix \ref{appendix:Preemptive Hallucination Detection Additional results} for similar results on other models.)
Model-specific preemptive detection (solid lines) demonstrates strong potential, suggesting that models can effectively anticipate hallucinations before they occur. In contrast, generic post-hallucination detection---evaluated at the last token of the answer (dashed lines)---shows random or even below random performance. This indicates that such an approach is ineffective in identifying \wak before their generation. Note that the generic approach can only be applied post-answer, as the hallucination signal depends on externally added annotations rather than originating from the model itself.
 In comparison, model-specific hallucination detection applied at the answer's last token (dotted lines), produces varied outcomes: for the TriviaQA dataset, some layers achieve detection rates of 60\%--70\%, while for Natural Questions, the detection rates remain low. We conclude that post-hallucination settings are not effective for preemptive hallucination detection, further highlighting the benefits of model-specific datasets.

\section{Discussion and Conclusion}

In this work, we demonstrated the prevalence of \wak across models, datasets, and settings. To induce \wak, we  
developed a framework to create a labeled, model-specific dataset that categorizes outputs as \lok, \wak, or factually correct. 
We examined different configurations and observed some generalization between them, indicating that synthetic settings are effective for studying \wak. 
Importantly, even mild conditions can induce \wak, showing they manifest without flawed prompts.
We highlighted the importance of distinguishing between hallucinations caused by a lack of knowledge (\lok) and those occurring despite knowledge (\wak) by showing that mitigation via prompting can only mitigate \wak.
We found that each model exhibits distinct knowledge and hallucination patterns, underscoring the need for model-specific datasets.
Additionally, we showed that generic datasets are less effective at detecting model-specific hallucinations, and that model-specific datasets enable preemptive detection.

\section{Limitations}\label{sec:limitations}
Our work has a few limitations. 
While we evaluated three popular models, the patterns may differ in other ones. 
Additionally, we used only four settings to induce hallucinations given a model’s correct knowledge; there may be many other ways to achieve similar aims. 
Finally, we only examined the two extremes of the knowledge spectrum and used one way of quantifying knowledge. Future work can investigate this middle region.

\section{Ethics Statement}
Our work provides insights into hallucinations that occur despite the model having the necessary knowledge. While this understanding could be misused to make models less reliable, our goal is to advance the study of the hallucination phenomenon to ultimately improve model reliability.

\bibliography{references,anthology}

\vfill 
\pagebreak 

\appendix

\section{Hyper Parameters Search for Knowledge Categorization}\label{appendix:knowledge_classification}

Knowledge detection typically relies on the model's output, either through logits or generation. We focus on the generation approach, assessing whether the model consistently produces a factually correct answer among multiple samples, similar to recent work \citep{gekhman2024does}. This method is influenced by various hyperparameters including (1) number of generations, (2) sampling temperature, (3) length of generation, and (4) prompt structure.

As directly accessing factually correct is challenging, we instead examined the consistency of knowledge classification across different hyperparameter settings. A high similarity in categorization across settings would suggest comparable proximity to ground truth, reducing the impact of specific hyperparameter choices.

We evaluated the following hyperparameters:
\begin{itemize}
\item Shots: two different 3-shot examples and one zero-shot example.
\item Temperature: $\{0.5,1,1.5\}$
\item Number of generations: $\{5,10\}$
\item Length of generated text: $\{5,10,20\}$ tokens. 
\end{itemize}

We started with a baseline configuration based on preliminary experiments: 3-shots, temperature of $0.5$, 5 generations, and 5 tokens generated. We then modified one parameter at a time to assess its impact on classification similarity.

We categorized knowledge into three classes: ``does not know'' if the model did not generate the correct answer in any of the generations; ``know'' if the model always generated the answer; and ``else'' for anything in between.

We tested this approach on 1000 random TriviaQA examples across our three models. The average similarity among all 8 configurations (28 unique combinations) was $93.6\%$ for Llama, $92.7\%$ for Mistral, and $92.2\%$ for Gemma, indicating a high consistency in knowledge classifications. The lowest similarity (about $80\%$) occurred with zero-shot configurations. Thus, indicating that the variation of configurations should not have significant effect on the knowledge classification.

Based on these results, we adopted the baseline setting as our knowledge detector, using the 3-shot prompt corresponding to the Snowballing used in subsequent hallucination classification. The high similarity between different few-shot prompts suggests that varying the few-shot examples should yield comparable results.
To enhance reliability, we supplemented this approach with one greedy generation, ensuring we capture the most likely output even if temperature-based generations fail to produce it.

Note that most examples are labeled as either ``know'' or ``does not know'' and only about 5\% are labeled as ``else''. Thus we leave the treatment of this category for future work.

One area for improvement in future research is the method of answer detection. While we use `exact match' for simplicity and achieve relatively good results, employing methods that allow for more flexible matching could enhance recall.

\section{Dataset Construction Details}\label{appendix:General Dataset Construction Specifics}
For TriviaQA we took 30K random examples from its training set as our initial dataset, making sure to use only examples where the answer was no longer than 5 tokens using the Mistral tokenizer. In addition, as we saw that some answers were written in uppercase, we also used the lower-case version of these answers if they contained more than 3 letters and did not contain numbers or the `/' symbol.

For the Natural Questions dataset, we also used 30K random examples, excluding examples with answers longer than 5 tokens as well as examples without an answer or with more than one answer. We again added lower-case versions of upper-case answers.

\subsection{Generic Dataset Construction}\label{Appendix:Non-Specific dataset construction}

To create the generic dataset, the key addition was obtaining an incorrect answer for each example. We generated these using Mistral \citep{mistral_7b_paper} with the following prompt:

\begin{center}
\fbox{\parbox{.4\textwidth}{
Question: \{question\}\\
Correct Answer: \{correct answer\}\\
Incorrect Answer:
}}
\end{center}

We accepted the model's greedy generation of 5 tokens as the incorrect answer if it did not contain the correct answer. During this process, we removed words such as `Questions' or `incorrect' that the model occasionally generated alongside the answer.
For examples of hallucinated answers generated by Mistral in the generic dataset of TriviaQA, refer to Table \ref{Generated answers static from triviaqa}.

Note that this process could potentially make the generic dataset more similar to the model-specific Mistral dataset. However, the main paper results show that this way of creating the wrong answers with Mistral provided no advantage for the generic dataset. In fact, this outcome further highlights the benefits of using a model-specific dataset. Despite the generic dataset being generated using responses from Mistral, it still performs worse than the model Mistral-specific dataset.

\begin{table*}

\centering
    \begin{tabular}  
  {p{0.5\linewidth} c c }
\toprule 
\multicolumn{1}{c}{Prompt} &Factually correct & Hallucination\\\midrule
question: Which instrument was primarily played by band leader Count Basie?&Piano&Trumpet\\
answer:  & & \\ 
\midrule
question: Into which body of water does the river Nile empty? & Mediterranean Sea &Atlantic Ocean\\
answer:  & & \\  
\midrule
question: Which planet has a 'great red spot'? &Jupiter&Saturn
  \\
 answer: & & \\  
  \bottomrule
  \end{tabular}
   \caption{Examples from the generic dataset of TriviaQA.}
  \label{Generated answers static from triviaqa}
 
\end{table*}

\subsection{Specific Dataset Construction}\label{Appendix:Specific dataset construction}

To construct a model-specific dataset, we began with all examples from the generic dataset. We then filtered these examples based on the model's knowledge, retaining only those classified as `known' by the model (for detailed information on this classification process,  refer to Appendix \ref{appendix:knowledge_classification}).

Subsequently, we employed a prompt-based approach utilizing Snowballing to generate responses from the model. For each example, the model produced a greedy generation of 5 tokens. We then evaluated these generations to categorize the examples: if the correct answer was present within the generated tokens, we labeled the example as ``factually correct''; otherwise, it was labeled as ``hallucinated''.

This process allowed us to create a dataset that specifically reflects the model's knowledge and generation tendencies, distinguishing between instances where the model correct responses and those where it generates hallucinated content.

We also provide the full prompts we used in each setting.

\textbf{Truthful Prompts}
\begin{enumerate}
    \item Provide responses that are completely accurate and fact-based.
    \item Deliver answers that are both truthful and exact.
    \item Generate replies that are entirely grounded in facts and precision
    \item Offer information that is fully correct and factual.
    \item Ensure all answers are accurate and based solely on verified facts.
    \item 
Craft responses that are entirely reliable and factually correct.
\item Compose answers with complete factual accuracy and precision.
\item Create responses that are truthful and meticulously accurate.
\item Formulate replies that are based solely on factual correctness and clarity.
\item Produce content that is both accurate and strictly fact-driven.
\end{enumerate}

\textbf{Persona Prompts}
\begin{enumerate}
    \item Hey, um, I know these questions are really tough, but they’re so important to me. Could you please take the time to explain them? I’m really relying on your knowledge and kindness here.
    \item Um... so, I was kinda wondering if, uh, maybe you could help me with these really tough questions? I mean, if it’s okay... I’d really like to understand them better.
    \item Hey! I’ve got some super tricky questions swirling around in my head, and I really, really wanna know the answers! You’re so good at explaining stuff—can you please help me figure them out? Pretty please?
    \item Well, I reckon I’d sure like to get the answers to these tough questions. Would you mind explaining ‘em to me? I’d sure appreciate it.
    \item Crikey, these are some tough nuts to crack, but I'd really love to get me head around the answers to these tricky questions. Would you mind explaining them to me? Much appreciated, mate!
    \item As someone who loves nature and has a deep curiosity for the world around us, I find myself pondering some truly challenging questions. I'd really appreciate it if you could help explain them to me!
    \item I crave to uncover the dark truths behind these devilishly difficult questions that haunt my mind. Would you dare to shed some light on them for me?
    \item Oh, my dear, I’m just an old granny who gets all jittery with anything too spooky or difficult, but I truly wish to understand these puzzling questions that seem so very hard. Could you kindly take a moment to explain them to me? I’d be ever so grateful, sweetie.
    \item Alright, folks, gather 'round! I’ve got a barrel of questions here that are giving me a real stir. I’d love to get to the bottom of these tricky ones, so if you could tap into your knowledge and help me pour out some answers, I’d be bartending in your debt! What do you say, can you mix me up some explanations?
    \item I find myself lost in these questions, so deep and elusive. Could you shed some light and help me find the answers I seek?
\end{enumerate}

Lastly, we provide the full list of the 20 bad shots and good shots we used:

\textbf{Good shots:}

question: What is the capital of France?\\answer: Paris\\question: How many continents are there?\\answer: 7\\question: Who wrote 'Romeo and Juliet'?\\answer: William Shakespeare\\question: What is the square root of 64?\\answer: 8\\question: Which element has the chemical symbol 'H'?\\answer: Hydrogen\\question: Who was the first President of the United States?\\answer: George Washington\\question: What is the powerhouse of the cell?\\answer: Mitochondria\\question: In what year did World War II end?\\answer: 1945\\question: What is the currency of Japan?\\answer: Japanese Yen\\question: Who painted the Mona Lisa?\\answer: Leonardo da Vinci\\question: What is the speed of light?\\answer: 299,792 kilometers per second\\question: How many sides does a hexagon have?\\answer: 6\\question: What is the boiling point of water in Celsius?\\answer: 100 degrees\\question: Who wrote 'To Kill a Mockingbird'?\\answer: Harper Lee\\question: What is the capital of Australia?\\answer: Canberra\\question: What is the largest ocean on Earth?\\answer: Pacific Ocean\\question: Who discovered penicillin?\\answer: Alexander Fleming\\question: What is the chemical symbol for gold?\\answer: Au\\question: What is the smallest prime number?\\answer: 2\\question: How many planets are there in our solar system?\\answer: 8\\

\textbf{Bad shots:}

question: What is the capital of France?\\answer: Berlin\\question: How many continents are there?\\answer: 6\\question: Who wrote 'Romeo and Juliet'?\\answer: Jane Austen\\question: What is the square root of 64?\\answer: 7\\question: Which element has the chemical symbol 'H'?\\answer: Helium\\question: Who was the first President of the United States?\\answer: Abraham Lincoln\\question: What is the powerhouse of the cell?\\answer: Golgi Apparatus\\question: In what year did World War II end?\\answer: 1939\\question: What is the currency of Japan?\\answer: Euro\\question: Who painted the Mona Lisa?\\answer: Pablo Picasso\\question: What is the speed of light?\\answer: 300,000 kilometers per second\\question: How many sides does a hexagon have?\\answer: 5\\question: What is the boiling point of water in Celsius?\\answer: 50 degrees\\question: Who wrote 'To Kill a Mockingbird'?\\answer: J.K. Rowling\\question: What is the capital of Australia?\\answer: Sydney\\question: What is the largest ocean on Earth?\\answer: Atlantic Ocean\\question: Who discovered penicillin?\\answer: Isaac Newton\\question: What is the chemical symbol for gold?\\answer: Ag\\question: What is the smallest prime number?\\answer: 1\\question: How many planets are there in our solar system?\\answer: 9\\

\subsection{Specific Dataset Qualitative Evaluation}\label{appendix:qualitative evaluation}

To illustrate the examples generated by our Snowballing setting and to highlight the differences between factually-correct, \wak, and \lok responses, we present sample generations with and without the Snowballing setting in Tables \ref{Generated answers using bad/non shots in the prompt gemma}, \ref{Generated answers using bad/non shots in the prompt Llama}, and \ref{Generated answers using bad/non shots in the prompt mistral}.
We observe that for \wak examples, the model generates a hallucination only in the Snowballing setting, while \lok examples produce hallucinations even without it. Additionally, factually-correct examples continue to generate the correct answer even in the Snowballing setting.

\begin{table*}[t]

\centering
% \small
  \begin{tabular}  
  {P{0.07\linewidth} P{0.20\linewidth}c c c c}
\toprule 
 & & & \multicolumn{2}{c}{Generation} \\ 
\cmidrule(lr){4-5}
Type&Prompt &Golden&w/ without addition & w/ Snowballing\\\midrule
 \wak &In the Old Testament, who was the mother of Solomon? & Bathsheba & Bathsheba & Mary\\\midrule
factually-correct & In humans, citguatera is an illness caused by eating contaminated what?& Fish & Fish & Fish\\\midrule
\lok & Who is the longest-serving captain in the history of Manchester United FC?&Bryan Robson &Roy Keane&Roy Keane\\

  \\\bottomrule

  \end{tabular}
  \caption{Generated answers using greedy decoding on TriviaQA with Gemma.}
  \label{Generated answers using bad/non shots in the prompt gemma}
  
\end{table*}

\begin{table*}[t]

\centering
% \small
  \begin{tabular}  
  {P{0.07\linewidth} P{0.20\linewidth}c c c c}
\toprule 
 & & & \multicolumn{2}{c}{Generation} \\ 
\cmidrule(lr){4-5}
Type&Prompt &Golden&w/ without addition & w/ Snowballing\\\midrule
 \wak &Who in 1990 became the first chancellor of a united Germany&Helmut Kohl&Helmut Kohl&John F. Kennedy\\\midrule
factually-correct & In the human body, what is the more common name for the 'Zygomatic Bone'?&Cheekbone&Cheekbone&Cheekbone\\\midrule
\lok &In the song The Twelve Days of Christmas, how many pipers are there?&Eleven&12 pipers&12 \\

  \\\bottomrule

  \end{tabular}
  \caption{Generated answers using greedy decoding on TriviaQA with Llama.}
  \label{Generated answers using bad/non shots in the prompt Llama}

\end{table*}

\begin{table*}[t]

\centering
% \small
  \begin{tabular}  
  {P{0.07\linewidth} P{0.20\linewidth}c c c c}
\toprule 
 & & & \multicolumn{2}{c}{Generation} \\ 
\cmidrule(lr){4-5}
Type&Prompt &Golden&w/ without addition & w/ Snowballing\\\midrule
 \wak & What color are the stars on an official United States flag?&White&White&Red\\\midrule
factually-correct &  A seriema is what type of creature? &Bird & A bird&A bird\\\midrule
\lok &Which insect does Isle of Wight Disease affect?&BEE& The Whitefly&Mosquito\\

  \\\bottomrule

  \end{tabular}
   \caption{Generated answers using greedy decoding on TriviaQA with Mistral.}
  \label{Generated answers using bad/non shots in the prompt mistral}
 
\end{table*}

\subsection{Dataset Statistics}\label{appendix:Alice-setting dataset statistics}
In this section, we show the dataset statistics for all the different settings. 
The results are in Tables \ref{Dataset_Statistic_snoaballing}, \ref{Dataset_Statistic_alice}, \ref{Dataset_Statistic_persona} and \ref{Dataset_Statistic_truthful}. We can see that the Snowballing setting contains the highest number of \wak examples, while the milder settings yield progressively fewer examples.

\begin{table*}[t!]
 
\centering
  \begin{tabular}
  {l r c r }
  \toprule

     & \multicolumn{1}{c}{\# Factually}  & \multicolumn{1}{c}{\# Hallucination}  & \multicolumn{1}{c}{\# Do-not-know} \\
   Dataset & \multicolumn{1}{c}{correct} & \multicolumn{1}{c}{(\wak)} & \multicolumn{1}{c}{(\lok)} \\ 
   \midrule
TriviaQA-Llama3-\bn&14154&1675&7356\\
Natural-Questions-Llama3-\bn&5934&1104&14739\\
TriviaQA-Gemma-\bn&13534&2563&6991\\
Natural-Questions-Gemma-\bn&6045&1859&13762\\
TriviaQA-Mistral-\bn&12652&2841&7650\\
Natural-Questions-Msitral-\bn&5562&1546&14689\\
\bottomrule
  \end{tabular}
   \caption{Dataset labels statistics using Snowballing setting.}
  \label{Dataset_Statistic_snoaballing}
\end{table*}

\begin{table*}[t!]
 
\centering

  \begin{tabular}
{l r c r }
  \toprule

     & \multicolumn{1}{c}{\# Factually}  & \multicolumn{1}{c}{\# Hallucination}  & \multicolumn{1}{c}{\# Do-not-know} \\
   Dataset & \multicolumn{1}{c}{correct} & \multicolumn{1}{c}{(\wak)} & \multicolumn{1}{c}{(\lok)} \\ 
   \midrule
TriviaQA-Llama3-\bn&14851&978&7356\\
Natural-Questions-Llama3-\bn&6059&979&14739\\
TriviaQA-Gemma-\bn&15418&679&6991\\
Natural-Questions-Gemma-\bn&7194&710&13762\\
TriviaQA-Mistral-\bn&14505&988&7650\\
Natural-Questions-Msitral-\bn&6232&876&14689\\

\bottomrule

  \end{tabular}
   \caption{Dataset label statistics on the Alice-Bob setting.}
  \label{Dataset_Statistic_alice}
\end{table*}

\begin{table*}[t!]
 
\centering

  \begin{tabular}
{l r c r }
  \toprule

     & \multicolumn{1}{c}{\# Factually}  & \multicolumn{1}{c}{\# Hallucination}  & \multicolumn{1}{c}{\# Do-not-know} \\
   Dataset & \multicolumn{1}{c}{correct} & \multicolumn{1}{c}{(\wak)} & \multicolumn{1}{c}{(\lok)} \\ 
   \midrule
TriviaQA-Llama3-\bn&15062&767&7356\\
Natural-Questions-Llama3-\bn&6153&885&14739\\
TriviaQA-Gemma-\bn&15445&652&6991\\
Natural-Questions-Gemma-\bn&7195&709&13762\\
TriviaQA-Mistral-\bn&14575&918&7650\\
Natural-Questions-Msitral-\bn&6218&890&14689\\

\bottomrule

  \end{tabular}
   \caption{Dataset label statistics on the Persona setting.}
  \label{Dataset_Statistic_persona}
\end{table*}

\begin{table*}[t!]
 
\centering

  \begin{tabular}
{l r c r }
  \toprule

     & \multicolumn{1}{c}{\# Factually}  & \multicolumn{1}{c}{\# Hallucination}  & \multicolumn{1}{c}{\# Do-not-know} \\
   Dataset & \multicolumn{1}{c}{correct} & \multicolumn{1}{c}{(\wak)} & \multicolumn{1}{c}{(\lok)} \\ 
   \midrule
TriviaQA-Llama3-\bn&14987&842&7356\\
Natural-Questions-Llama3-\bn&6203&835&14739\\
TriviaQA-Gemma-\bn&15458&639&6991\\
Natural-Questions-Gemma-\bn&7253&651&13762\\
TriviaQA-Mistral-\bn&14728&765&7650\\
Natural-Questions-Msitral-\bn&6421&687&14689\\

\bottomrule

  \end{tabular}
   \caption{Dataset label statistics on the Truthful setting.}
  \label{Dataset_Statistic_truthful}
\end{table*}

\section{Implementation Details}\label{sec:Implementation Details}

We aim to show the importance of separating the two hallucination types and using our model's specific dataset to create better detectors. 
In the following sections, we report on various experiments for detecting different types of hallucinations by training classifiers on inner model states. 
Each experiment was repeated with three random seeds (42,100,200) for the SVM and split into training/test sets. We report average results with standard deviations. To maintain consistency with the prompts used in the creation of the WACK dataset, all examples incorporate similar prompts. In addition, unless stated otherwise the results are shown under the Snowballing setting and the dashed black line is the baseline. We used Sklearn \citep{pedregosa2011scikit} LinearSVC with 1000000 maximum iterations and 1e-5 as the tolerance for stopping criteria.
Lastly, unless stated otherwise, given an example we detect at the answer ($\tilde{a}$) last token, which may or may not be a hallucination.

All experiments were run on NVIDIA RTX 6000 Ada (49GB) with 4 CPUs. 
Generating all the datasets and results takes approximately 2-4 weeks on one GPU.

We use the datasets under the Apache License and the models under each one's agreement terms to follow each artifact's terms.

Lastly, We used AI assistants only for simple paraphrasing as part of writing this paper.

\section{Mitigation Additional Results}\label{appendix:Mitigation Additional results}
To show the broadness of the mitigation results shown in Section \ref{Mitigation fails for lok hallucinations},  we present results using an alternative mitigation prompt: ``Produce responses that are completely factual and accurate, irrespective of any problems in the text.'' instead of ``Generate answers that are entirely factual and precise, regardless of any issues in the text.''. The results, shown in Table \ref{mitigation results additional prompt}, are consistent with those presented in the main paper. Once again, we observe that mitigation is significantly more effective for \wak examples than for \lok examples, reinforcing the importance of distinguishing between these two types.

\begin{table*}[t]
\small

\centering
\begin{tabular}{cccc}
\toprule
Model & Data Set & \wak & \lok \\ \hline
\multirow{2}{*}{Gemma} & TriviaQA &\textbf{15.4 / 20.0 / 19.6 / 7.8} & 2.2 / 1.6 / 1.8 / 2.2 \\ 
 & Natural Questions &\textbf{17.2 / 23.2 / 19.0 / 11.6}&2.2 / 1.0 / 1.6 / 1.8\\ \midrule
\multirow{2}{*}{Llama} & TriviaQA & \textbf{18.6 / 26.2 / 16.4 / 11.6}& 1.4 / 1.6 / 0.6 / 2.2\\ 
 & Natural Questions &\textbf{16.8 / 25.4 / 20.2 / 18.4}& 1.0 / 1.4 / 0.6 / 1.0\\ \midrule
\multirow{2}{*}{Mistral} & TriviaQA &\textbf{24.6 / 17.4 / 18.6 / 18.2} & 1.8 / 3.2 / 1.0 / 3.4\\ 
 & Natural Questions & \textbf{23.6 / 20.0 / 15.4 / 16.4}& 1.4 / 2.2 / 0.8 / 1.4\\ \bottomrule
\end{tabular}
\caption{Comparison of \wak and \lok mitigation results using alternative mitigation prompt on Truthful / Persona / Alice-Bob / Snowballing settings. We can see that \wak mitigation is significantly higher than \lok.}
\label{mitigation results additional prompt}

\end{table*}

\section{Distinguishing HK$^+$ from HK$^-$}\label{appendix distinguish}
\begin{figure*}
    \centering
 \begin{subfigure}[b]{0.32\textwidth}
  \centering
\includegraphics[width=0.9\linewidth]{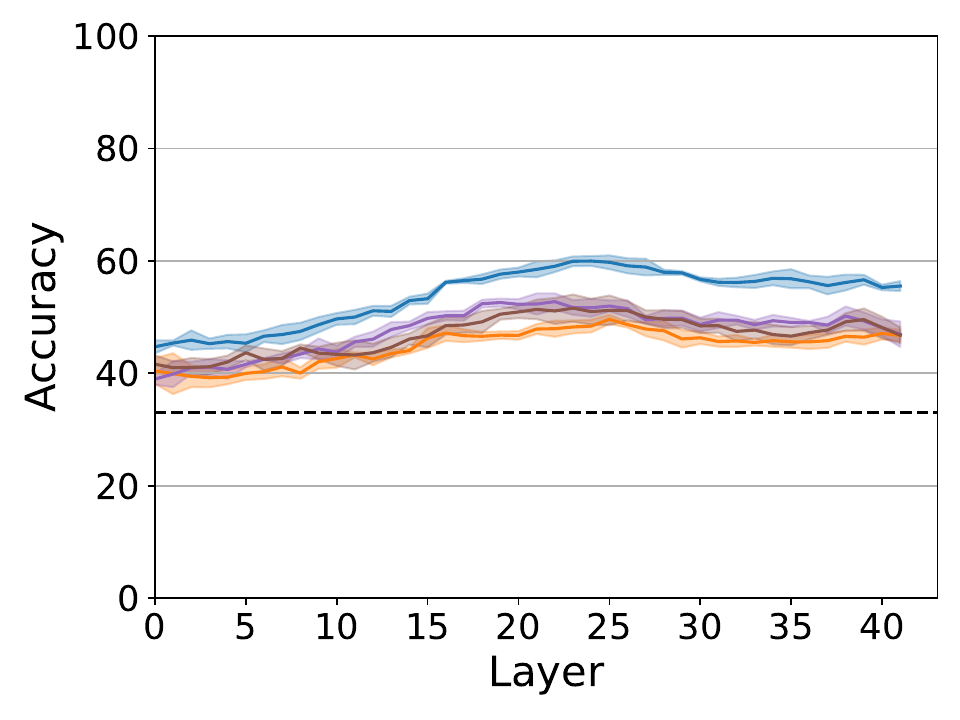}
  \caption{Gemma.}
 \end{subfigure}
 \begin{subfigure}[b]{0.32\textwidth}
  \centering
\includegraphics[width=0.9\linewidth]{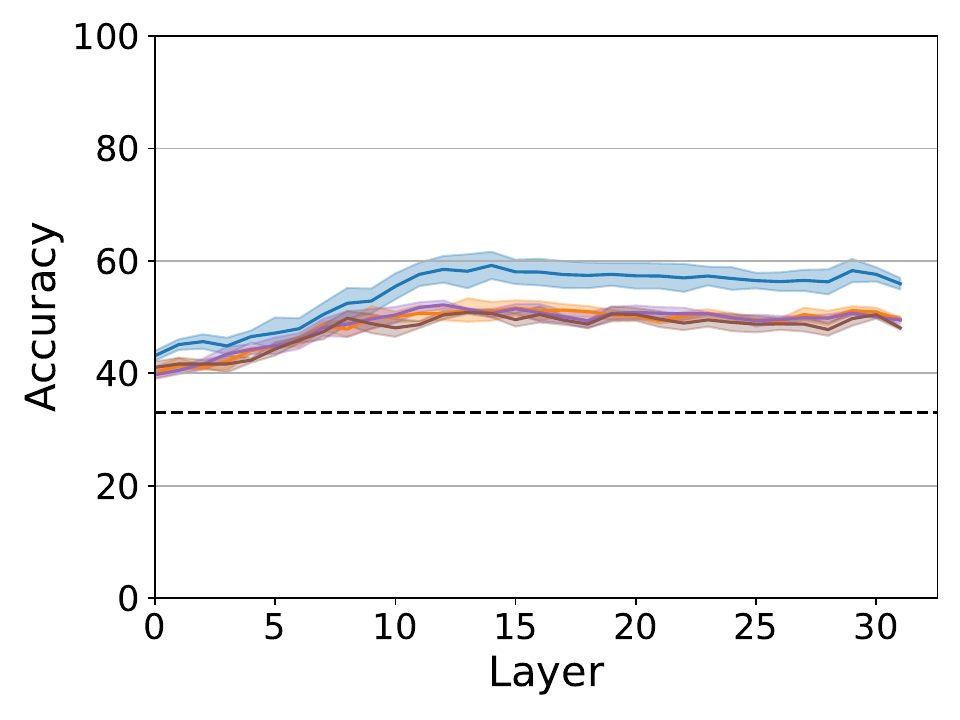}
  \caption{Llama.}
 \end{subfigure}
 \begin{subfigure}[b]{0.32\textwidth}
  \centering
\includegraphics[width=0.9\linewidth]{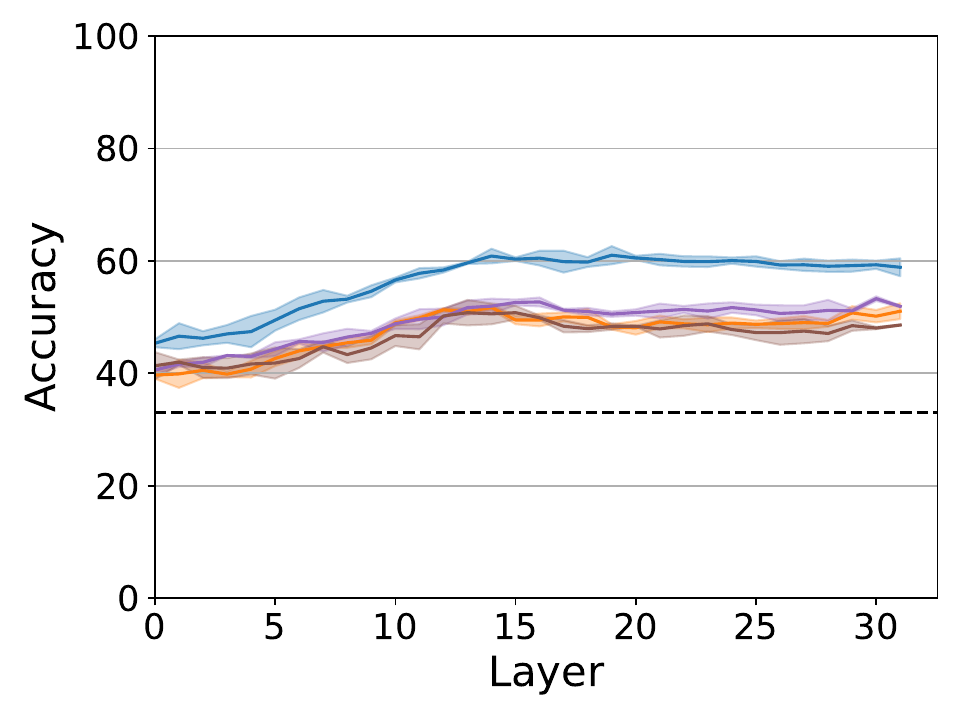}
  \caption{Mistral.}
 \end{subfigure}
 \caption{3-way classification results on Natural Questions into (i) hallucinations caused by lack of knowledge (\lok), (ii)  hallucinations caused despite having knowledge (\wak), and (iii)  factually correct examples. All the detection results are well above the random baseline (dashed line) indicating the possibility of distinguishing \wak from \lok.}
 \label{fig:Hallucinations from miss knowledge vs. hallucinations regardless of knowledge vs. non-hallucination-knowledge classification naturalqa}
\end{figure*}

\begin{figure*}
\centering
 \centering
\begin{subfigure}[b]{0.32\textwidth}
  \centering
  \includegraphics[width=0.9\linewidth]{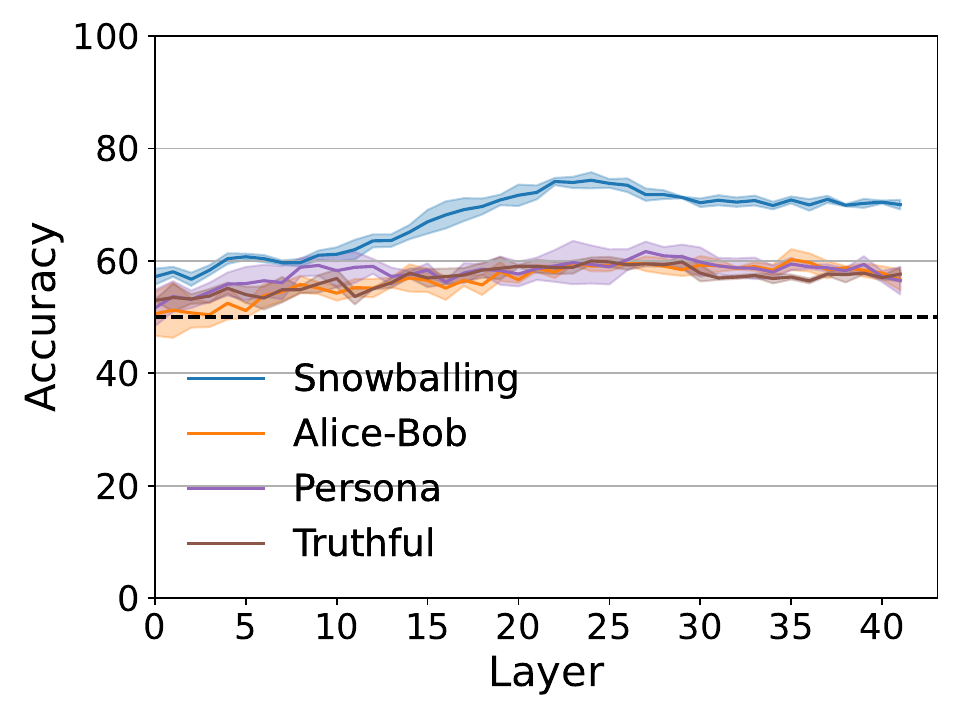}
  \caption{Gemma, TriviaQA.}
 \end{subfigure}
 \begin{subfigure}[b]{0.32\textwidth}
  \centering
\includegraphics[width=0.9\linewidth]{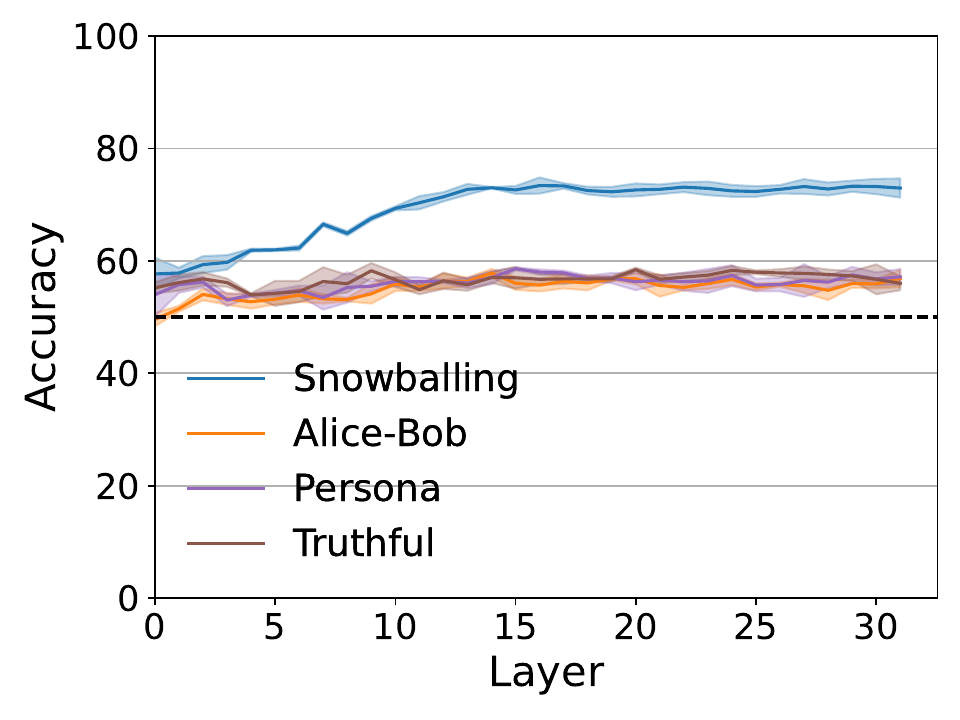}
  \caption{Llama, TriviaQA.}
 \end{subfigure}
 \begin{subfigure}[b]{0.32\textwidth}
  \centering
  \includegraphics[width=0.9\linewidth]{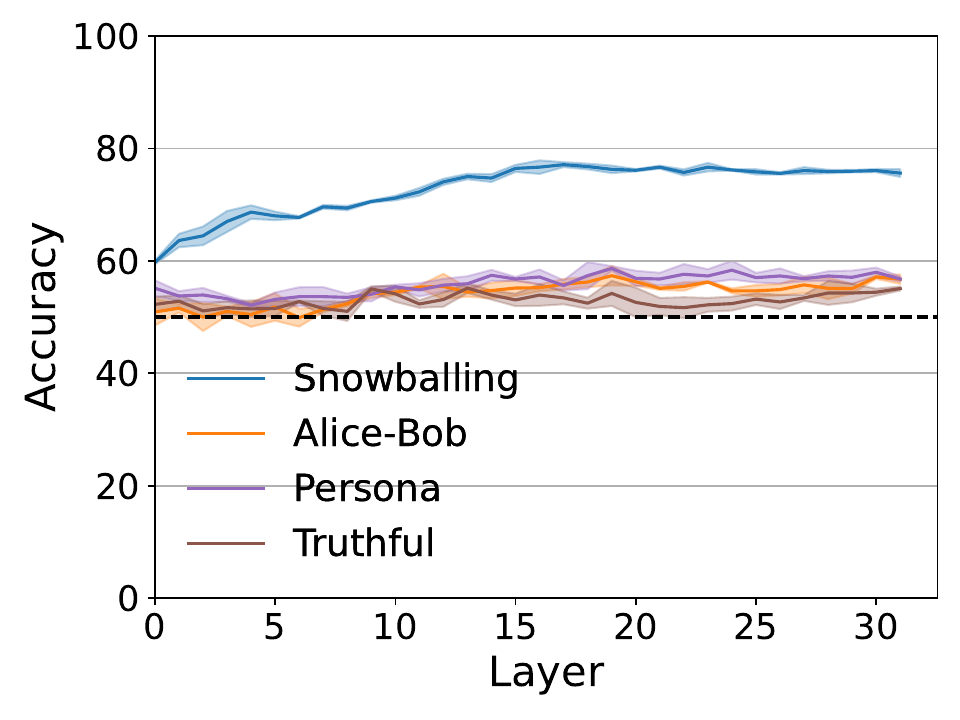}
  \caption{Mistral, TriviaQA.}
 \end{subfigure}\\
 \begin{subfigure}[b]{0.32\textwidth}
  \centering
\includegraphics[width=0.9\linewidth]{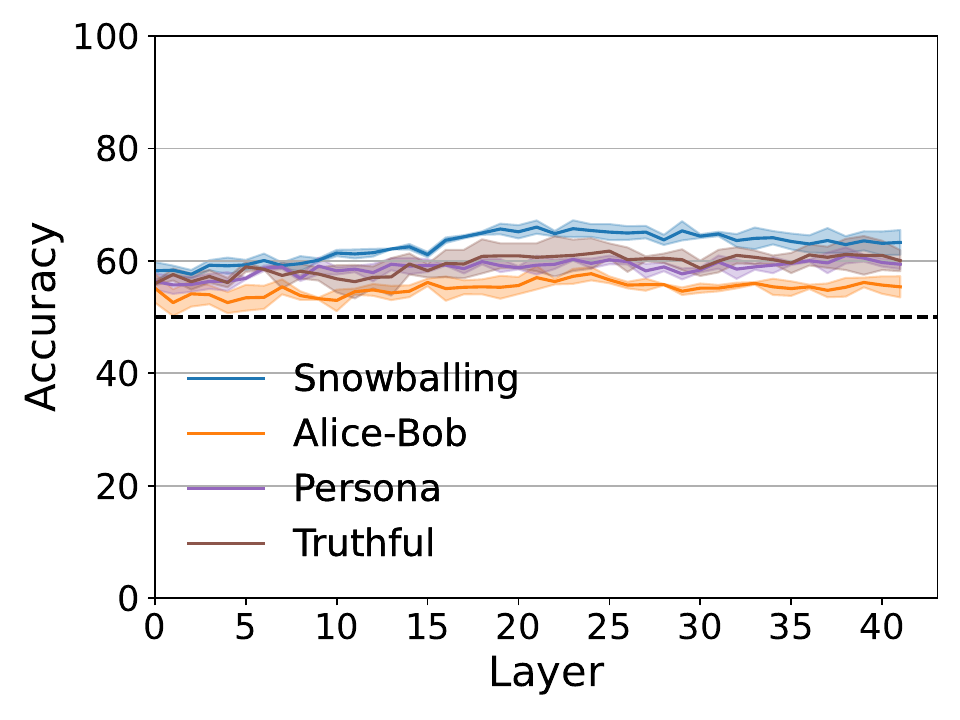}
  \caption{Gemma, Natural Questions.}
 \end{subfigure}
 \begin{subfigure}[b]{0.32\textwidth}
  \centering
\includegraphics[width=0.9\linewidth]{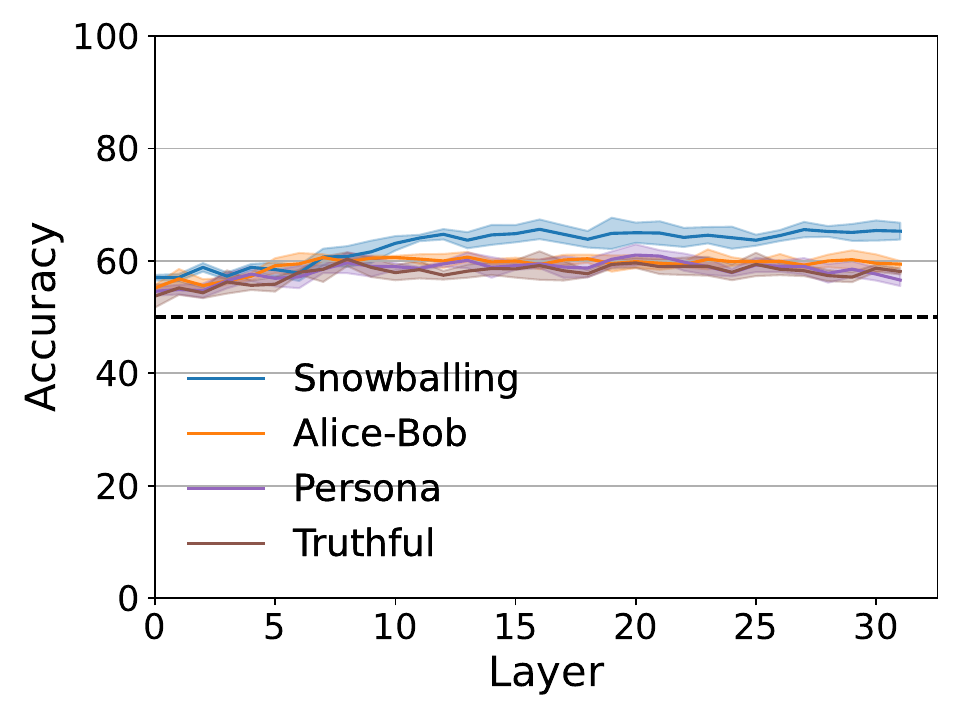}
  \caption{Llama, Natural Questions.}
 \end{subfigure}
 \begin{subfigure}[b]{0.32\textwidth}
  \centering
\includegraphics[width=0.9\linewidth]{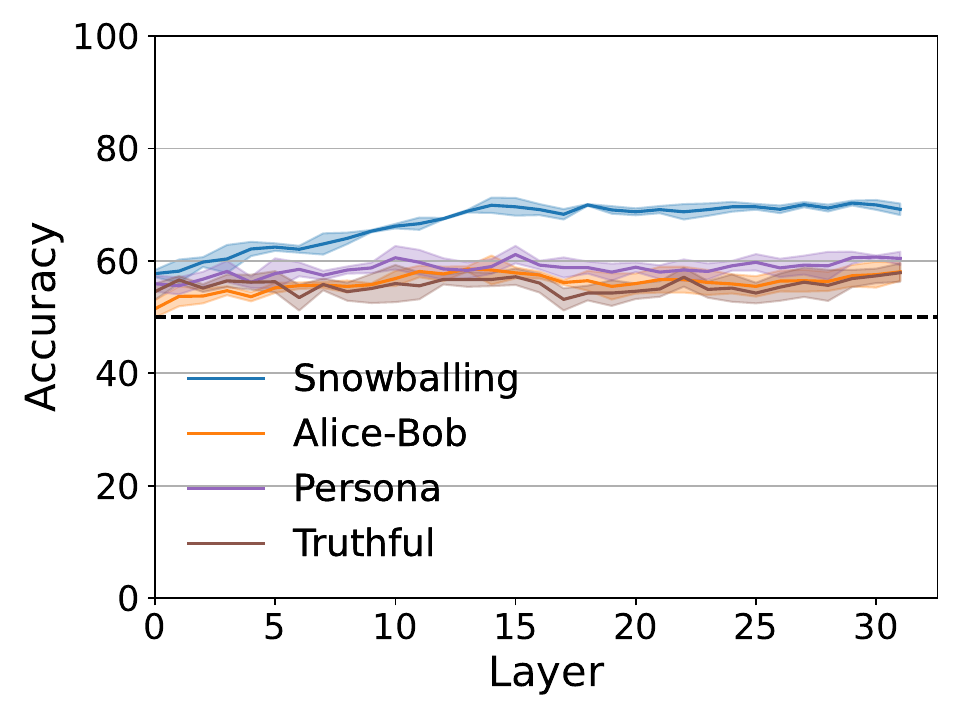}
  \caption{Mistral, Natural Questions.}
 \end{subfigure}
 \caption{Classification results into (i) hallucinations caused by lack of knowledge (\lok), (ii)  hallucinations caused despite having knowledge (\wak). We can see that all the detection lines are well above the random baseline (dashed line) indicating the possibility of distinguishing \wak from \lok.}
 \label{fig:Hallucinations from miss knowledge vs. hallucinations regardless of knowledge classification}
\end{figure*}

 In Section \ref{sec_detect_halu_types}, we showed detection results on TriviaQA of distinguishing \wak from \lok from factually-correct. To complement this,  Figure \ref{fig:Hallucinations from miss knowledge vs. hallucinations regardless of knowledge vs. non-hallucination-knowledge classification naturalqa} shows similar results on Natural Questions dataset and
 Figure \ref{fig:Hallucinations from miss knowledge vs. hallucinations regardless of knowledge classification} shows classification results when distinguishing between the two hallucination types,  \wak and \lok.
 Distinguishing these two types is easier under the snowballing setting (sometimes getting close to 80\%), but even under other settings, results are above chance.  

\section{Additional Results on the Difference between Models Hallucinations}\label{appendix:Additional Results on the Difference between Models Hallucinations}

In Section \ref{sec:halu-similarity-across-models}, we demonstrate that different models possess distinct knowledge and exhibit unique types of hallucinations. Additionally, even within the same model, the hallucinations can vary depending on the specific settings used (e.g., Alice-Bob, Truthful, etc.).

\begin{figure*}[t]
\centering
% \begin{figure*}
 \centering
\begin{subfigure}[b]{0.2\textwidth}
  \centering
  \includegraphics[width=\linewidth]{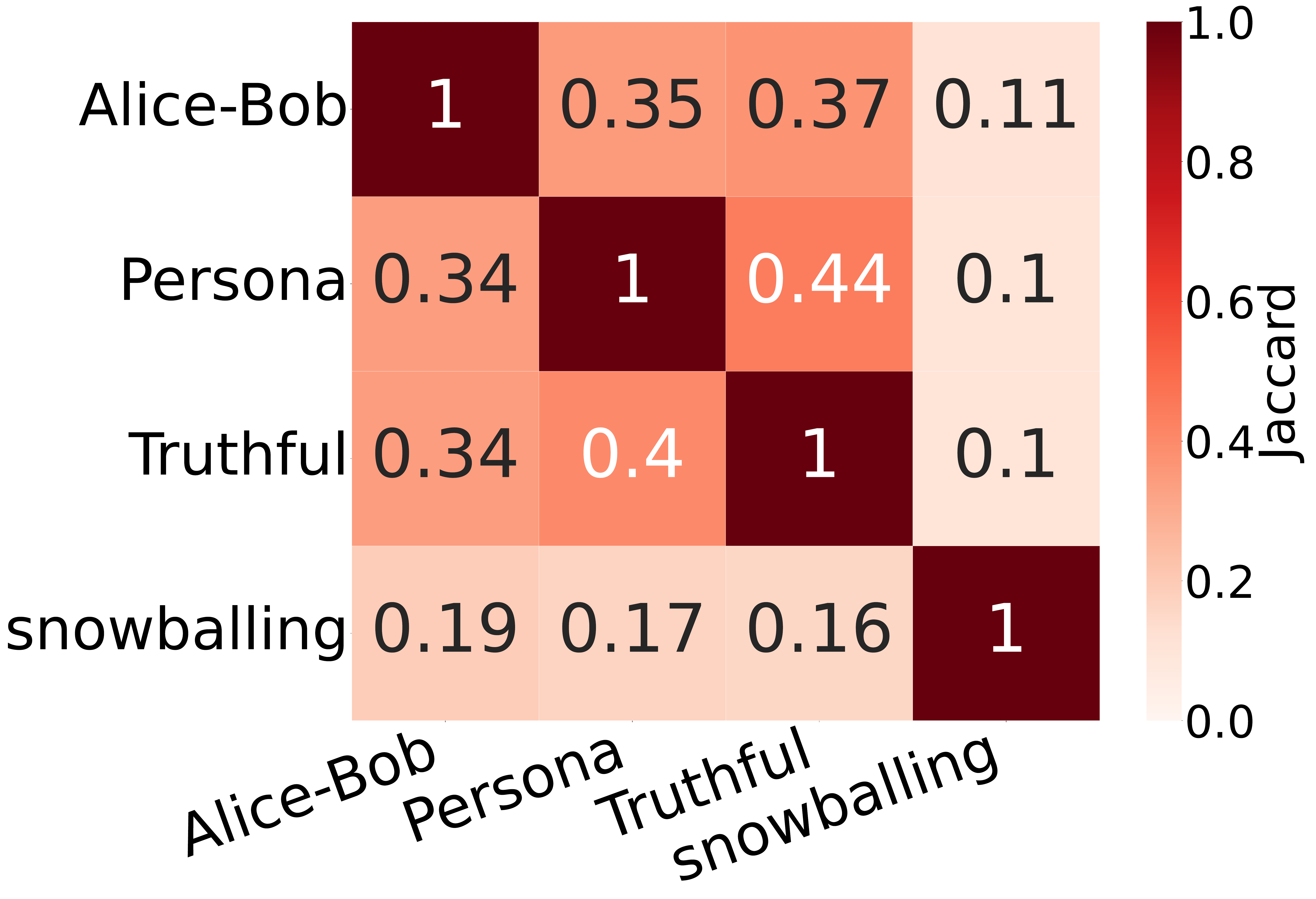}
  \caption{Gemma \wak similarity between settings.}
  \label{g}
 \end{subfigure}%
 \hfill
 \begin{subfigure}[b]{0.2\textwidth}
  \centering
  \includegraphics[width=\linewidth]{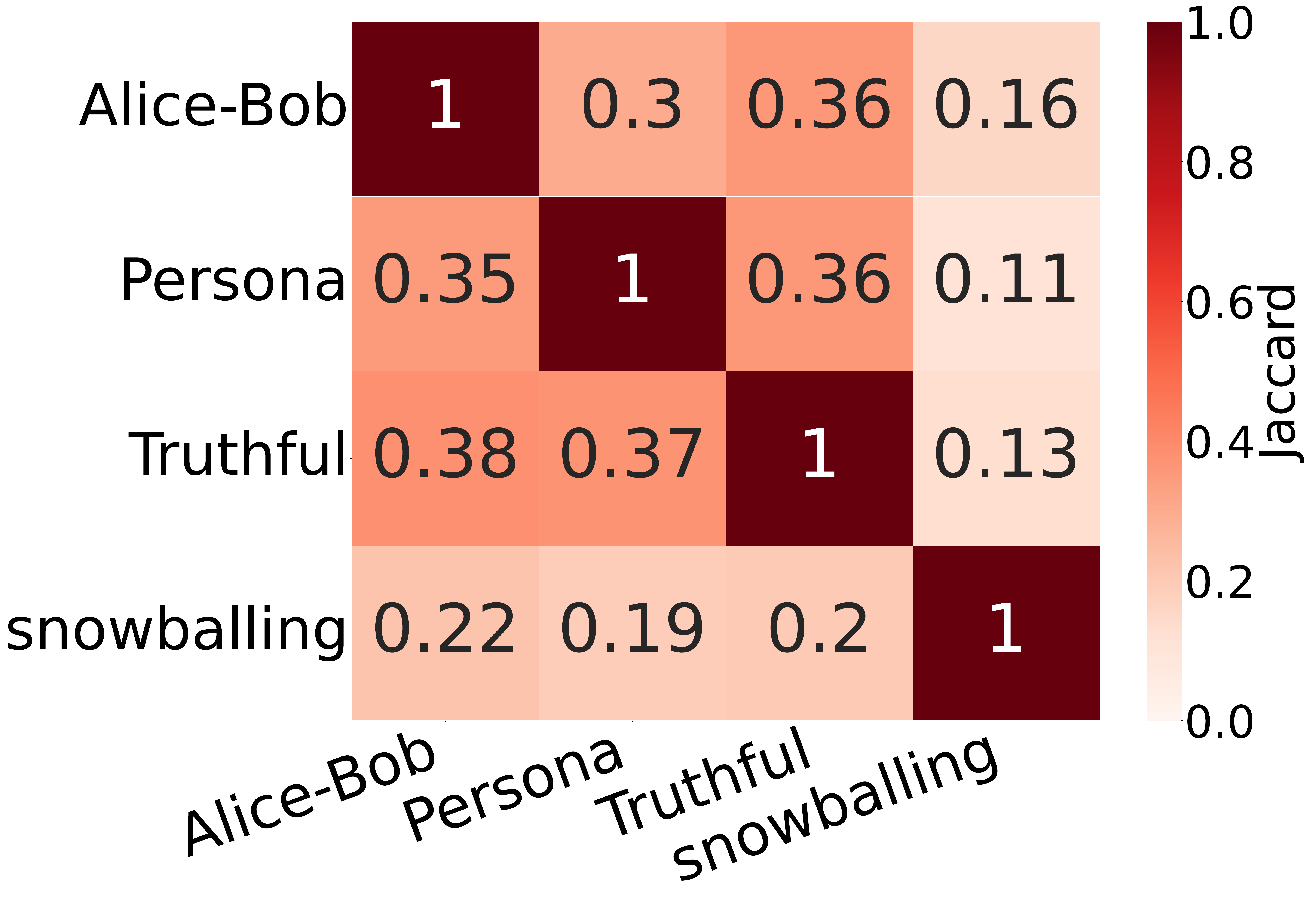}
  \caption{Llama \wak similarity between settings.}
  \label{l}
 \end{subfigure}
 \hfill
  \begin{subfigure}[b]{0.2\textwidth}
  \centering
  \includegraphics[width=\linewidth]{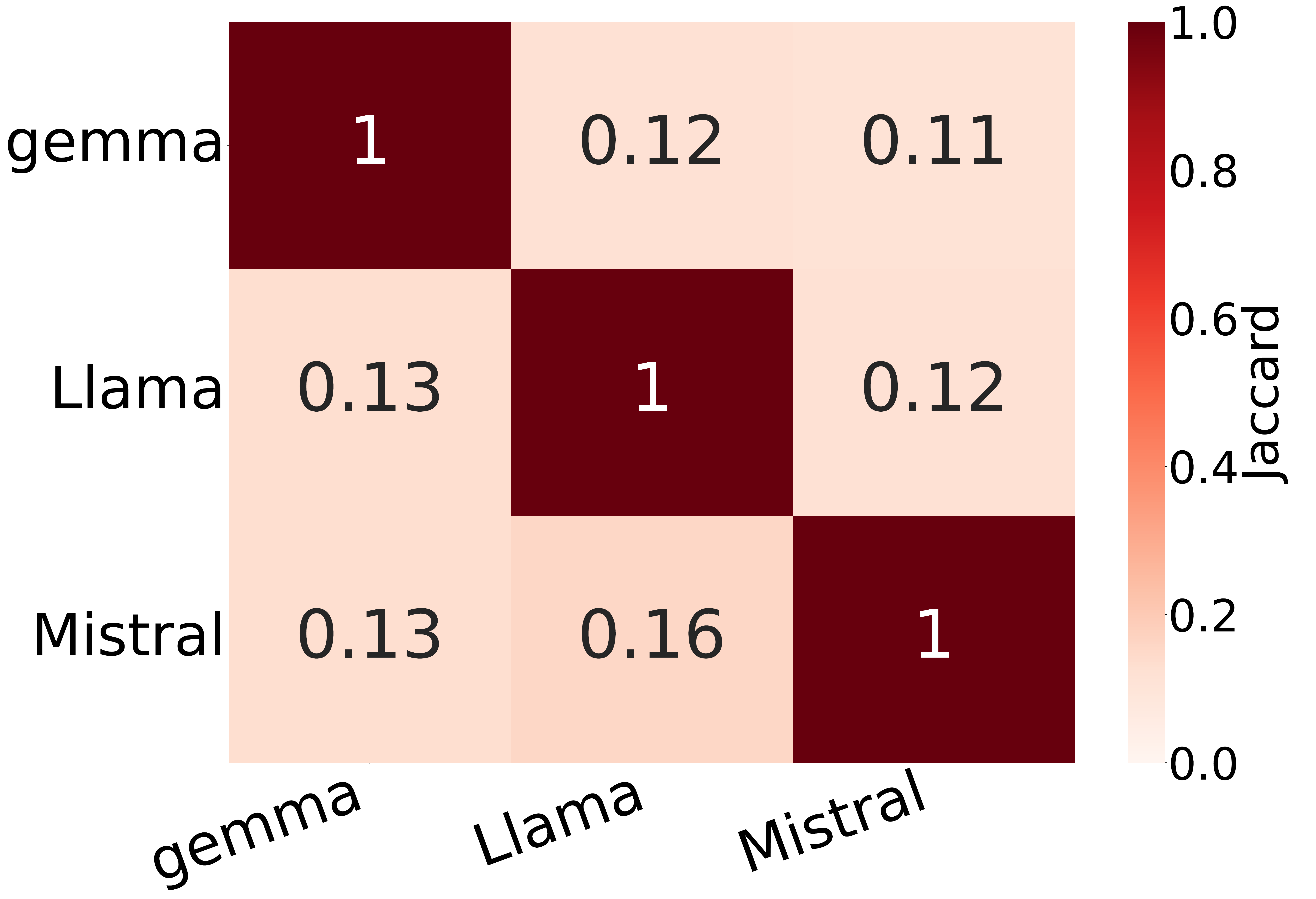}
  \caption{\wak similarity under Alice-Bob.}
  \label{hallucination similarity_alice}
 \end{subfigure}
\hfill
  \begin{subfigure}[b]{0.2\textwidth}
  \centering
  \includegraphics[width=\linewidth]{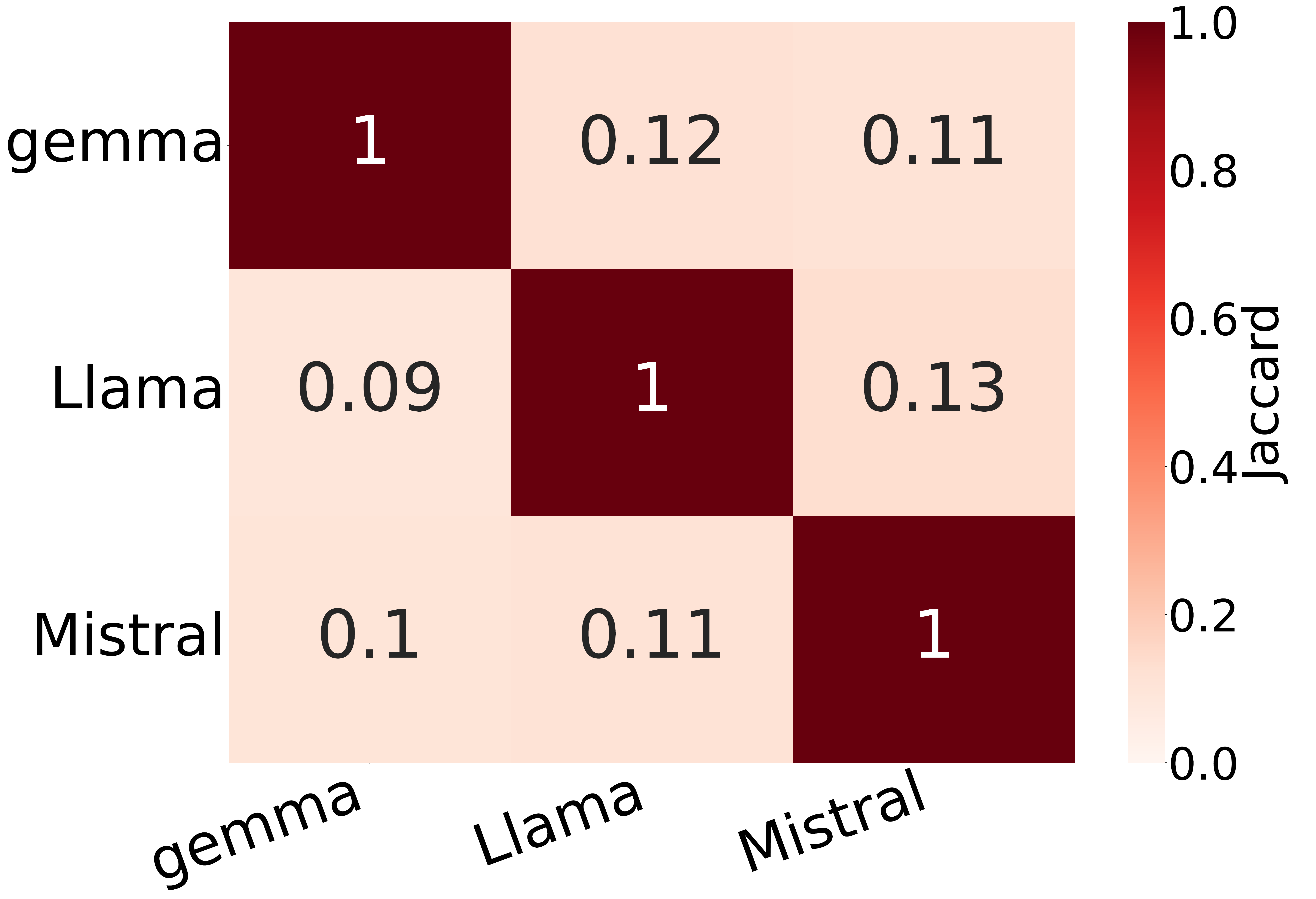}
  \caption{\wak similarity under Truthful setting.}
  \label{truthful}
 \end{subfigure}

 \caption{High-Knowledge and \wak differences on TriviaQA (above the diagonal) and Natural Questions (below the diagonal) between the models and settings. We can see that different models have different knowledge and different \wak examples in shared knowledge. }

 \label{fig:knowledge and hallucination differences appendix}
\end{figure*}

Figure \ref{fig:knowledge and hallucination differences appendix} presents additional results for configurations beyond those discussed in the main paper. Specifically, Figures \ref{g} and \ref{l} illustrate the similarity of settings under the Gemma and Metta models, while Figures \ref{hallucination similarity_alice} and \ref{truthful} highlight the hallucination similarity across models using Alice-Bob and Truthful settings. These findings mirror the patterns observed in the main paper and further underscore the differences between models and settings. This emphasizes the necessity of developing model- and setting-specific datasets.

\section{\wak Detection Additional Results} \label{appendix:wak Detection Additional Results}
\begin{figure*}
    \centering
    
\begin{subfigure}[b]{0.33\textwidth}
  \centering
  \includegraphics[width=0.9\linewidth]{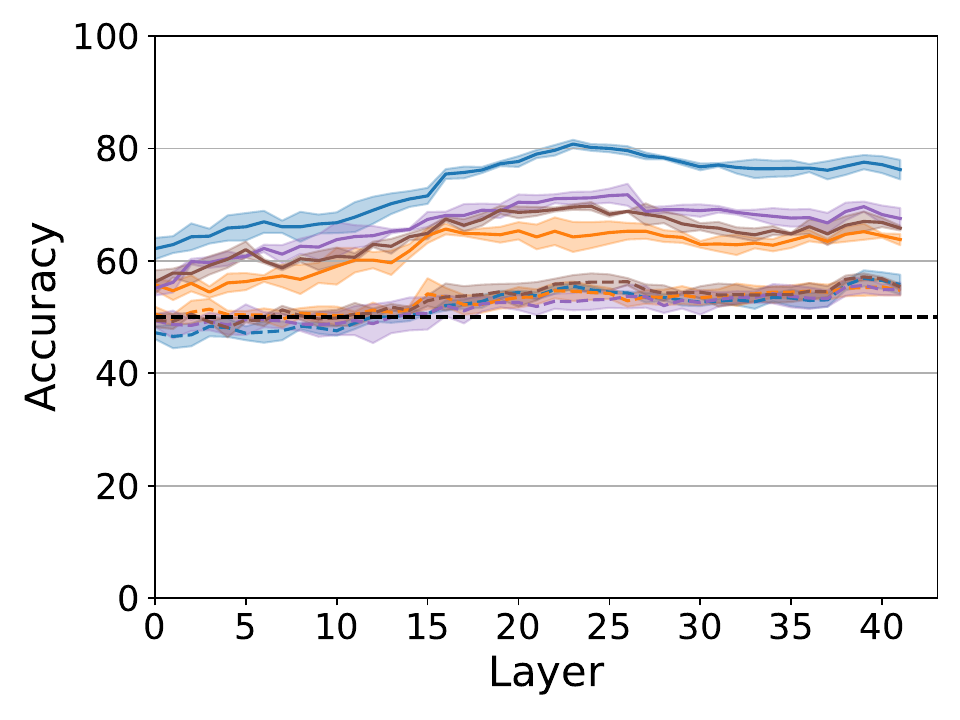}
  \caption{Gemma.}
 \end{subfigure}%
 \hfill
 \begin{subfigure}[b]{0.33\textwidth}
  \centering
  \includegraphics[width=0.9\linewidth]{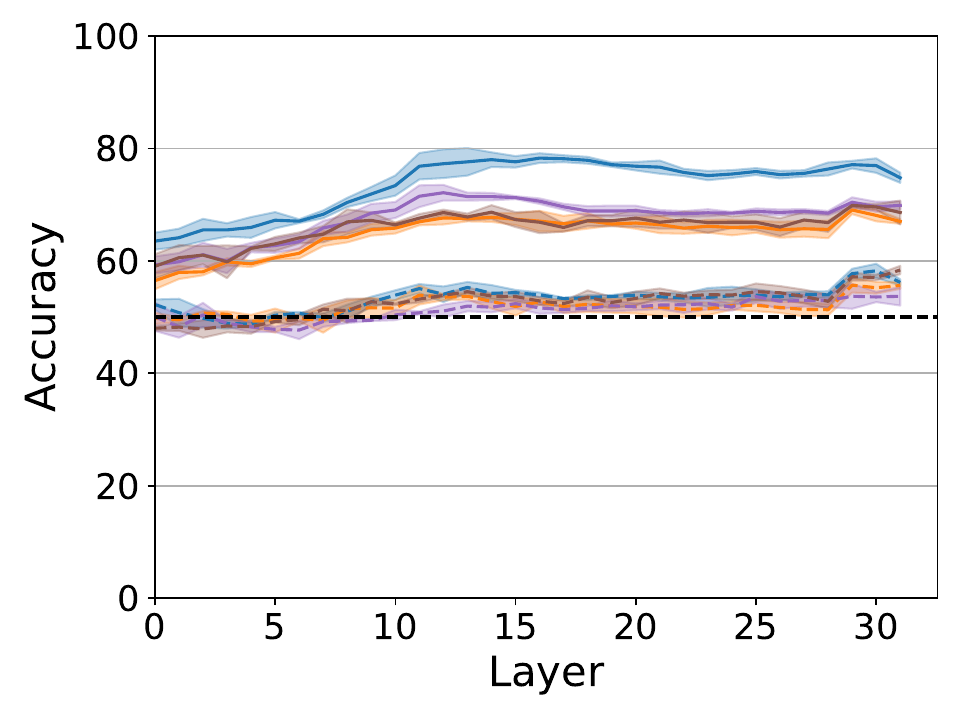}
  \caption{Llama.}   \end{subfigure}
  \hfill
 \begin{subfigure}[b]{0.33\textwidth}
  \centering
  \includegraphics[width=0.9\linewidth]{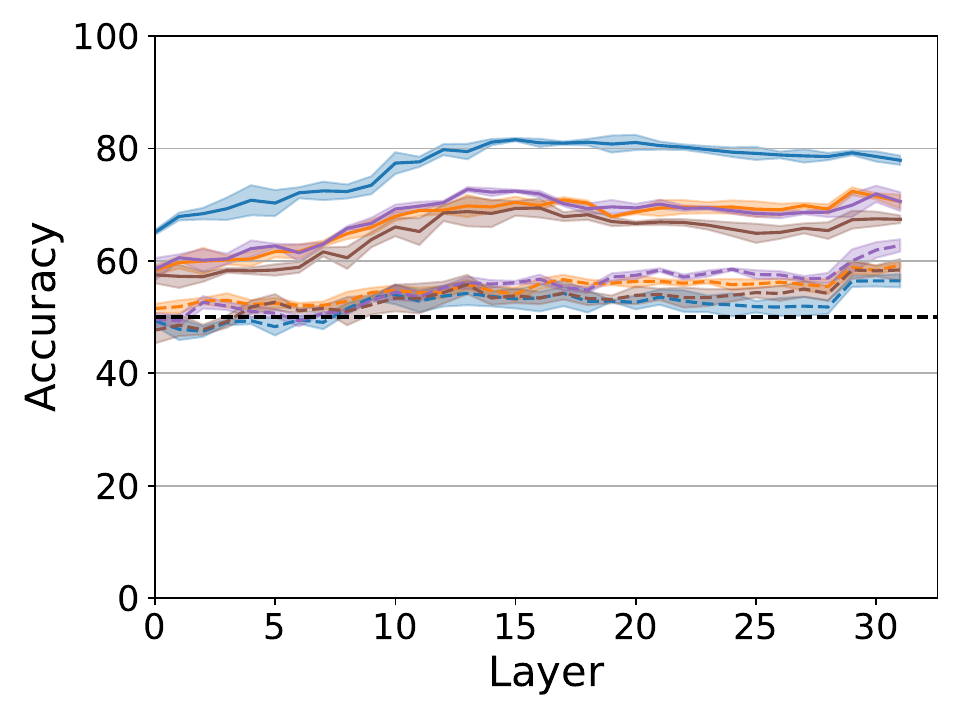}
  \caption{Mistral.}

 \end{subfigure}

 \caption{Distinguishing factually correct from \wak using classifiers trained on generic vs.\ model-specific datasets on Natural Questions. We can see that specific dataset accuracy is significantly higher than generic dataset. }

 \label{fig:non-spesific_results_natural}
\end{figure*}

To complete the results from Section \ref{Detect Model's Hallucinations works better with model's specific dataset} on TriviaQA, we show in Figure \ref{fig:non-spesific_results_natural} similar detection results that compare generic and model's specific dataset on Natural Questions.

\section{Preemptive Hallucination Detection Additional Results}\label{appendix:Preemptive Hallucination Detection Additional results}

\begin{figure*}
\centering
\begin{subfigure}[b]{0.5\textwidth}
  \centering
  \includegraphics[width=0.8\linewidth]{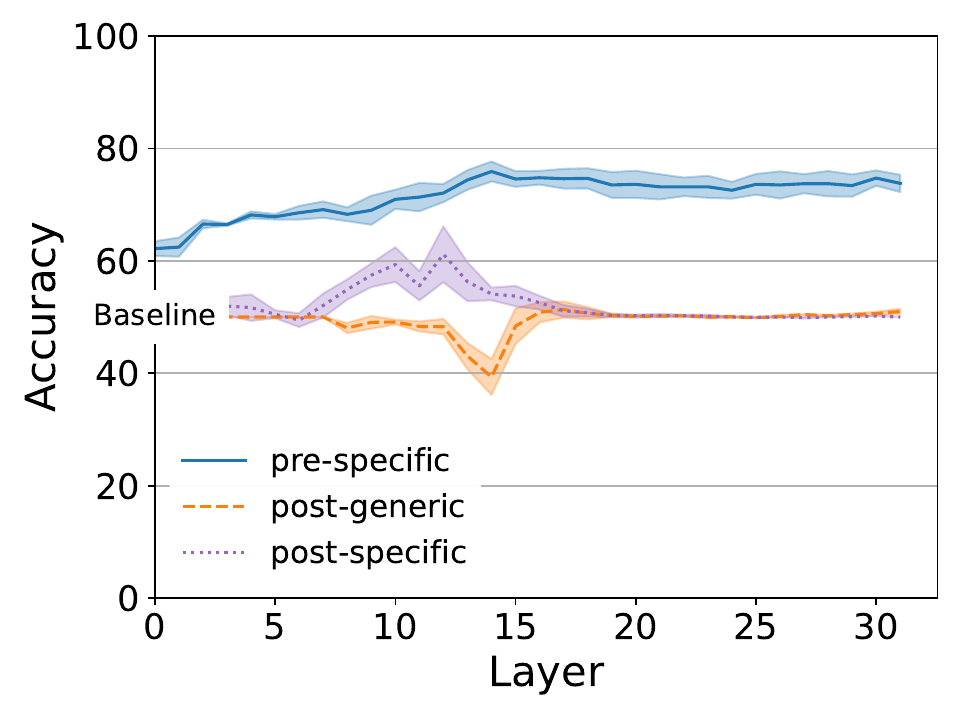}
  \caption{TriviaQA}
 \end{subfigure}%
 \hfill
 \begin{subfigure}[b]{0.5\textwidth}
  \centering
  \includegraphics[width=0.8\linewidth]{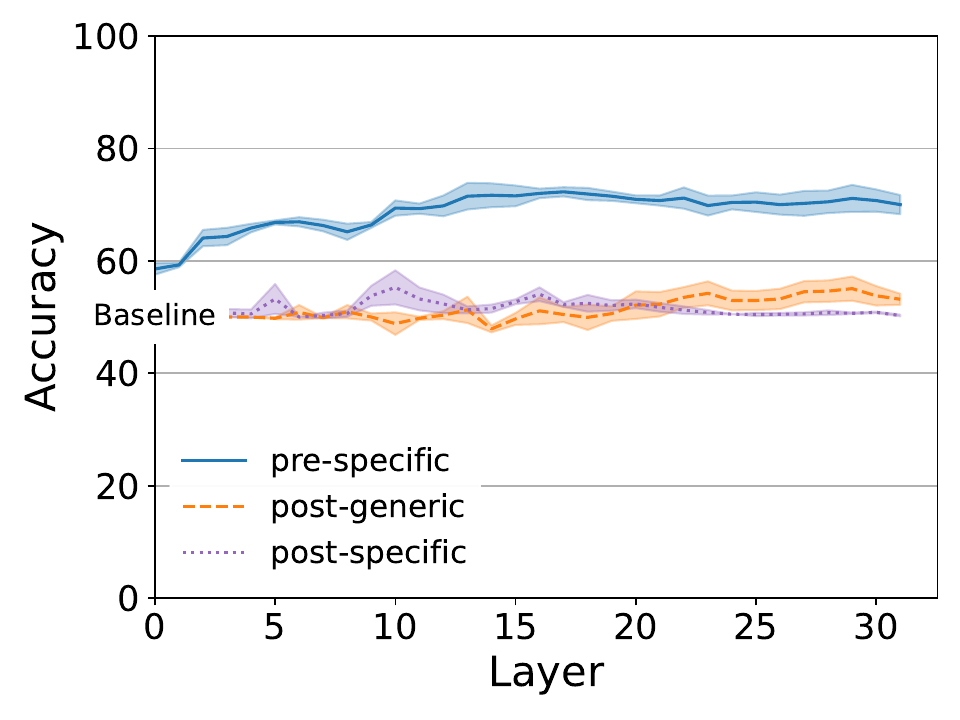}
  \caption{Natural Questions}
 \end{subfigure}

 \caption{Comparing \wak detection before generation using classifiers trained on model-specific pre-hallucination, generic post-hallucination, and model-specific post-hallucination examples on Llama. We can see that preemptive hallucination is effective and that generic dataset can not detect it. }
 \label{fig:spesific_results_prior_metta}
% \end{figure*}
\end{figure*}

\begin{figure*}
\centering
\begin{subfigure}[b]{0.5\textwidth}
  \centering
  \includegraphics[width=0.8\linewidth]{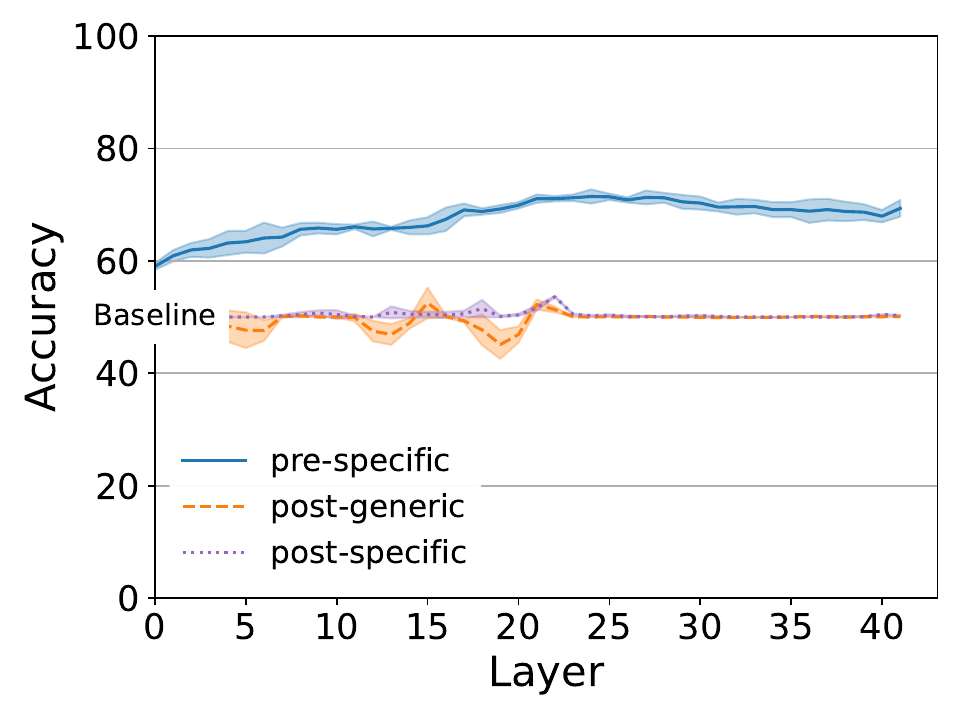}
  \caption{TriviaQA}
 \end{subfigure}%
 \hfill
 \begin{subfigure}[b]{0.5\textwidth}
  \centering
  \includegraphics[width=0.8\linewidth]{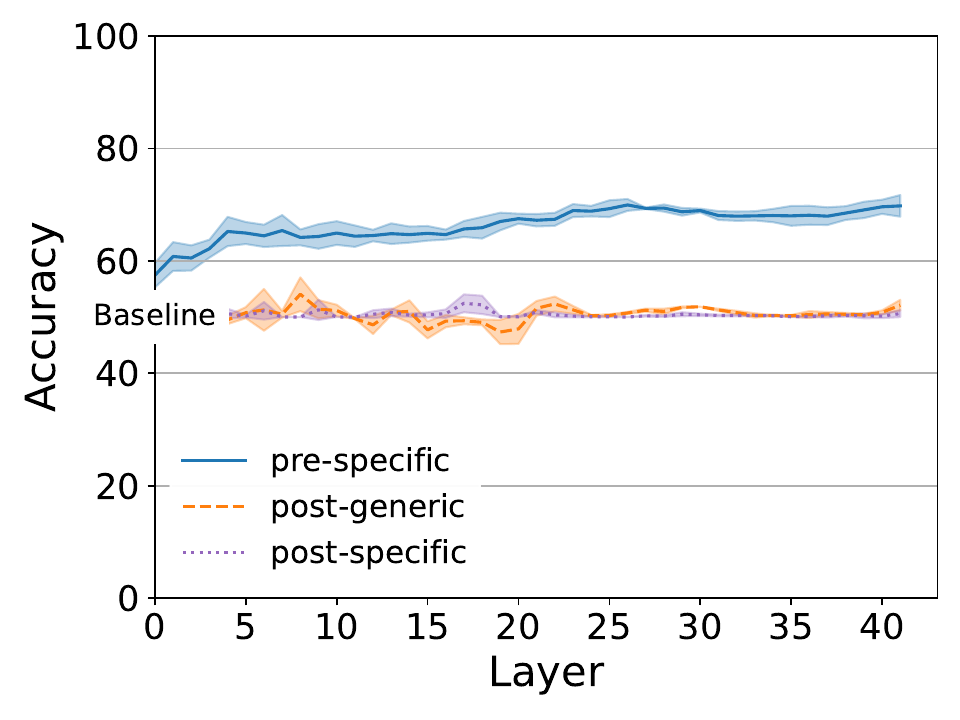}
  \caption{Natural Questions}
 \end{subfigure}

 \caption{Comparing \wak detection before generation using classifiers trained on model-specific pre-hallucination, generic post-hallucination, and model-specific post-hallucination examples on Gemma. We can see that preemptive hallucination is effective and that generic dataset can not detect it. }
 \label{fig:spesific_results_prior_gemma}
% \end{figure*}
\end{figure*}

In Section \ref{Preemptive Hallucination Detection Using Model-Specific Datasets}, we demonstrate that hallucinations can be detected before they are generated, whereas post-hallucination settings are less effective for this purpose. While the main paper presents results using the Mistral model, we extend this evaluation to Llama (Figure \ref{fig:spesific_results_prior_metta}) and Gemma (Figure \ref{fig:spesific_results_prior_gemma}). The results for these models exhibit similar patterns to those observed with Mistral, reinforcing the generality of this phenomenon.

\section{Detection results on the MLP and Attention Components}\label{appendix:Detection results on the MLP and Attention components}
The results in the main paper are only shown using hidden states from the residual component of the LLMs, that is, the representations after each transformer block. To complete the picture, we provide detection results also for the MLP and attention components using the representations that are output by the component.

\begin{figure*}
\centering
% \begin{figure*}
 \centering
\begin{subfigure}[b]{0.32\textwidth}
  \centering
  \includegraphics[width=0.9\linewidth]{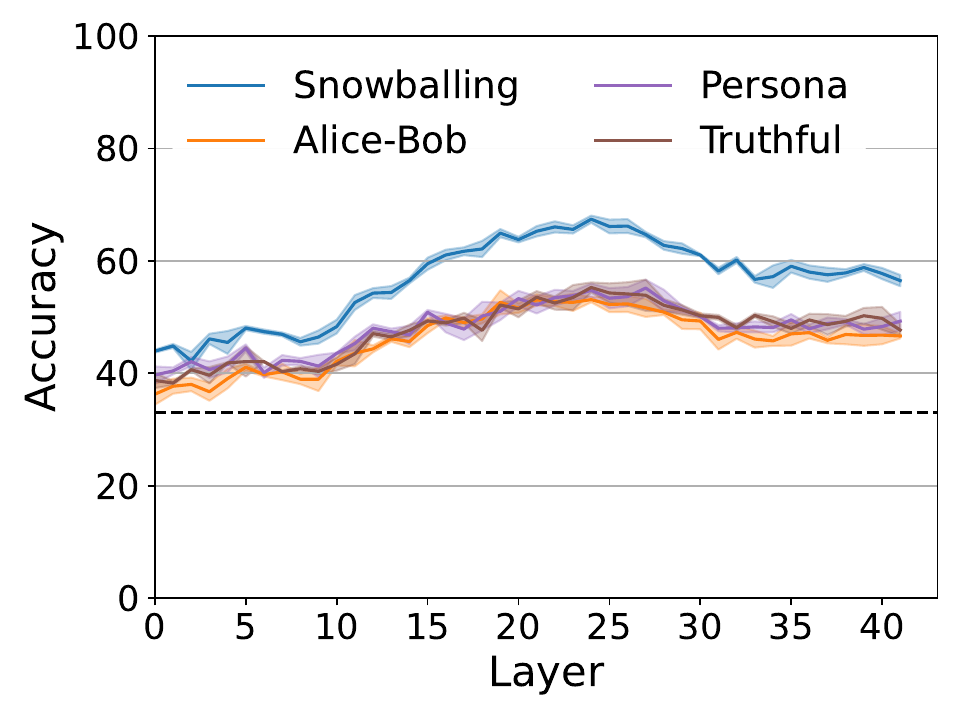}
  \caption{Gemma, TriviaQA}
 \end{subfigure}%
 % \hfill
 \begin{subfigure}[b]{0.32\textwidth}
  \centering
\includegraphics[width=0.9\linewidth]{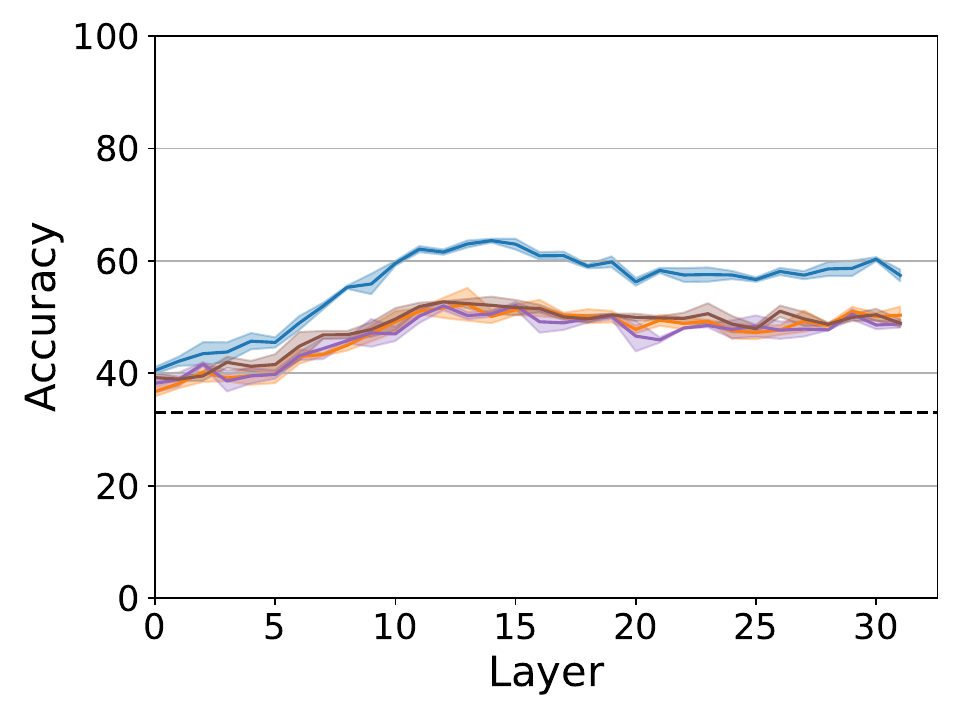}
  \caption{Llama, TriviaQA}
 \end{subfigure}
 \begin{subfigure}[b]{0.32\textwidth}
  \centering
  \includegraphics[width=0.9\linewidth]{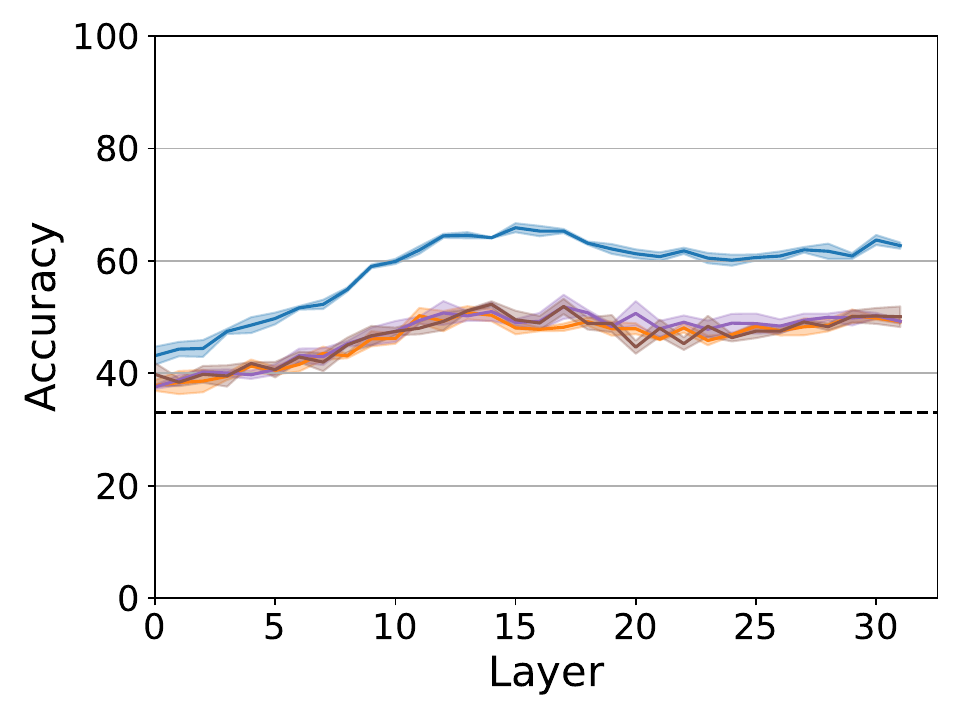}
  \caption{Mistral, TriviaQA}
 \end{subfigure}\\
 % \hfill
 \begin{subfigure}[b]{0.32\textwidth}
  \centering
\includegraphics[width=0.9\linewidth]{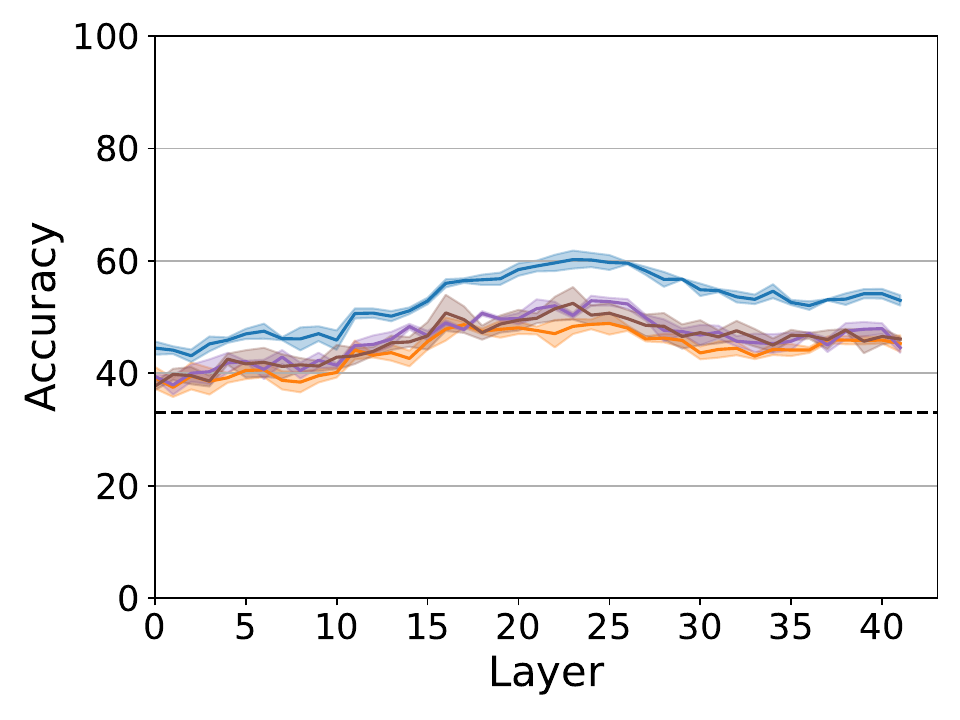}
  \caption{Gemma, Natural Questions}
 \end{subfigure}
 % \hfill
 \begin{subfigure}[b]{0.32\textwidth}
  \centering
\includegraphics[width=0.9\linewidth]{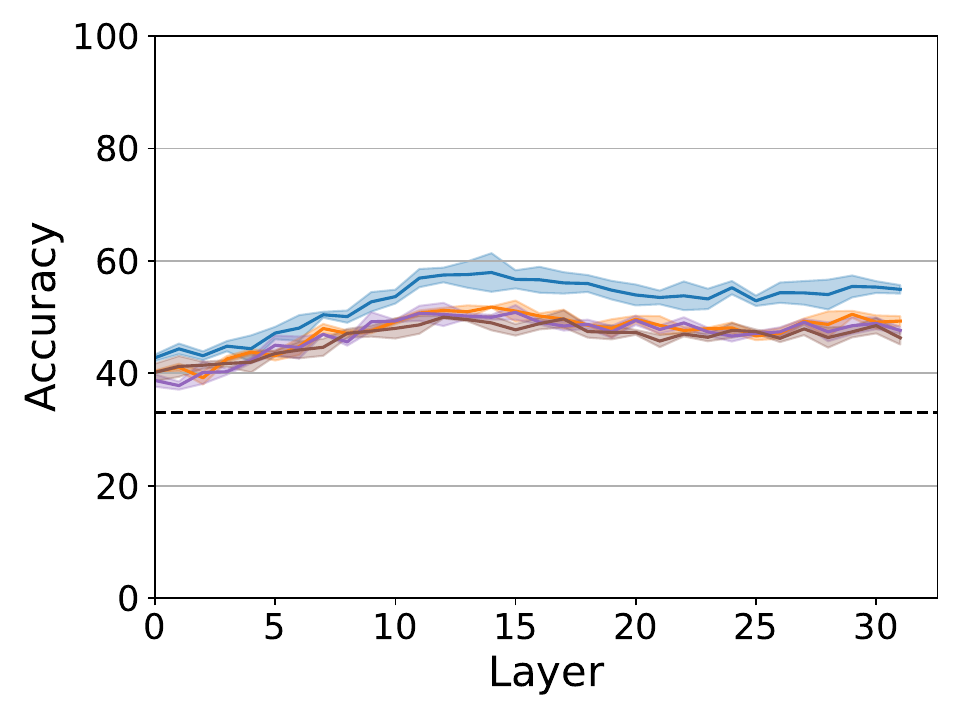}
  \caption{Llama, Natural Questions}
 \end{subfigure}
 % \hfill
 \begin{subfigure}[b]{0.32\textwidth}
  \centering
\includegraphics[width=0.9\linewidth]{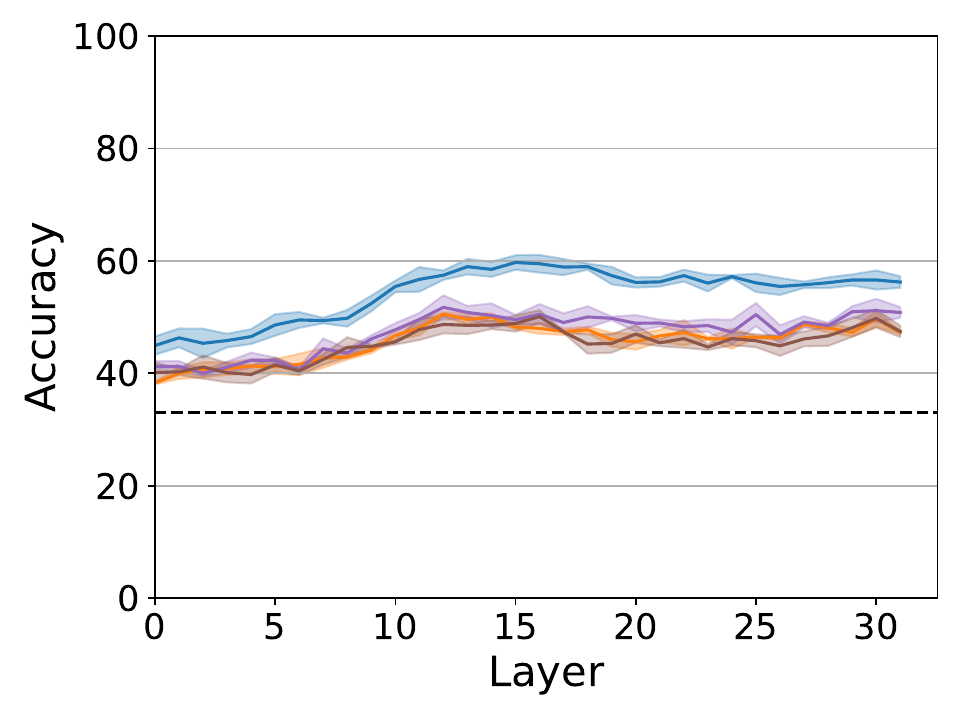}
  \caption{Mistral, Natural Questions}
 \end{subfigure}
 \caption{3-way classification results into (i) hallucinations caused by lack of knowledge (\lok), (ii)  hallucinations caused despite having knowledge (\wak), and (iii)  factually correct examples on the MLP component. We can see that all the detection lines are well above the baseline (dashed line) indicating the possibility of distinguishing \wak from \lok.}
 \label{fig: mlp Hallucinations from miss knowledge vs. hallucinations regardless of knowledge vs. non-hallucination-knowledge classification}
\end{figure*}

\begin{figure*}
\centering

 \centering
\begin{subfigure}[b]{0.32\textwidth}
  \centering
  \includegraphics[width=0.9\linewidth]{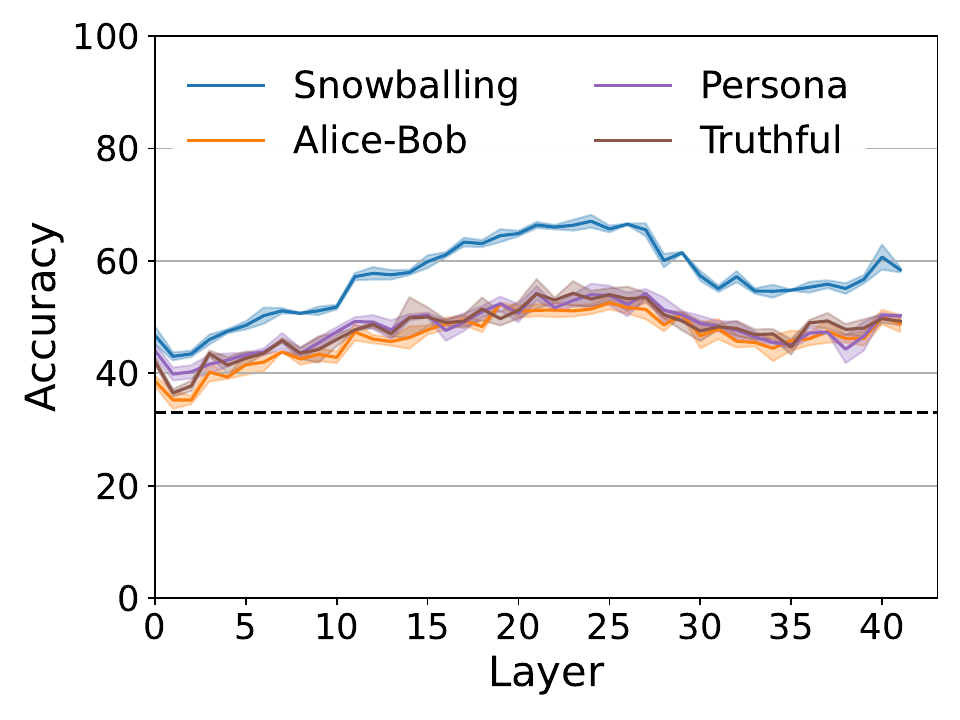}
  \caption{Gemma, TriviaQA}
 \end{subfigure}%
 \begin{subfigure}[b]{0.32\textwidth}
  \centering
\includegraphics[width=0.9\linewidth]{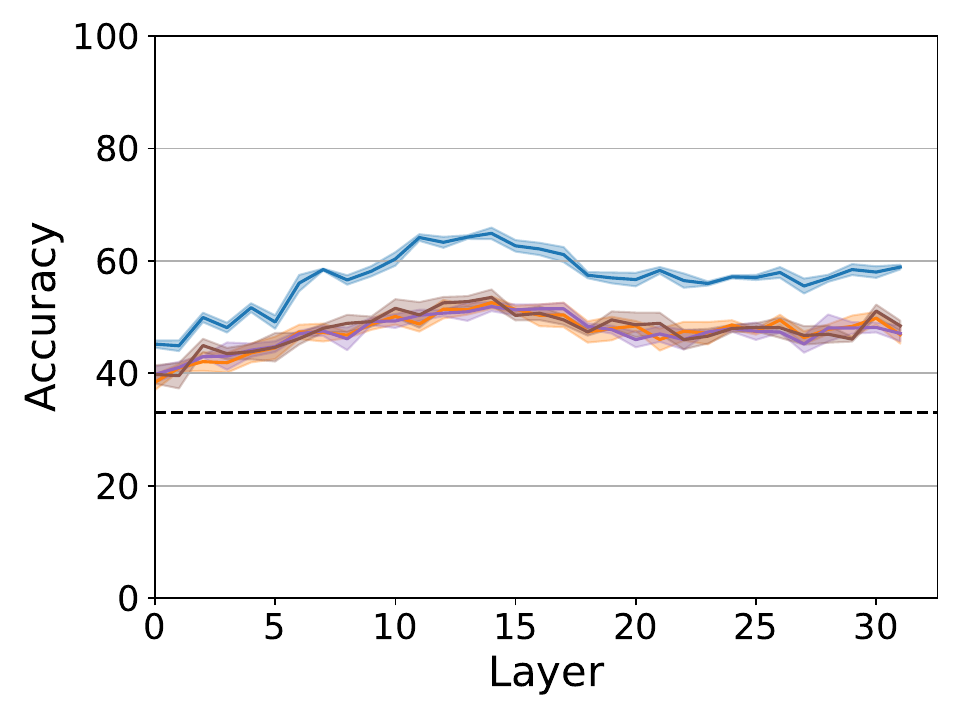}
  \caption{Llama, TriviaQA}
 \end{subfigure}
 \begin{subfigure}[b]{0.32\textwidth}
  \centering
  \includegraphics[width=0.9\linewidth]{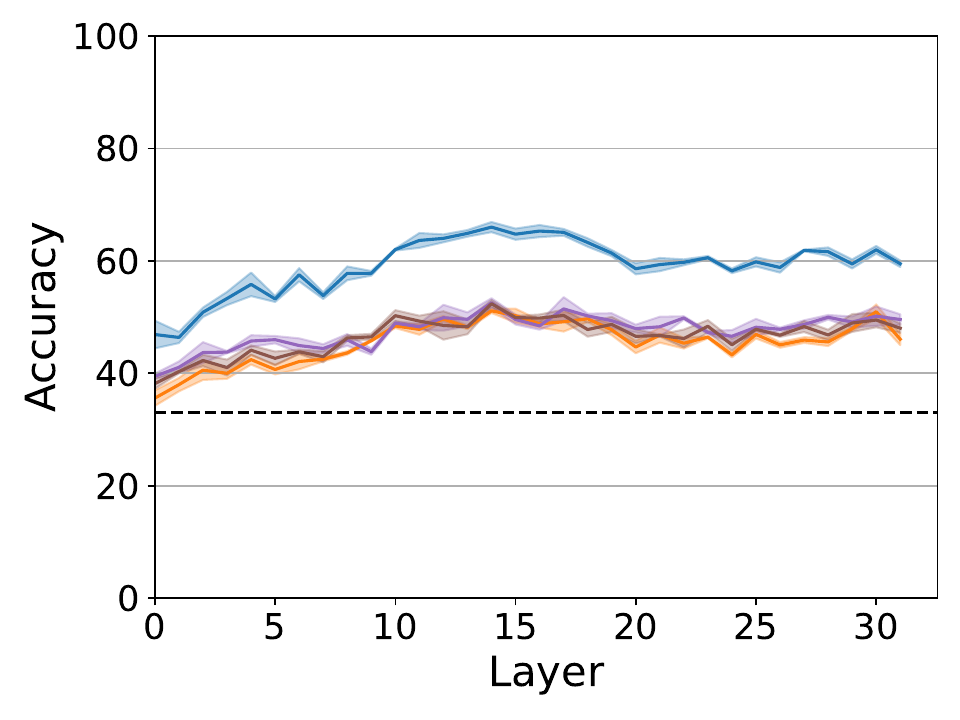}
  \caption{Mistral, TriviaQA}
 \end{subfigure}\\
 \begin{subfigure}[b]{0.32\textwidth}
  \centering
\includegraphics[width=0.9\linewidth]{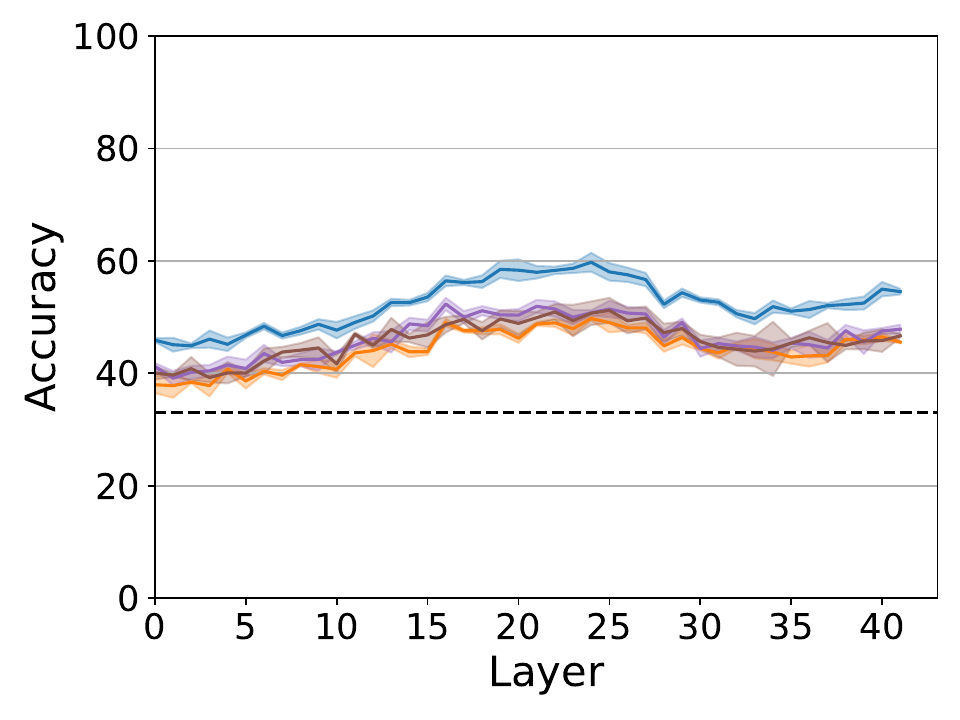}
  \caption{Gemma, Natural Questions}
 \end{subfigure}
 % \hfill
 \begin{subfigure}[b]{0.32\textwidth}
  \centering
\includegraphics[width=0.9\linewidth]{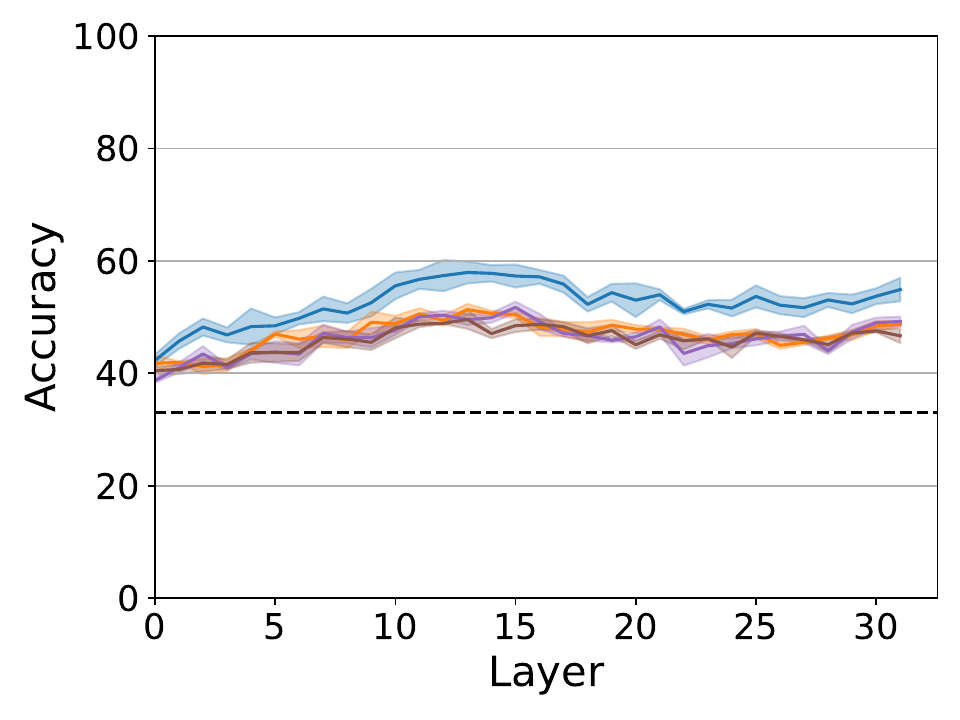}
  \caption{Llama, Natural Questions}
 \end{subfigure}
 % \hfill
 \begin{subfigure}[b]{0.32\textwidth}
  \centering
\includegraphics[width=0.9\linewidth]{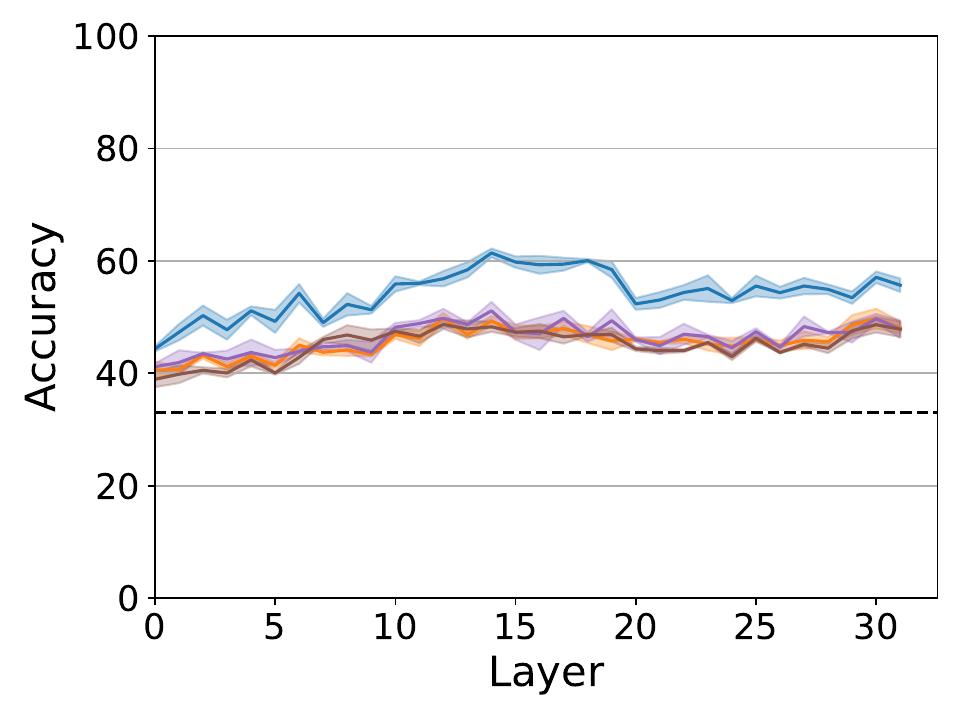}
  \caption{Mistral, Natural Questions}
 \end{subfigure}
 \caption{3-way classification results into (i) hallucinations caused by lack of knowledge (\lok), (ii)  hallucinations caused despite having knowledge (\wak), and (iii)  factually correct examples on the Attention component. We can see that all the detection lines are well above the baseline (dashed line) indicating the possibility of distinguishing \wak from \lok. }
 \label{fig: attention Hallucinations from miss knowledge vs. hallucinations regardless of knowledge vs. non-hallucination-knowledge classification}
\end{figure*}

The results are shown in Figures \ref{fig: mlp Hallucinations from miss knowledge vs. hallucinations regardless of knowledge vs. non-hallucination-knowledge classification} and \ref{fig: attention Hallucinations from miss knowledge vs. hallucinations regardless of knowledge vs. non-hallucination-knowledge classification} for the classification into the two hallucination types and factually correct examples, for all the combinations of datasets and models. 

\begin{figure}
\centering

 \centering
\begin{subfigure}[b]{0.5\textwidth}
  \centering
  \includegraphics[width=0.8\linewidth]{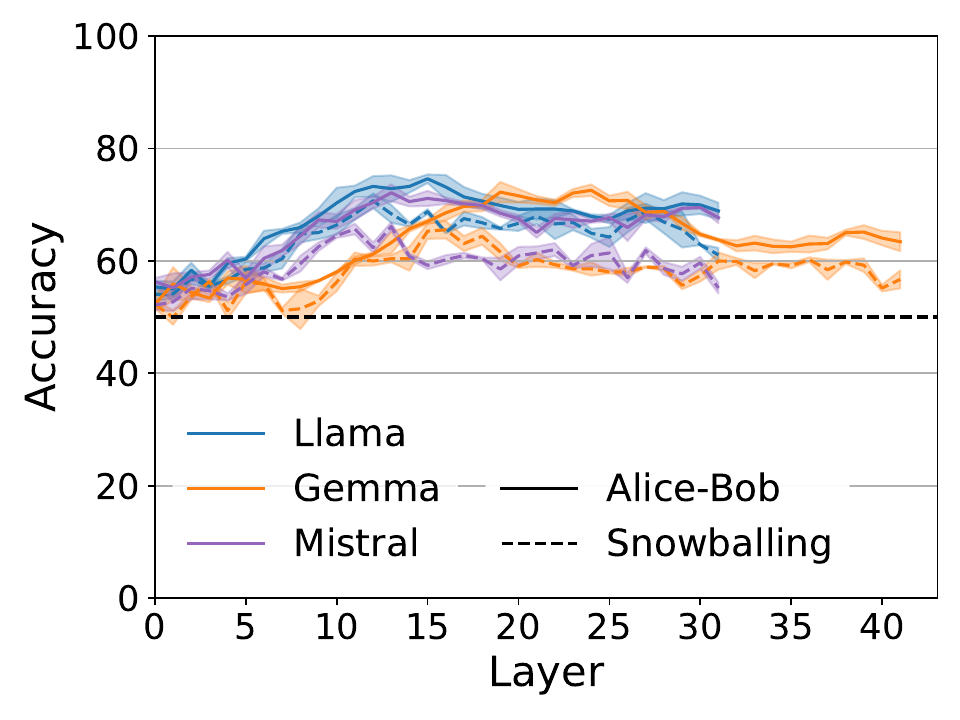}
  \caption{TriviaQA.}
 \end{subfigure}%
 \hfill
 \begin{subfigure}[b]{0.5\textwidth}
  \centering
  \includegraphics[width=0.8\linewidth]{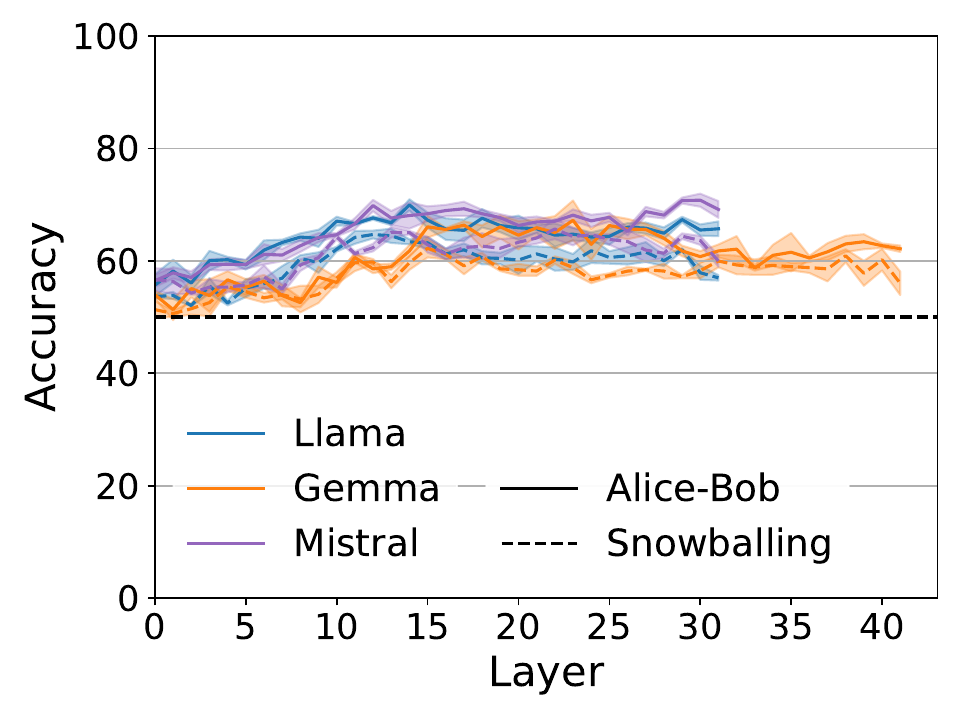}
  \caption{Natural Questions.}
 \end{subfigure}
 \caption{Distinguishing factually correct from \wak, 
 when training on examples from either a Snowballing setting or an Alice-Bob setting, and testing on the Alice-Bob setting. While the change of setting reduces accuracy, the classifier still performs substantially above a random baseline on MLP component.}

 \label{fig:alice_results_mlp}
\end{figure}

\begin{figure}
\centering

 \centering
\begin{subfigure}[b]{0.5\textwidth}
  \centering
  \includegraphics[width=0.8\linewidth]{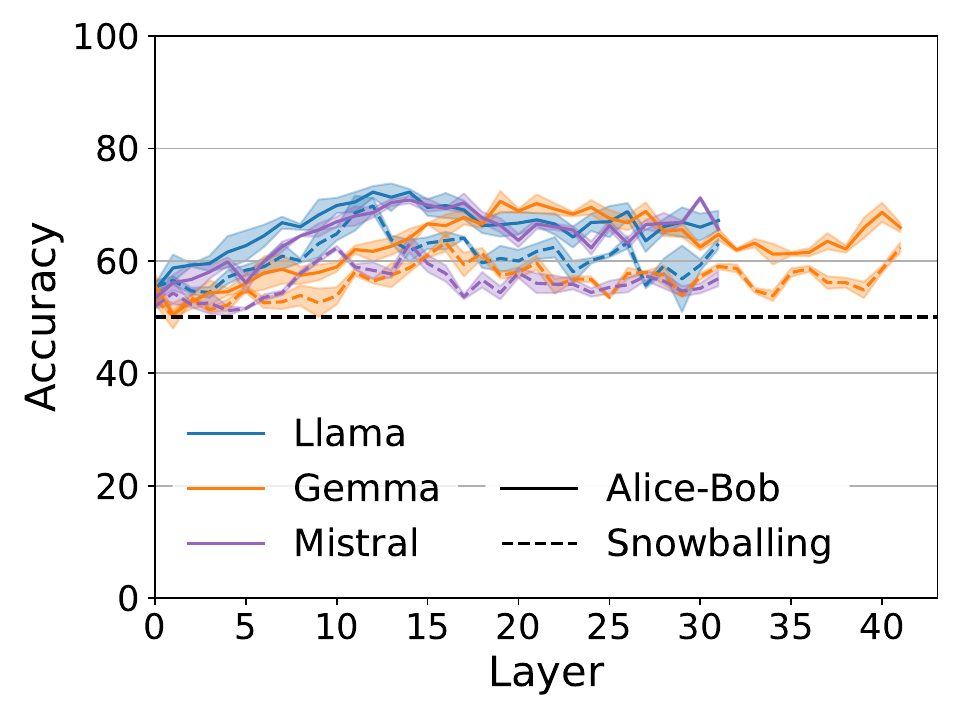}
  \caption{TriviaQA.}
 \end{subfigure}%
 \hfill
 \begin{subfigure}[b]{0.5\textwidth}
  \centering
  \includegraphics[width=0.8\linewidth]{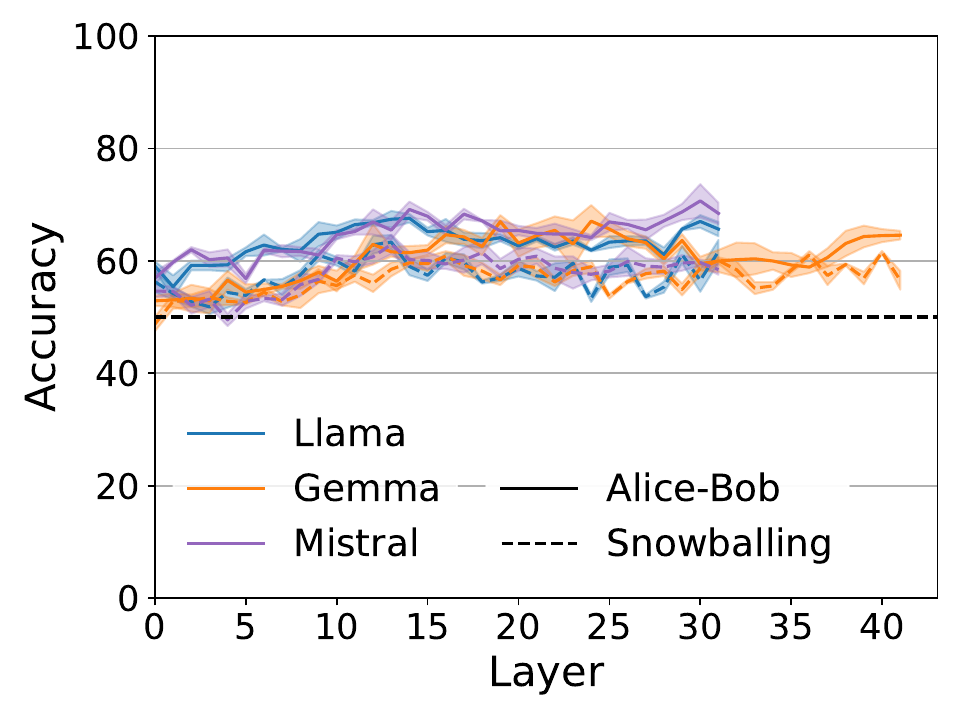}
  \caption{Natural Questions.}
 \end{subfigure}
 \caption{Distinguishing factually correct from \wak, 
 when training on examples from either a Snowballing setting or an Alice-Bob setting, and testing on the Alice-Bob setting. While the change of setting reduces accuracy, the classifier still performs substantially above a random baseline on Attention component.}

 \label{fig:alice_results_attention}
\end{figure}

Next in Figures \ref{fig:alice_results_mlp} and \ref{fig:alice_results_attention} we see the results of the generalization of the Snowballing setting to the Alice-Bob setting using the MLP and Attention components.

\begin{figure*}
\centering

 \centering
\begin{subfigure}[b]{0.3\textwidth}
  \centering
  \includegraphics[width=0.9\linewidth]{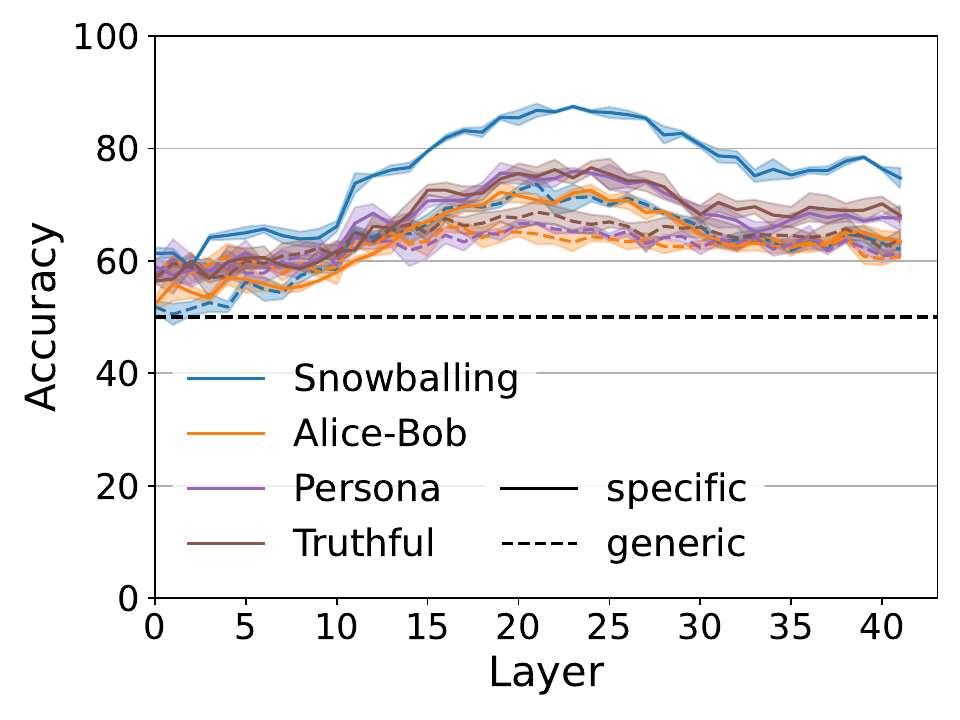}
  \caption{Gemma, TriviaQA}
 \end{subfigure}%
 \hfill
 \begin{subfigure}[b]{0.3\textwidth}
  \centering
  \includegraphics[width=0.9\linewidth]{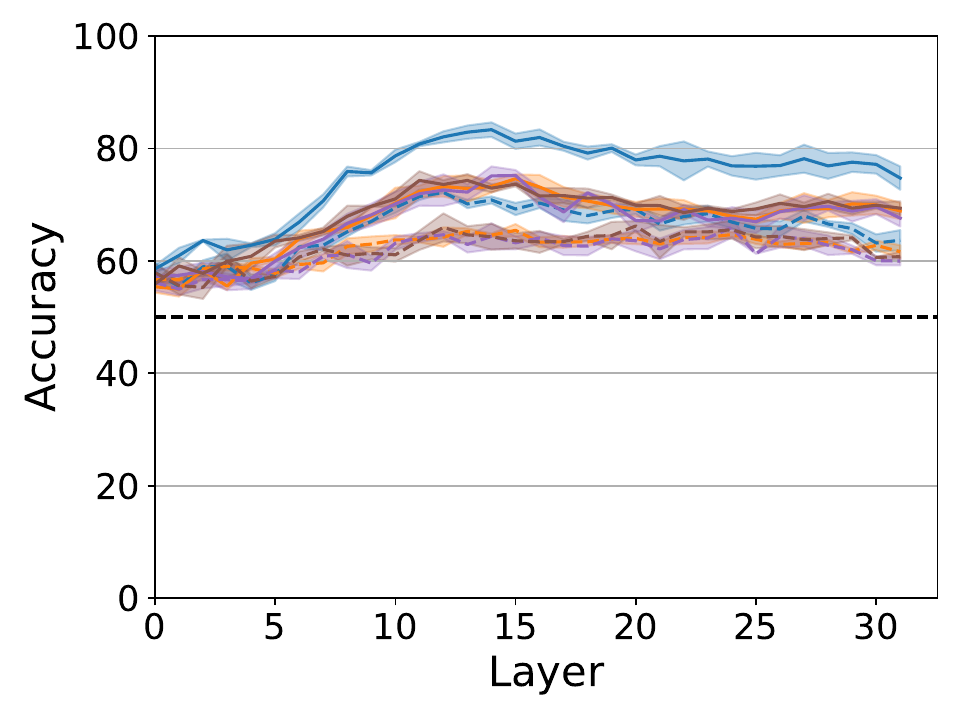}
  \caption{Llama, TriviaQA}
   \end{subfigure}
  \hfill
 \begin{subfigure}[b]{0.3\textwidth}
  \centering
  \includegraphics[width=0.9\linewidth]{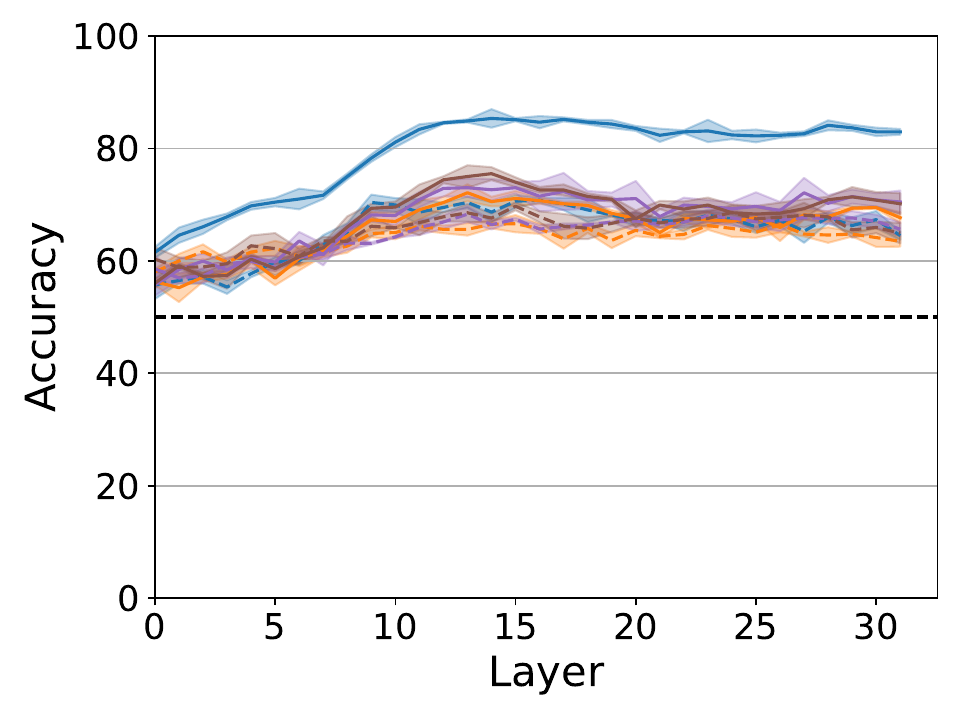}
  \caption{Mistral, TriviaQA}

 \end{subfigure}\\
 \begin{subfigure}[b]{0.3\textwidth}
  \centering
  \includegraphics[width=0.9\linewidth]{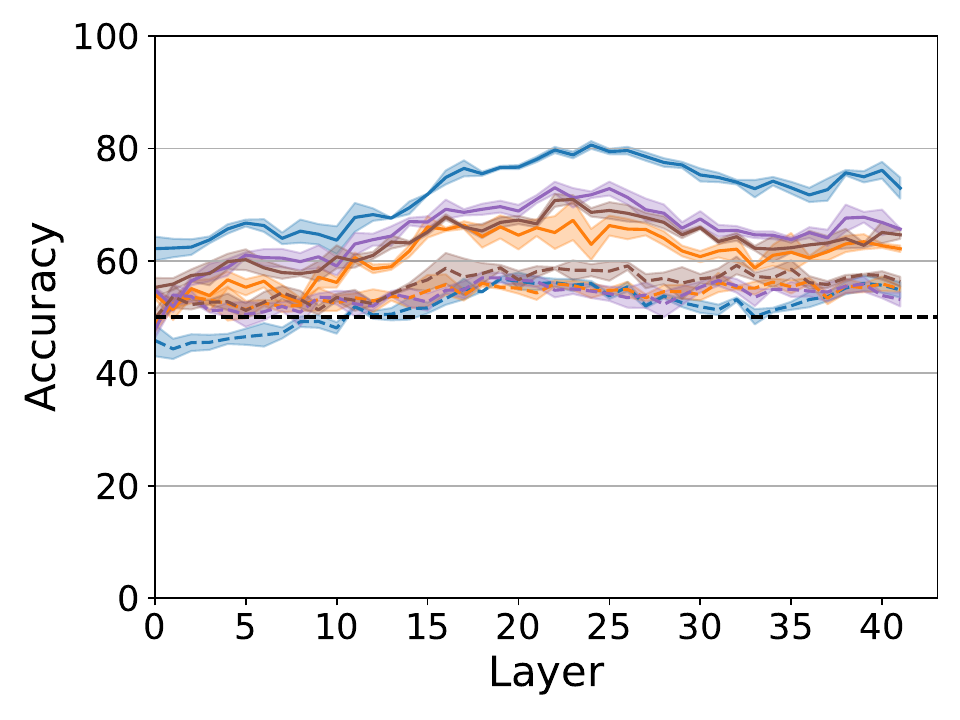}
  \caption{Gemma, Natural Questions}
 \end{subfigure}%
 \hfill
 \begin{subfigure}[b]{0.3\textwidth}
  \centering
  \includegraphics[width=0.9\linewidth]{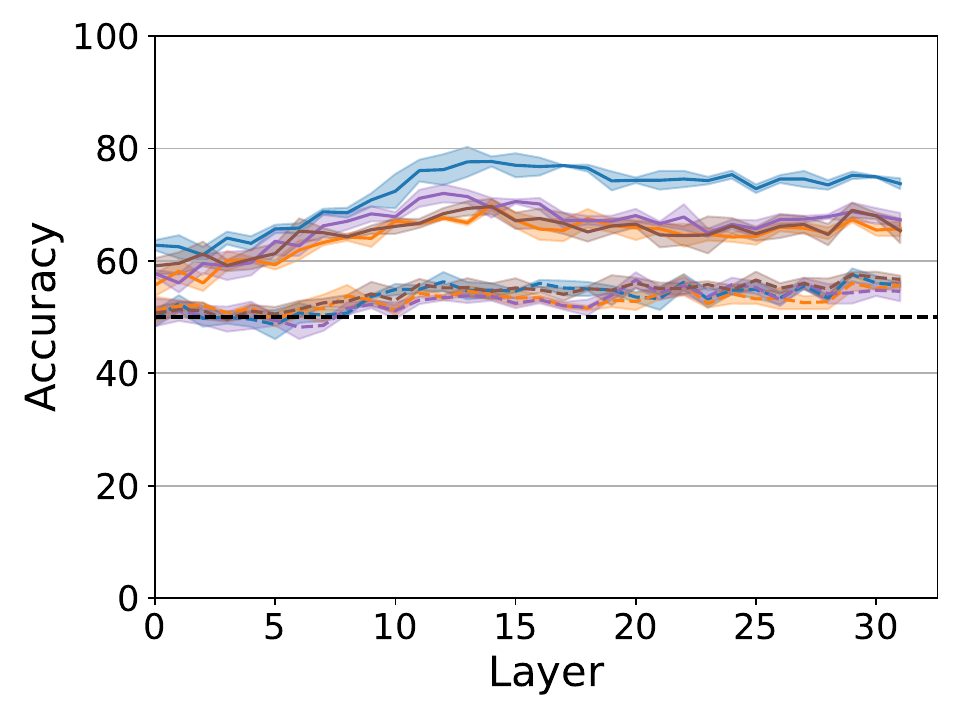}
  \caption{Llama, Natural Questions}
   \end{subfigure}
  \hfill
 \begin{subfigure}[b]{0.3\textwidth}
  \centering
  \includegraphics[width=0.9\linewidth]{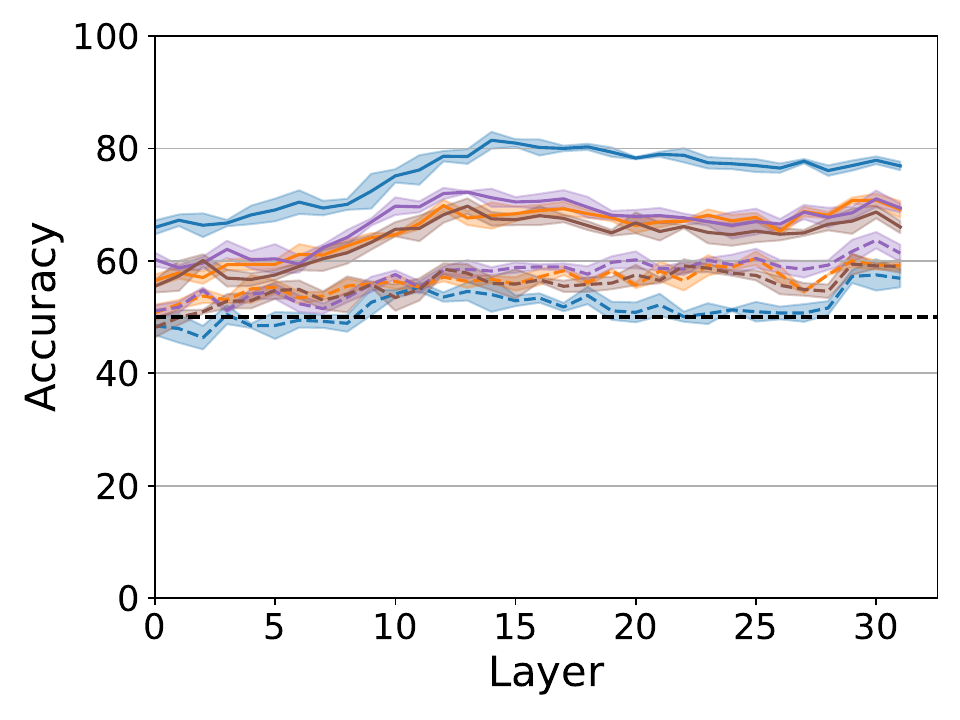}
  \caption{Mistral, Natural Questions}

 \end{subfigure}

 \caption{Distinguishing factually correct from \wak using classifiers trained on generic vs.\ model-specific datasets on MLP component. We can see that specific dataset accuracy is significantly higher than generic dataset. }

 \label{fig:non-spesific_results_mlp}
\end{figure*}

\begin{figure*}
\centering
% \begin{figure*}
 \centering
\begin{subfigure}[b]{0.3\textwidth}
  \centering
  \includegraphics[width=0.9\linewidth]{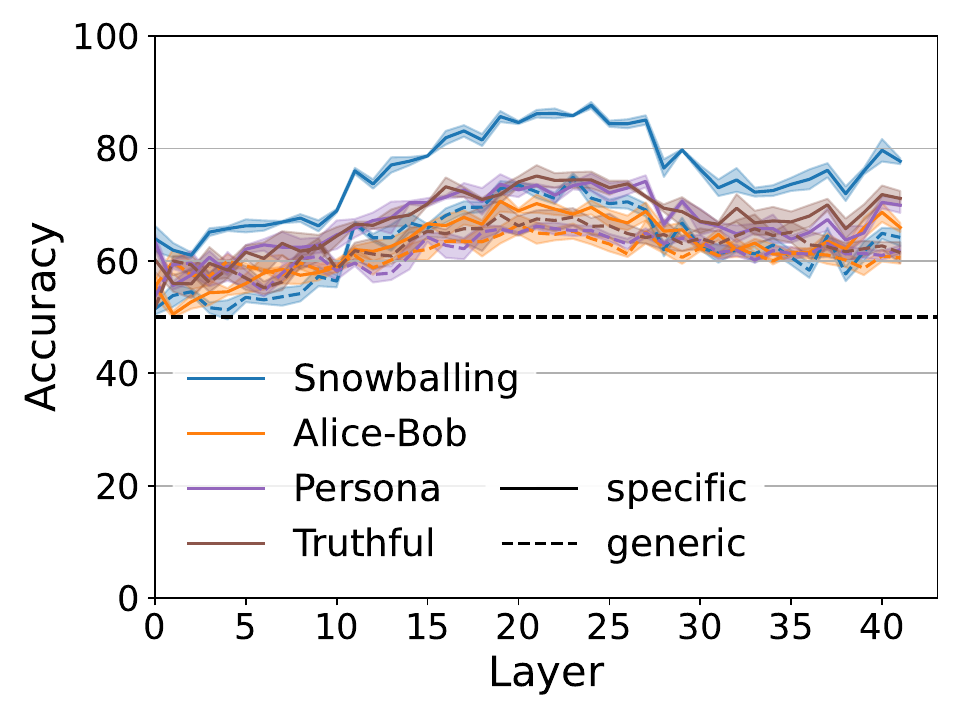}
  \caption{Gemma, TriviaQA}
 \end{subfigure}%
 \hfill
 \begin{subfigure}[b]{0.3\textwidth}
  \centering
  \includegraphics[width=0.9\linewidth]{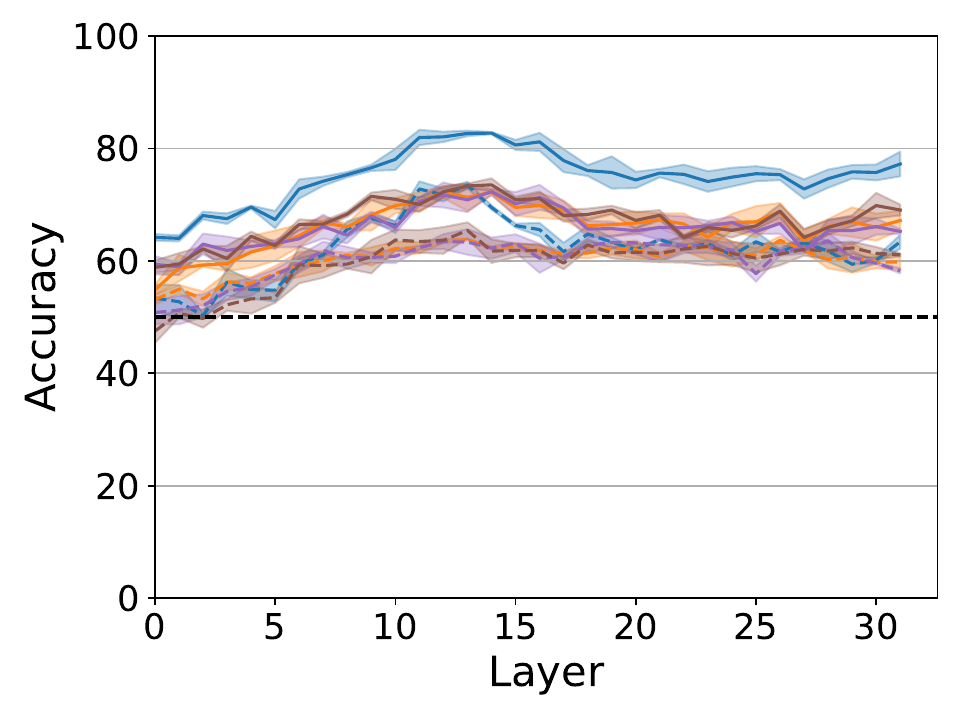}
  \caption{Llama, TriviaQA}
   \end{subfigure}
  \hfill
 \begin{subfigure}[b]{0.3\textwidth}
  \centering
  \includegraphics[width=0.9\linewidth]{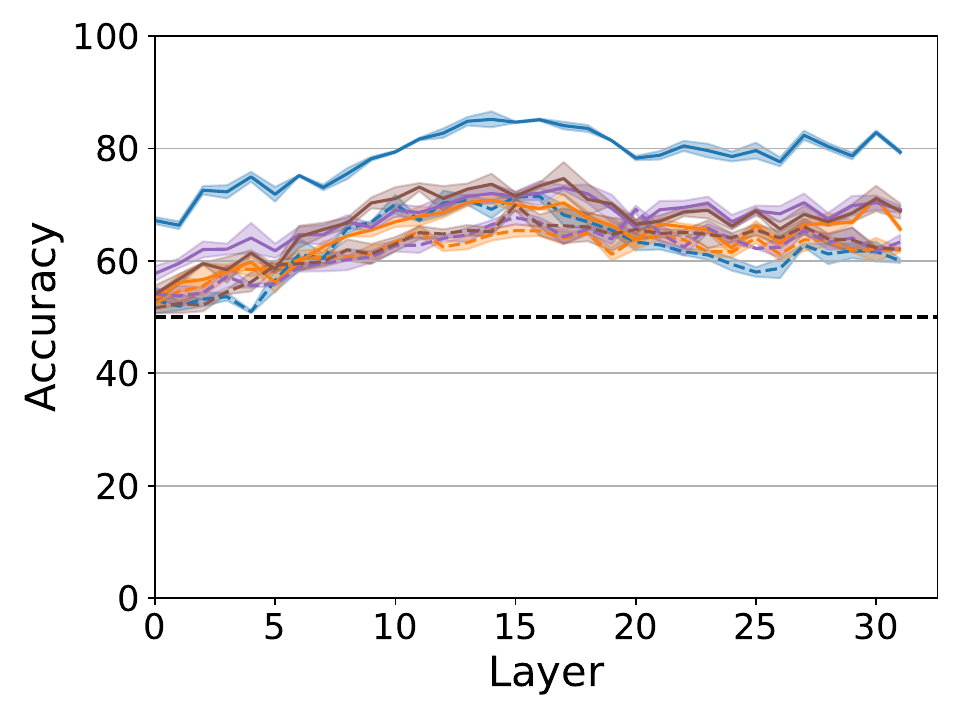}
  \caption{Mistral, TriviaQA}

 \end{subfigure}\\
 \begin{subfigure}[b]{0.3\textwidth}
  \centering
  \includegraphics[width=0.9\linewidth]{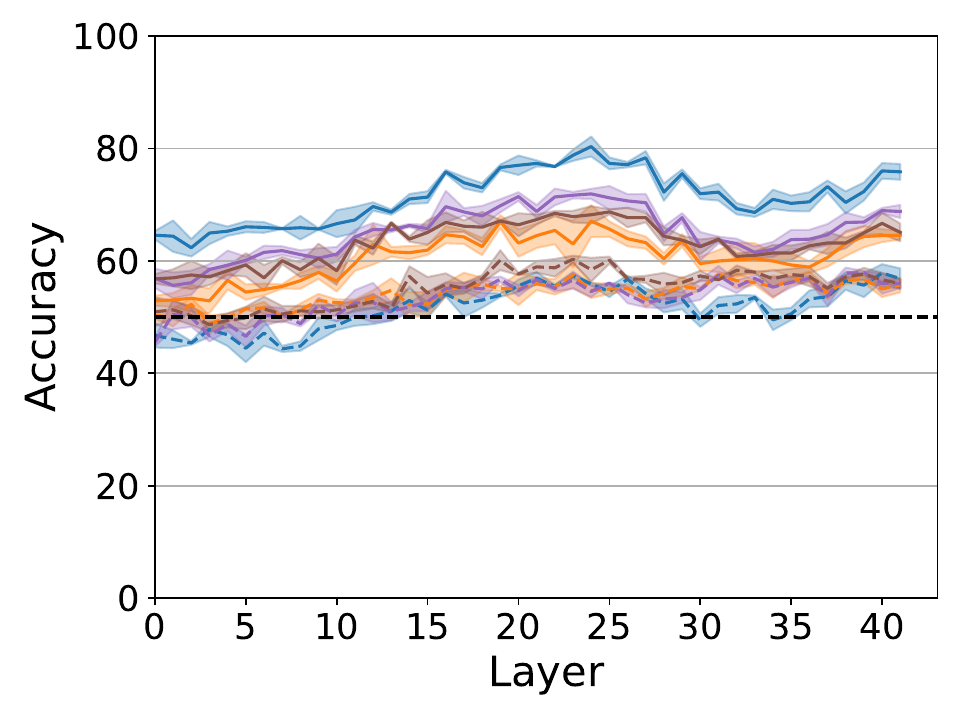}
  \caption{Gemma, Natural Questions}
 \end{subfigure}%
 \hfill
 \begin{subfigure}[b]{0.3\textwidth}
  \centering
  \includegraphics[width=0.9\linewidth]{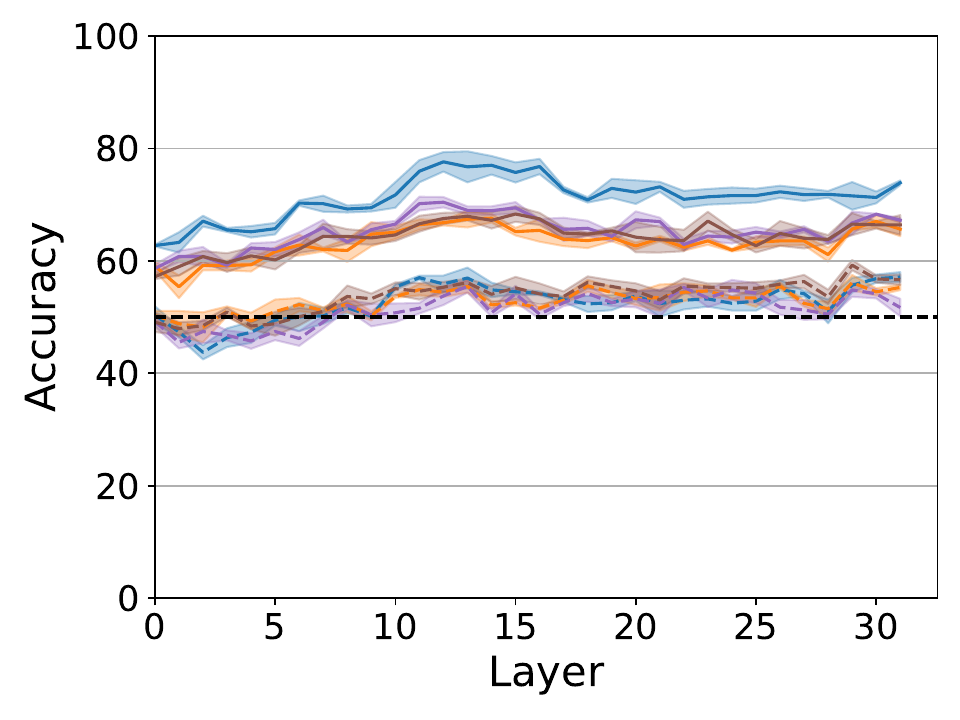}
  \caption{Llama, Natural Questions}
   \end{subfigure}
  \hfill
 \begin{subfigure}[b]{0.3\textwidth}
  \centering
  \includegraphics[width=0.9\linewidth]{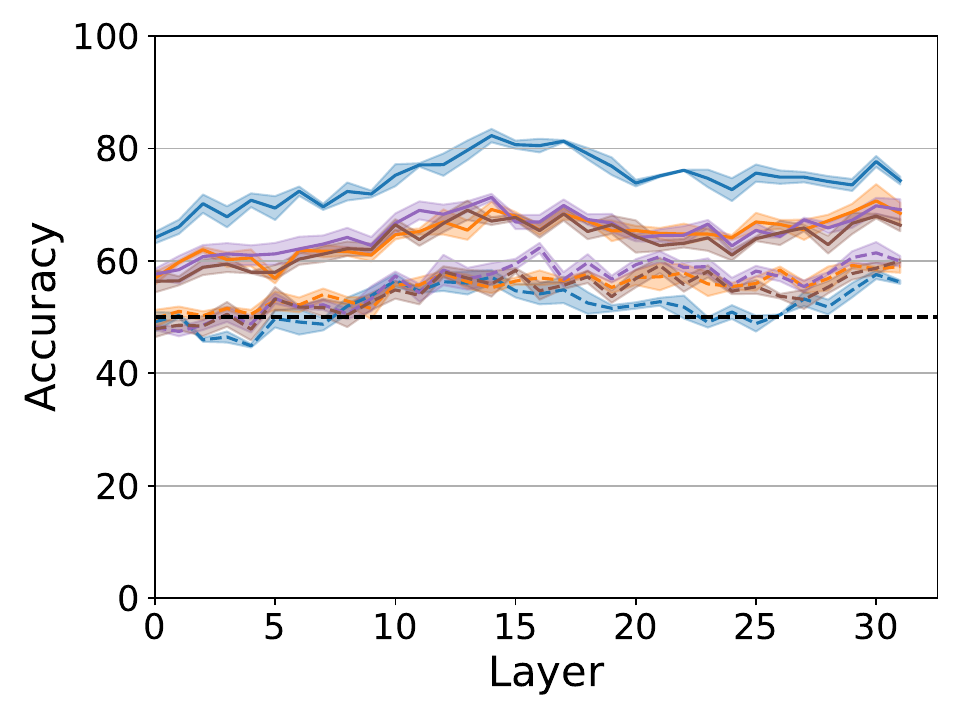}
  \caption{Mistral, Natural Questions}

 \end{subfigure}

 \caption{Distinguishing factually correct from \wak using classifiers trained on generic vs.\ model-specific datasets on Attention component. We can see that specific dataset accuracy is significantly higher than generic dataset. }

 \label{fig:non-spesific_results_attention}
\end{figure*}

Lastly,  Figures \ref{fig:non-spesific_results_mlp} and \ref{fig:non-spesific_results_attention}
show detection of \wak results using classifiers trained on specific and generic datasets.

In all these figures, the results with the MLP and attention components yield similar trends to the ones in the main paper using the residual component, albeit with moderately lower accuracy. This implies that the detection results are not limited to a specific component and are a broader phenomenon across components.

\section{Generalization between Settings}\label{appendix:Generalization between Settings}
\begin{figure*}
\centering

 \centering
\begin{subfigure}[b]{0.5\textwidth}
  \centering
  \includegraphics[width=0.8\linewidth]{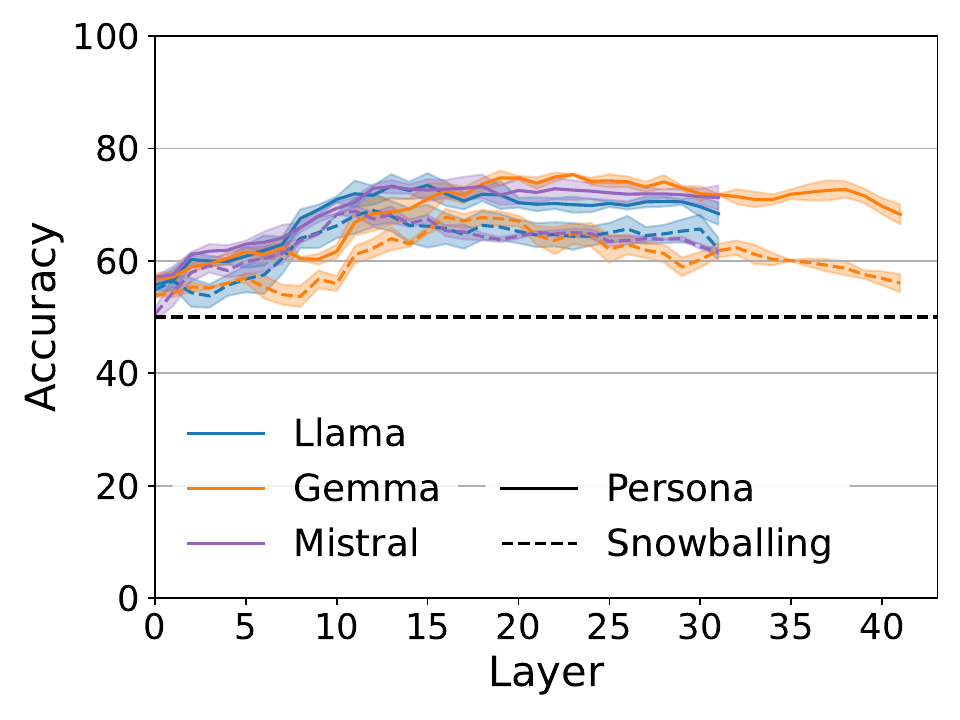}
  \caption{TriviaQA.}
 \end{subfigure}%
 \hfill
 \begin{subfigure}[b]{0.5\textwidth}
  \centering
  \includegraphics[width=0.8\linewidth]{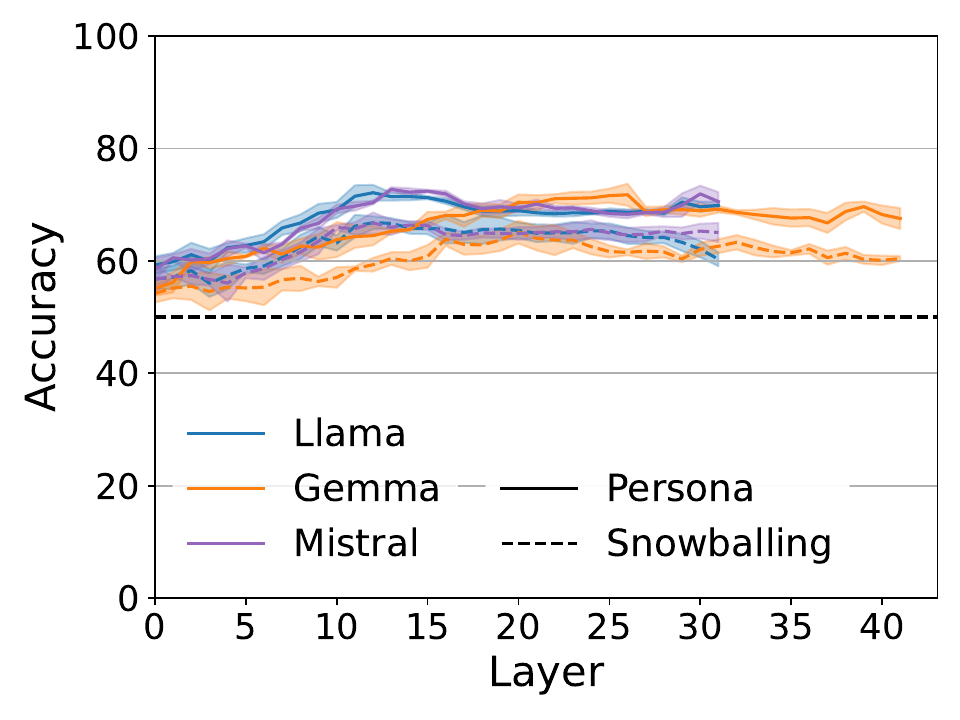}
  \caption{Natural Questions.}
 \end{subfigure}
 \caption{Distinguishing factually correct from \wak, 
 when training on examples from either a Snowballing setting or Persona setting, and testing on the Persona setting. While the change of setting reduces accuracy, the classifier still performs substantially above a random baseline.}
 \label{generalizationpersona}
 \end{figure*}

\begin{figure*}
\centering

 \centering
\begin{subfigure}[b]{0.5\textwidth}
  \centering
  \includegraphics[width=0.8\linewidth]{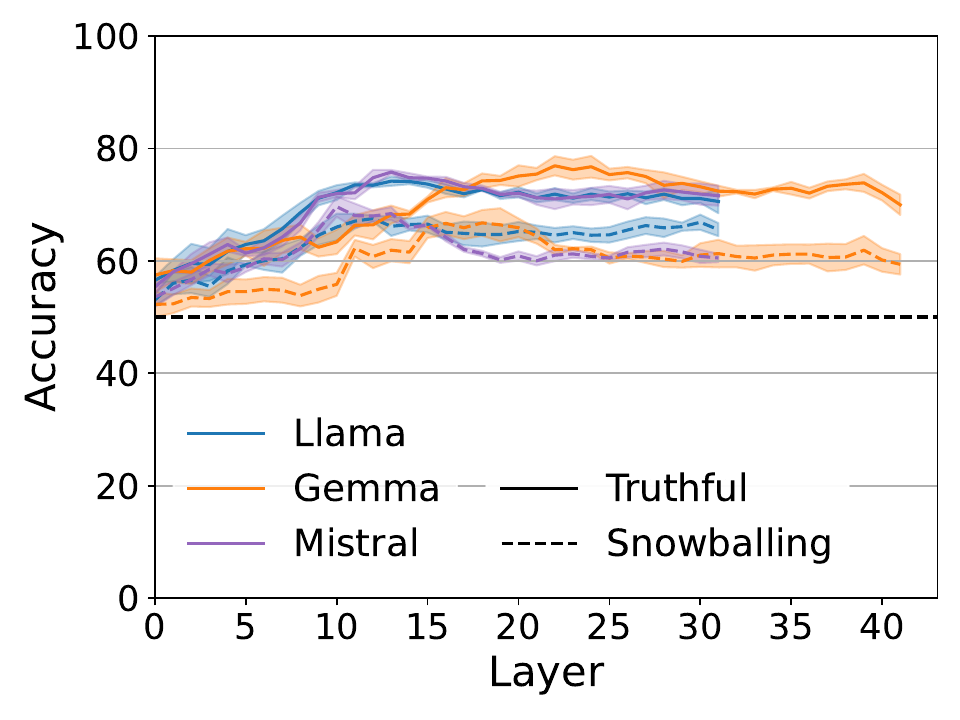}
  \caption{TriviaQA.}
 \end{subfigure}%
 \hfill
 \begin{subfigure}[b]{0.5\textwidth}
  \centering
  \includegraphics[width=0.8\linewidth]{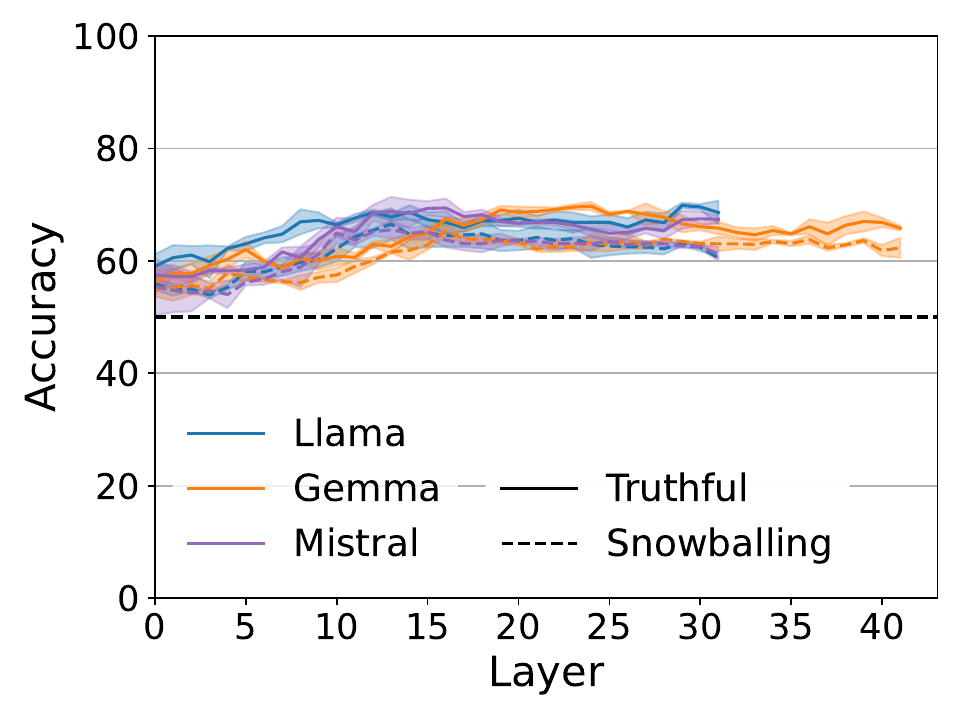}
  \caption{Natural Questions.}
 \end{subfigure}
 \caption{Distinguishing factually correct from \wak, 
 when training on examples from either a Snowballing setting or a Truthful setting, and testing on the Truthful setting. While the change of setting reduces accuracy, the classifier still performs substantially above a random baseline.}
 \label{generalizationTruthful}
 \end{figure*}

In Section \ref{Generalization of WACK hallucinations across hallucination settings}, we demonstrated that the Snowballing setting is generalizable to the Alice-Bob setting. To further validate this generalization, we present Figures \ref{generalizationpersona} and \ref{generalizationTruthful}, which show that Snowballing is also generalizable to the Persona and Truthful settings.

\section{Different snowballing Configuration 
 have Similar Results}\label{appendix:Generalization between Snowballing settings}
In the main paper, we used a single configuration with 3 bad shots, randomly selected. To demonstrate that our results are not dependent on this specific configuration, we conducted additional experiments with a different Snowballing setting. Specifically, we increased the number of shots to five and used a different random seed in the dataset creation process to sample different Snowballing. Our main objective was to show that the classifier trained under this new configuration performs similarly to the original 3-Snowballing-shot setting. Demonstrating this would suggest that variations in Snowballing configurations do not impact classifier performance.

\begin{figure*}
\centering

 \centering
\begin{subfigure}[b]{0.5\textwidth}
  \centering
  \includegraphics[width=0.8\linewidth]{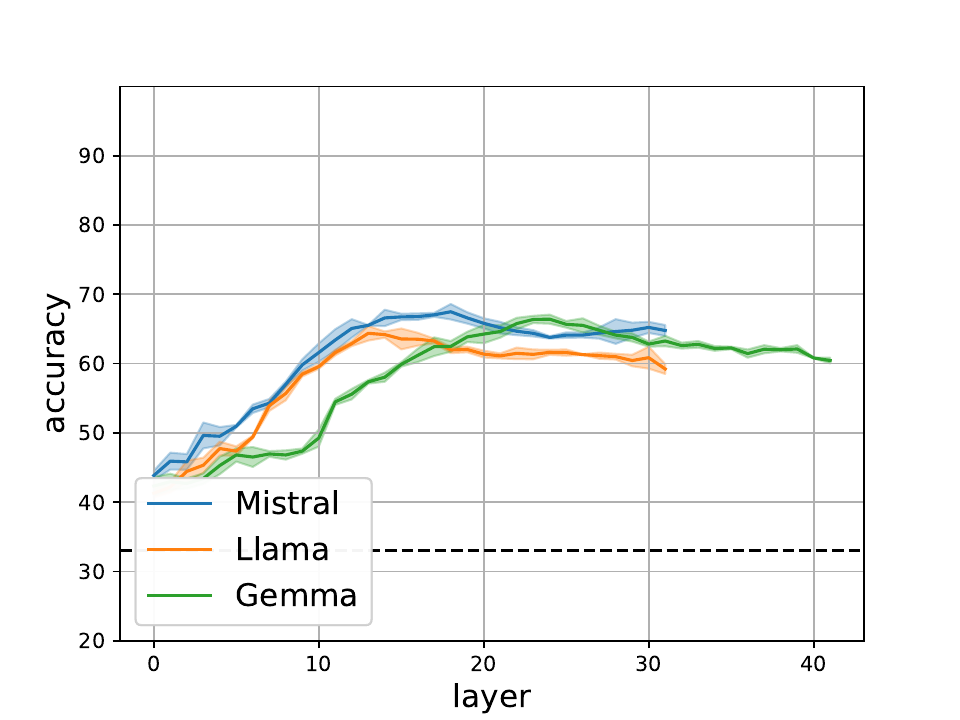}
  \caption{TriviaQA, 5-Snowballing-shots.}
 \end{subfigure}%
 \hfill
 \begin{subfigure}[b]{0.5\textwidth}
  \centering
\includegraphics[width=0.8\linewidth]{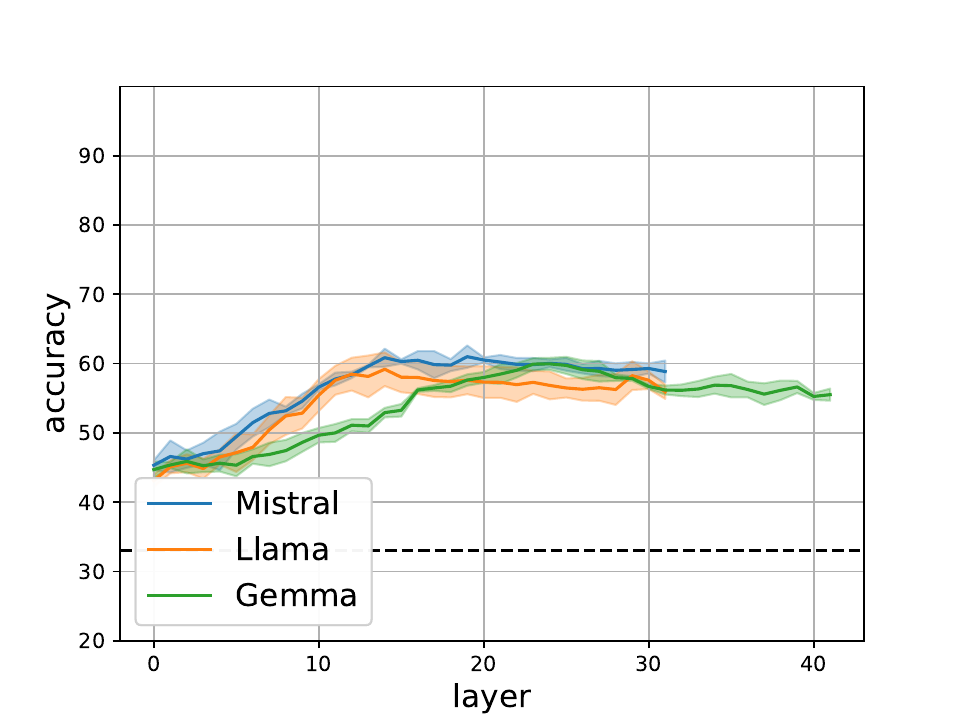}
  \caption{Natural Questions, 5-Snowballing-shots.}
 \end{subfigure}\\
 
 \caption{3-way classification results into (i) hallucinations caused by lack of knowledge (\lok), (ii)  hallucinations caused despite having knowledge (\wak), and (iii)  factually correct examples. We show results using a 5-Snowballing setting.}
 \label{fig:Hallucinations from miss knowledge vs. hallucinations regardless of knowledge vs. non-hallucination-knowledge classification_5 bad shots}
\end{figure*}

In Figure \ref{fig:Hallucinations from miss knowledge vs. hallucinations regardless of knowledge vs. non-hallucination-knowledge classification_5 bad shots} we present the performance of a classifier trained in the 5-Snowballing-shot configuration to distinguish between factually correct answers, \lok, and \wak. These results are comparable to those reported in the main paper in Section \ref{sec_detect_halu_types} for the Snowballing configurations. This consistency indicates that the 5-Snowballing-shots configuration does not affect the classifier's ability to differentiate between the various types of hallucinations.

\end{document}